\documentclass[12pt]{article}

\usepackage{times}

\usepackage[utf8]{inputenc}
\usepackage[T1]{fontenc}
\usepackage{soul, color}
\usepackage{multirow}
\usepackage[ruled,vlined]{algorithm2e}
\usepackage{colortbl}
\usepackage{tabu}
\usepackage{bm}
\usepackage{subfig}
\usepackage{algorithmic}
\usepackage[normalem]{ulem}
\usepackage{pifont}
\usepackage{listings}
\usepackage{makecell}
\usepackage{graphicx}
\usepackage{amsmath}
\usepackage{url}
\usepackage{amssymb}
\usepackage{hyperref}

\newenvironment{sciabstract}{%
\begin{quote} \bf}
{\end{quote}}

\title{SpikingJelly: An open-source machine learning infrastructure platform for spike-based intelligence}

\author
{Wei Fang,$^{1, 2, 3}$ Yanqi Chen,$^{1, 2}$ Jianhao Ding,$^{1}$ Zhaofei Yu,$^{4}$ Timothée Masquelier,$^{5}$ \\Ding Chen,$^{2, 6}$ Liwei Huang,$^{1, 2}$ Huihui Zhou,$^{2}$ Guoqi Li,$^{7, 8\ast}$ Yonghong Tian$^{1, 2, 3\ast}$\\
\\
\normalsize{$^{1}$School of Computer Science, Peking University, China}\\
\normalsize{$^{2}$Peng Cheng Laboratory, China}\\
\normalsize{$^{3}$School of Electronic and Computer Engineering, Shenzhen Graduate School, Peking University, China}\\
\normalsize{$^{4}$Institute for Artificial Intelligence, Peking University, China}\\
\normalsize{$^{5}$Centre de Recherche Cerveau et Cognition (CERCO), UMR5549 CNRS - Univ. Toulouse 3, France}\\
\normalsize{$^{6}$Department of Computer Science and Engineering, Shanghai Jiao Tong University, China}\\
\normalsize{$^{7}$Institute of Automation, Chinese Academy of Sciences, China}\\
\normalsize{$^{8}$School of Artificial Intelligence, University of Chinese Academy of Sciences, China}\\
\normalsize{$^\ast$E-mail: guoqi.li@ia.ac.cn, yhtian@pku.edu.cn}
}
\date{}

\begin{document} 

% Double-space the manuscript.

\baselineskip24pt

% Make the title.

\maketitle

% Place your abstract within the special {sciabstract} environment.

\begin{sciabstract}
Spiking neural networks (SNNs) aim to realize brain-inspired intelligence on neuromorphic chips with high energy efficiency by introducing neural dynamics and spike properties. As the emerging spiking deep learning paradigm attracts increasing interest, traditional programming frameworks cannot meet the demands of the automatic differentiation, parallel computation acceleration, and high integration of processing neuromorphic datasets and deployment. In this work, we present the SpikingJelly framework to address the aforementioned dilemma. We contribute a full-stack toolkit for pre-processing neuromorphic datasets, building deep SNNs, optimizing their parameters, and deploying SNNs on neuromorphic chips. Compared to existing methods, the training of deep SNNs can be accelerated $11\times$, and the superior extensibility and flexibility of SpikingJelly enable users to accelerate custom models at low costs through multilevel inheritance and semiautomatic code generation. SpikingJelly paves the way for synthesizing truly energy-efficient SNN-based machine intelligence systems, which will enrich the ecology of neuromorphic computing.
\end{sciabstract}

% In setting up this template for *Science* papers, we've used both
% the \section* command and the \paragraph* command for topical
% divisions.  Which you use will of course depend on the type of paper
% you're writing.  Review Articles tend to have displayed headings, for
% which \section* is more appropriate; Research Articles, when they have
% formal topical divisions at all, tend to signal them with bold text
% that runs into the paragraph, for which \paragraph* is the right
% choice.  Either way, use the asterisk (*) modifier, as shown, to
% suppress numbering.

\section*{Introduction}
%\subsection*{Intruduction of SNNs and ANNs: version C}
Recently, artificial neural networks (ANNs), such as convolutional neural networks (CNNs)\cite{krizhevsky2012imagenet}, recurrent neural networks (RNNs)\cite{hochreiter1997long} and transformers\cite{vaswani2017attention}, have defeated most other methods and even surpassed the average ability levels of humans in some areas, including image classification \cite{krizhevsky2012imagenet, simonyan2014very, szegedy2015going}, object detection \cite{girshick2014rich, liu2016ssd, redmon2016you}, machine translation \cite{sutskever2014sequence, bahdanau2014neural, sennrich2015neural, vaswani2017attention}, speech recognition \cite{graves2013speech, graves2013hybrid}, and gaming \cite{mnih2015human, silver2016mastering}. These achievements are computer-science-oriented because ANNs are mainly driven by gradient-based numerical optimization methods\cite{rumelhart1986learning, kingma2014adam}, big data\cite{chen2014big, russakovsky2015imagenet} and massively parallel computing with graphics processing units (GPUs) \cite{raina2009large, krizhevsky2017imagenet}. 
Although neuroscience plays a diminished role in ANNs\cite{Goodfellow-et-al-2016}, insights from neuroscience are critical for building general human-level artificial intelligence (AI) systems \cite{goertzel2014artificial, hassabis2017neuroscience}. The human brain is one of the most intelligent systems, possessing overwhelming advantages over any other artificial system in cognition and learning tasks such as transfer learning and continual learning\cite{hassabis2017neuroscience} . The neuroscientific community has been exploring biologically plausible computational paradigms to understand, mimic, and exploit the impressive feats of the human brain. Correspondingly, neuroscience-oriented spiking neural networks (SNNs) have been derived and are sometimes regarded as the next generation of neural networks\cite{maas1997networks}. SNNs process and transmit information using short electrical pulses, or spikes, making them similar to biological neural systems.
%Meanwhile, it cannot be ignored that ANNs are originated in neuroscience, as the name \textit{neural network} indicates\cite{hassabis2017neuroscience}. Neuroscientific observations and mechanisms have made great contribution to ANNs\cite{hassabis2017neuroscience}, e.g., CNNs \cite{lecun1989generalization} inspired from the primary visual cortex\cite{hubel1959receptive} and attention mechanism \cite{larochelle2010learning, denil2012learning} inspired from primate visual system \cite{koch1987shifts, posner1990attention}. Insights from neuroscience are critial to build the human-level general AI \cite{goertzel2014artificial, hassabis2017neuroscience}. In particular, the fact that biological neurons use short electrical pulses, or spikes, to process and transmit information may be key. To mimic these processes, Spiking Neural Networks (SNNs) have been derived and regarded as the next generation of neural networks\cite{maas1997networks}.
The neurons in SNNs are lower-level abstractions of biological neurons that collect signals from dendrites and process stimuli with nonlinear neuronal dynamics, which enable SNNs to be competitive candidates for processing spatiotemporal data \cite{tavanaei2019deep, 10.7554/eLife.65459}. Spike-based biological neural systems are extremely energy-efficient, e.g., the human brain only consumes a power budget of approximately $20$ W\cite{cox2014neural}. Benefiting from event-driven computing on spikes, SNNs are up to 1000$\times$ more power efficient than ANNs \cite{pei2019towards} when running on tailored neuromorphic chips, including True North \cite{merolla2014million}, Loihi \cite{loihi} and Tianjic \cite{pei2019towards}.
The biological plausibility, spatiotemporal information processing capabilities, and event-driven computational paradigm of SNNs have attracted increasing research interest in recent years.

\textbf{Emerging Spiking Deep Learning Methods} Due to their nondifferentiable spike trigger mechanisms and the complex spatiotemporal propagation processes, designing high-performance learning algorithms for SNNs is challenging. Traditional learning algorithms mainly incorporate biologically plausible unsupervised learning rules and primitive gradient-based supervised learning methods.
Unsupervised learning algorithms inspired by the biological nervous system are applied in SNNs, including Hebbian learning \cite{hebb1949the}, Spike Timing Dependent Plasticity (STDP) \cite{bi1998synaptic}, and their variants \cite{Masquelier2007, Kheradpisheh2018,  liu2020effective, 8891738, taherkhani2020review}. Primitive supervised learning methods including SpikeProp \cite{BOHTE200217}, Tempotron \cite{tempotron}, ReSume \cite{ponulak2010supervised}, and SPAN \cite{mohemmed2012span}, achieve higher performance than biologically plausible unsupervised methods. However, these approaches are limited. Most SpikeProp-based methods only allow spiking neurons to fire no more than one spike, while Tempotron, ReSume, and SPAN cannot train SNNs with more than one layer. Thus, these primitive supervised learning methods can only solve tasks that are no harder than classifying the Modified National Institute of Standards and Technology (MNIST) dataset \cite{MNIST}.

One of the key technologies that has led to the rapid progress of ANNs is deep learning \cite{deep-learning-nature}, which optimizes the parameters of multilayer ANNs via backpropagation and learns high-dimensional representations of data with multiple levels of abstraction.
To overcome the challenge of training SNNs, researchers have explored the application of deep learning methods to SNNs and achieved substantial performance improvements. Two of the most commonly used deep learning methods for SNNs are the surrogate gradient method \cite{STBP, shrestha2018slayer, neftci2019surrogate} and the ANN-to-SNN conversion (ANN2SNN) \cite{hunsberger2015spiking, cao2015spiking, Bodo2017Conversion, deng2021optimal, li2021free, stockl2021optimized}. SNNs trained by the surrogate gradient method achieve high performance on complex datasets such as the Canadian Institute for Advanced Research (CIFAR) dataset \cite{CIFAR10}, the Dynamic Vision Sensor (DVS) Gesture dataset \cite{Amir_2017_CVPR}, and the challenging ImageNet dataset \cite{russakovsky2015imagenet} using only a few simulation time steps \cite{fang2021incorporating, neunorm, 10.3389/fnins.2019.00095, zheng2020going, SEWResNet, zhou2023spikformer}, while SNNs converted from ANNs attain almost the same accuracy as that of the original ANNs on the ImageNet dataset with dozens of simulation time steps \cite{han2020rmp, han2020deep, deng2021optimal}. Due to the rapid progress achieved by deep learning methods, the applications of SNNs have been expanded beyond classification to other tasks including object detection\cite{kim2020spiking, cordone2022object, barchid2022spiking}, object segmentation\cite{patel2021spiking, parameshwara2021spikems}, depth estimation \cite{ranccon2021stereospike} and optical flow estimation\cite{lee2020spike}. The boom exhibited by the research community indicates that spiking deep learning has become a promising research hot spot.

% 3. issues: lack of frameworks to: accelerate SNNs with GPU, deploy SNNs to neuromorphic chips, a full-stack toolkit
\textbf{Demands for Frameworks} Experience derived from the development of ANNs shows that software frameworks play a vital role in deep learning.
Modern frameworks, including TensorFlow \cite{abadi2016tensorflow}, Keras\cite{chollet2015keras} and PyTorch\cite{NEURIPS2019_9015}, provide user-friendly frontend application programming interfaces (APIs) implemented by Python and high-performances backend accelerated by C++ libraries, e.g., Intel MKL and Nvidia CUDA. Statistics show that the numbers of both new adopters and projects increase exponentially after the release of modern frameworks \cite{8595219}, indicating that these frameworks substantially reduce the workload required to build and train ANNs, help users quickly realize ideas, and make great contributions to the growth of deep learning research. The rapid development of machine learning frameworks has additionally accelerated the progress of the research community, which also highlights the importance of frameworks for spiking deep learning.

However, there is no mature framework available for spiking deep learning. Most existing frameworks for SNNs, including NEURON \cite{carnevale2006neuron}, NEST \cite{gewaltig2007nest}, Brian1/2 \cite{10.3389/neuro.11.005.2008, Stimberg2019}, and GENESIS\cite{cornelis2012python}, can build detailed spiking neurons with complex neuronal dynamics, use numerical methods to approximate ordinary differential equations (ODEs), and simulate the biological neural system with high precision but do not integrate automatic differentiation, which is the core component required for gradient-based deep learning. These frameworks construct SNNs that are highly biologically plausible and can be used to investigate the functionality of real neural systems, but are not designed to solve machine learning tasks. Nengo \cite{bekolay2014nengo}, SpykeTorch \cite{10.3389/fnins.2019.00625} and BindsNET \cite{10.3389/fninf.2018.00089} use simplified neurons with smaller numbers of ODEs than detailed neurons. With their low computational complexity brought by simplified neurons, these frameworks can implement some primary machine learning and reinforcement learning algorithms but still lack the modern deep learning capabilities of SNNs. 

Due to the lack of available frameworks, researchers who want to combine advanced deep learning methods with SNNs have to build basic spiking neurons and synapses from scratch, resulting in repetitive and uncoordinated efforts.
Deep SNNs involve a large number of matrix operations in both spatial and temporal dimensions of the data, which requires researchers to refine their codes to create high-performance programs that are accelerated by GPUs. Such workloads increase the burden on researchers. In the neuromorphic community, SNNs are frequently employed to process data from neuromorphic sensors and deployed on neuromorphic computing chips, but data processing and deployment also require considerable time and effort.
After skillful researchers implement their projects, inconsistent programming languages, coding styles, and model definitions generated by different authors will be difficult to reuse and divide the community. 
The efficiency of scientific research can be greatly improved if there exists a modern spiking deep learning framework that possesses at least the following three characteristics: exploits and accelerates spike-based operations; supports both simulations on CPUs/GPUs and deployment on neuromorphic chips; and provides a full-stack toolkit for building, training, and analyzing deep SNNs.

\begin{figure}
	\centering
	% \begin{minipage}{0.62\linewidth}}
	% \centering
	% \subfloat[]{\includegraphics[width=1..\textwidth,trim=0 500 720 30,clip]{./fig/framework.pdf}}
	% \end{minipage}
% \begin{minipage}{0.37\linewidth}
	% \centering
	% \subfloat[]{\includegraphics[width=1..\textwidth,trim=18 540 218 18,clip]{./fig/api_example3.pdf}}
	% 
	% \subfloat[]{\includegraphics[width=1.\textwidth,trim=100 660 220 22,clip]{./fig/code_example_figure.pdf}}
	% \end{minipage}
% \subfloat[]{\includegraphics[width=0.5\textwidth,trim=30 610 340 60,clip]{./fig/cmp_frameworks.pdf}}
% \subfloat[]{\includegraphics[width=0.5\textwidth,trim=30 200 310 40,clip]{./fig/application.pdf}}
% \vspace{-14cm}
\includegraphics[width=1.\textwidth,trim=18 318 1 26,clip]{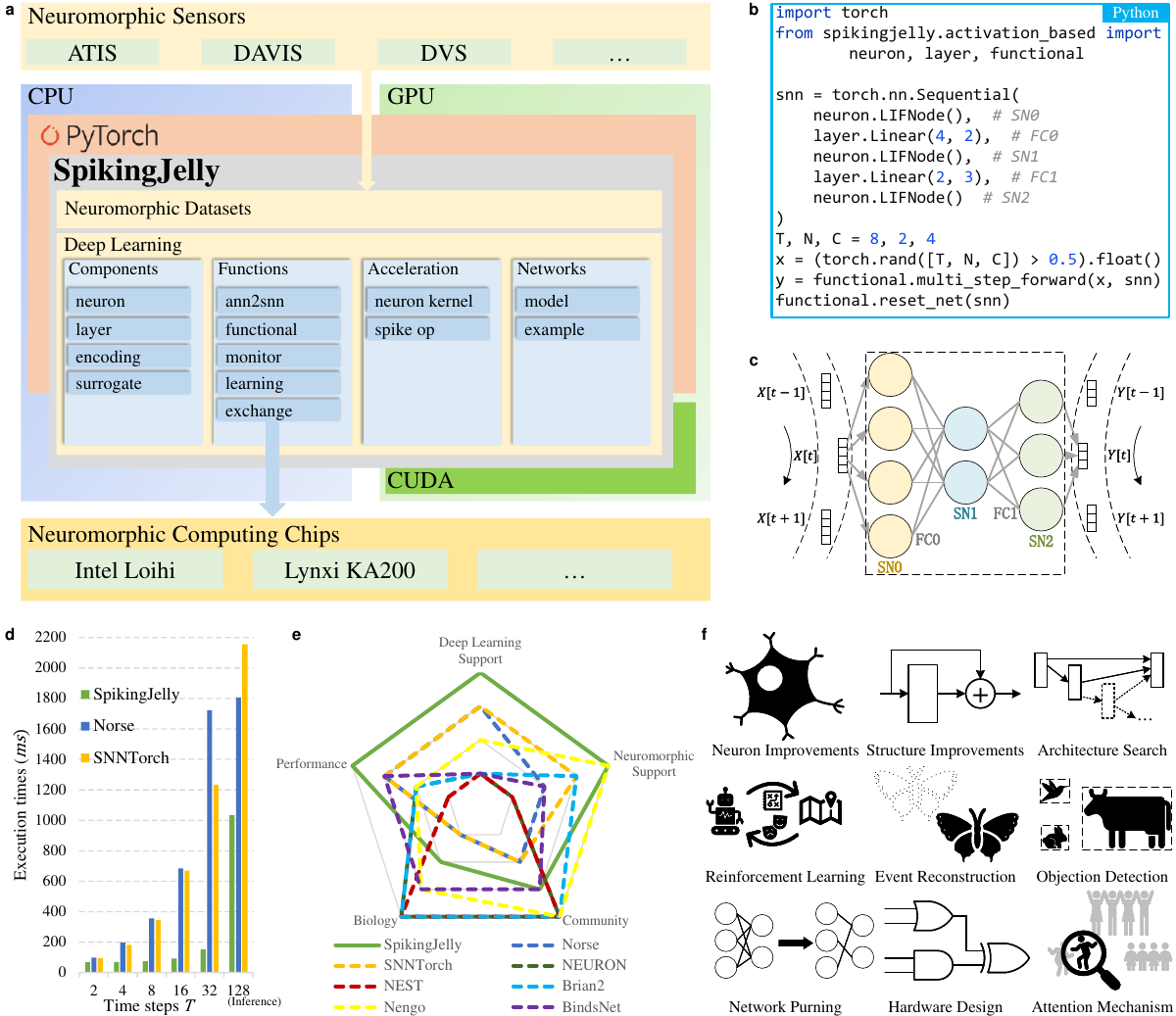}
\caption{\textbf{Overview of SpikingJelly.} 
	\textbf{a}. Architecture of the whole framework.  
	\textbf{b}. A code example of building and running an SNN, whose architecture is shown in \textbf{c}. 
	\textbf{d}. The execution times required for a single training iteration with $T=2,4,8,16,32$ and inference with $T=128$ using Spiking ResNet-18 built by SpikingJelly, Norse and SNNTorch. 
	\textbf{e}. Comparisons of the ecological niche of SpikingJelly with those of other frameworks.
	\textbf{f}. Typical adoptions from research based on SpikingJelly.}
	\label{figure: framework}
\end{figure}

% 4. introduction of spikingjelly (fig1)
\textbf{SpikingJelly: A Modern Framework for Spiking Deep Learning} To solve the above issues and promote research on spiking deep learning, we present SpikingJelly, an open-source deep learning framework, to bridge deep learning and SNNs. 
Fig.~\ref{figure: framework}\textbf{a} shows a hierarchical overview of the architecture of SpikingJelly. Based on one of the most commonly used machine learning frameworks, PyTorch, SpikingJelly supports the simulation of SNNs on both CPUs and GPUs with autograd-enabled computation. 
Additional CUDA kernels are employed for GPU-level acceleration beyond that provided by PyTorch. 
To achieve a balance between ease of use, flexible extensibility and high performance, the deep learning aspects in SpikingJelly are elaborated into four sections: \textit{Components}, \textit{Functions}, \textit{Acceleration}, and \textit{Networks} (see \textit{Subpackages of Deep Learning} in the supplementary materials). \textit{Components} provide essential modules such as spiking neurons and synapses to build deep SNNs. \textit{Functions} contain practical functions for training, simulating, analyzing, converting, quantizing and deploying SNNs. Some modules in \textit{Functions} are the functional formulations of the corresponding modules in \textit{Components}. In this way, both object-oriented and procedural programming are supported to satisfy the diverse demands of users.
\textit{Acceleration} accelerates the SNN simulation process with extra semiautomatically generated CUDA kernels, thus exploiting the efficiency of low-level programming languages and reducing the development cost via the code generation technique.
Based on the above subpackages, \textit{Networks} provide classic and large-scale network structures such as Spiking ResNet for fast model reuse as well as primary SNN applications for beginners, which is favourably received by the community.
Considering that neuromorphic datasets obtained from neuromorphic sensors \cite{lichtsteiner2008128, posch2010qvga, brandli2014240} such as ATIS, DAVIS, and DVS are widely used in SNNs, SpikingJelly integrates neuromorphic dataset processing methods, including downloading, unifying data layouts and reading interfaces with the general NumPy \cite{harris2020array} format. With the \textit{quantize} and \textit{exchange} packages provided in \textit{Functions}, compatibility with neuromorphic chips is also effectively implemented by providing low-bit quantizers for network weights and exchange functions for deploying SNNs.

As a full-stack framework, SpikingJelly enables researchers to build SNNs with flexible and convenient APIs, simulate SNNs with extremely high efficiency, and deploy SNNs to edge AI devices. With SpikingJelly, a method for synthesizing a truly spike-based energy-efficient machine intelligence paradigm enriches the ecology of the research in this field.

\section*{Results}
\subsection*{Convenient and Flexible SNN Construction}

SpikingJelly provides convenient and flexible APIs for users to build SNNs with a few steps. These APIs are designed in the acclaimed PyTorch style. Thus, those who are proficient in deep learning can also quickly start working with SpikingJelly. Fig.~\ref{figure: framework}\textbf{b} is an example of building and running an SNN, whose structure is shown in Fig.~\ref{figure: framework}\textbf{c}. Note that we input a 3-D tensor $X$ with a shape of $(T, N, C)$, where $T$ is the number of time steps, $N$ is the batch size, and $C$ is the number of features. Then, we use the \textit{multi\_step\_forward} function to input $X$ into the SNN, which implements a for-loop over time steps to send the inputs $X[t]$ to the network and concatenate the outputs $Y[t]$ to $Y$. In SpikingJelly, all hidden states are stored inside modules, which makes it convenient to build a network with sequential layers and access attributes from a specific module. Note that the hidden states need to be reset before sending a new sample, and we call the \textit{reset\_net} function in this example to reset the SNN.

%Modules in SpikingJelly are defined in multilevel inheritance with generic functions implemented by the superclass and specific functions implemented by the subclass, which provides sufficient augmentability for define the new model with marginal codes completion. Fig.~\ref{figure: framework}\textbf{c} shows how to implement the Exponential Integrate-and-Fire (EIF) spiking neuron in SpikingJelly. The discrete-time neuronal dynamic of the EIF neuron is defined by
%
%\begin{equation}
%	V[t] = V[t-1] + \frac{1}{\tau}(X[t] - (V[t-1] - V_{rest}) + \Delta_{T}\exp{(\frac{V[t-1]-\theta_{rh}}{\Delta_{T}})}),
%\end{equation}
%where $\tau$ is the membrane time constant, $\Delta_{T}$ is the sharpness parameter, $\theta_{rh}$ is the rheobase threshold, $V_{rest}$ is the rest membrane potential and $V[t]$ is the membrane potential at time step $T$. After we inherit the base neuron, we only need to add the parameters in the initialization function and implement the neuronal dynamic in the \textit{neuronal\_charge} function.

\subsection*{High-Performance Simulation}
SpikingJelly leverages the massively parallel computing power of GPUs to achieve extremely high SNN simulation performance.
For quantitative assessment purposes, we evaluated the training and inference performance of modern spiking deep learning frameworks on the Spiking ResNet, which is the spiking version of the classic ResNet ANN structure\cite{he2016deep} and is used frequently by SNN researchers\cite{10.3389/fnins.2019.00095, han2020rmp, pmlr-v139-li21d, hu2020spiking, SEWResNet}. The compared frameworks were Norse\cite{norse2021} and SNNTorch\cite{eshraghian2021training}, which were proposed at almost the same time as SpikingJelly or afterward. The versions of Norse and SNNTorch are 1.0.0 and 0.6.0, respectively, which are the latest versions. We used leaky-integrate-and-fire (LIF) neurons \cite{gerstner2014neuronal} in the network, as they are among the most commonly used spiking neurons in deep SNNs. The experiments are performed on an Ubuntu 18.04 server with an Intel Xeon Silver 4210R CPU, an NVIDIA A100-SXM-80GB GPU, and 256 GB of memory.

Researchers usually simulate SNNs in two ways. The first approach is to train SNNs directly via the surrogate gradient method with backpropagation through time (BPTT). Limited by the fact that the memory consumption is approximately proportional to the number of time steps $T$, direct training often uses few time steps; e.g., $T \leq 32$ for high-resolution datasets and $T \leq 64$ for small-size datasets. The second approach is to run SNNs with weights obtained via ANN2SNN. The ANN2SNN evaluation is restricted to inference, since this procedure does not require training. Most of the ANN2SNN methods are based on rate encoding and need far more time steps, e.g., $T \geq 128$, than the surrogate gradient method. Conversion methods based on latency encoding rather than rate coding are also reported \cite{lew2022time}. Note that recent research has achieved a substantial reduction of $T$, such as surrogate learning methods \cite{9556508} with $T = 5$ and ANN2SNN methods \cite{ding2021optimal, deng2021optimal} with $T \leq 64$. 
Our evaluation involved two steps for the most common usage: 1) training with small time steps for surrogate learning and 2) performing inference with large time steps for ANN2SNN.

First, we trained SNNs with $T = 2, 4, 8, 16, 32$ using the surrogate gradient method. Fig.~\ref{figure: framework}\textbf{d} shows the execution times required by the three frameworks for a single training iteration on Spiking ResNet-18. It can be found that SpikingJelly has a notable training speed advantage over other frameworks, yielding higher speedup when $T$ is larger, e.g., up to $11\times$ when $T=32$. Second, we tested SNNs in terms of inference with time steps of $T = 128, 256, 512, 1024$ to simulate the usage of ANN2SNN. As the experimental results suggest, the inference times of the three frameworks are all proportional to the number of time steps $T$. We therefore report only the results obtained with $T=128$ for simplicity. As Fig.~\ref{figure: framework}\textbf{d} shows, SpikingJelly also achieves up to 2$\times$ speedup over other frameworks. The source codes and results of the experiments are provided in the supplementary materials.

\subsection*{Neuromorphic Device Support}

Neuromorphic devices, including sensors and computing chips, are the key components of neuromorphic engineering and are also frequently used in spiking deep learning. SpikingJelly provides an efficient interface to incorporate SNNs with neuromorphic devices.

Most neuromorphic datasets are collected from neuromorphic sensors (e.g., N-MNIST \cite{10.3389/fnins.2015.00437} was collected from the ATIS sensor, and DVS Gesture was collected from the DVS128 camera), while others, such as ES-ImageNet\cite{10.3389/fnins.2021.726582}, are converted from static images by software algorithms. Raw data collected through the sensor is stored in a specific format, e.g., AEDAT, which requires a specific binary decoding method. To reduce the cost of use, decoding methods for different formats are implemented in SpikingJelly. SpikingJelly provides both event-based and downsampled frame-based representations of datasets. The event representation is identical to the address event representation (AER) format, which is also the default neuromorphic interchip communication protocol. The $i$-th event is $\textbf{E}[i] = (x_{i}, y_{i}, t_{i}, p_{i})$, where $(x_{i}, y_{i})$ is the coordinate, $t_{i}$ is the timestamp, and $p_{i}$ is the polarity. Frame representations are widely used \cite{SNN-IIR, STBP, neunorm, DECOLLE, fang2021incorporating, SEWResNet} and are usually temporally downsampled from event representations. The frame $\textbf{F}$ is a 4-D tensor containing event counts with a shape of $(T, C, H, W)$, where $T$ is the number of frames, $C$ is the number of channels and $H$ and $W$ are the height and width of the frame, respectively. Prevalent event-to-frame downsampling methods are also provided by SpikingJelly. 

The advantages of SNNs largely depend on the event-driven fashion, which can only be satisfied when neuromorphic computing chips are employed. To achieve this goal, researchers first train SNNs on powerful CPUs/GPUs to obtain their optimized weights and then deploy the pretrained SNNs on neuromorphic computing chips for inference purposes. Such a pipeline is also supported by SpikingJelly. 
SpikingJelly provides the necessary conversion functionality to convert native SpikingJelly modules to the supported formats of neuromorphic computing chips. This allows native SpikingJelly SNNs to run on neuromorphic chips (see \textit{Exchange Modules} in the supplementary materials). Currently, two of the most commonly used neuromorphic computing chips are supported by SpikingJelly: Loihi from Intel and Lynxi KA200 from the Tianjic \cite{pei2019towards} group. Deployment on other neuromorphic computing chips can also be implemented by developing specific exchange modules.

\subsection*{Ecological Niche}
A wide variety of SNN frameworks with distinctive features and specific advantages for certain usages are available, which we denote as ecological niches of frameworks. To illustrate the unique ecological niche of SpikingJelly among others, we summarize the characteristics of commonly used SNN frameworks. In general, frameworks can be divided into three categories.

The first category contains classic biological frameworks including NEURON, NEST, and Brian2, which use the lowest-level abstractions of biological neuron models and integrate (or not integrate but easy to implement) biologically plausible learning rules such as STDP. To support GPUs, some classic frameworks provide subframeworks, e.g., CoreNEURON for NEURON and Brian2GENN for Brian2, to accelerate parts of the modules derived from the parent framework. Brian2 also provides Brian2Loihi to support the Loihi chip. With rapid development in recent decades, the classical frameworks have been widely employed by neuroscientists and have grown into a large and active research community.

The second category, which can be considered as the intersection of neuroscience and computer science, includes Nengo and BindsNet. These frameworks are designed in the NumPy style and use neuron models of moderate complexity, which leads to a lower computational cost than classical frameworks. These frameworks are more compatible with GPUs since they provide GPU-supported backends such as OpenCL or are directly implemented based on modern machine learning frameworks that fully support GPUs. More specifically, Nengo employs OpenCL-based NengoOCL or TensorFlow-based NengoDL to use GPUs, and BindsNet is based on PyTorch. In addition, Nengo supports Loihi, which is implemented by the NengoLoihi subframework. Biologically plausible rules are the main training algorithms in these frameworks. It is worth noting that Nengo also supports ANN2SNN via the NengoDL subframework. The rich ecosystem of Nengo makes it one of the most commonly used frameworks. In addition, the rising BindsNet has attracted thousands of followers within the GitHub community.

The third category considers the intersection of neuroscience and deep learning, which includes Norse, SNNTorch, and SpikingJelly. All of these frameworks are based on PyTorch and support GPUs as well as automatic differentiation. High-level neuron model abstraction is employed, allowing these frameworks to work easily with backpropagation. All these frameworks support at least one deep learning method. These frameworks have received a lot of attention from the research community due to the increasing interest in spiking deep learning in recent years. Compared to other frameworks, SpikingJelly has the advantage of full-stack integration, which supports neuromorphic datasets and chips, ANN2SNN and surrogate gradients, as well as biologically plausible learning rules, and maximizes its simulation efficiency with specific optimization techniques for spike-based operations.

For a clear illustration purpose, we compare available SNN frameworks for SNNs in terms of five aspects: simulation performance (\textit{Performance}), support for neuromorphic sensors and computing chips (\textit{Neuromorphic Support}), community scale(\textit{Community}), biological abstraction level (\textit{Biology}), and support for surrogate learning and ANN2SNN (\textit{Deep Learning Support}). The results are presented in Fig.~\ref{figure: framework}\textbf{e}, which clearly shows the ecological niche of SpikingJelly. For more details about the difference with frameworks, please refer to \textit{Comparison of SNN Frameworks} in the supplementary materials.

\subsection*{Adoptions by the Community}
The increasing number of adoptions by the community is symbolic of the success of a framework.
Since being open-sourced in December 2019, SpikingJelly has been widely used in many spiking deep learning studies, including adversarial attack\cite{abad2022poster, abad2023sneaky}, ANN2SNN  \cite{ding2021optimal, bu2021optimal, bu2022optimized, tang2022snn2ann, hao2023reducing, hao2023bridging}, attention mechanisms\cite{zhu2022tcja, 10032591}, depth estimation from DVS data\cite{ranccon2021stereospike, wu2022mss}, development of innovative materials  \cite{D2NR06498G}, emotion recognition \cite{wang2022spiking}, energy estimation \cite{lemaire2022analytical}, event-based video reconstruction\cite{zhu2022event}, fault diagnosis\cite{9994622}, hardware design\cite{li2021feas, kaiser2023neuromorphic, 10.1007/978-3-031-14903-0_5}, network structure improvements\cite{SEWResNet, han2021cascade, vicente2021keys, li2022spikeformer, yu2022stsc, zhou2023spikformer}, spiking neuron improvements\cite{fang2021incorporating, jin2022sit, tang2022relaxation, wu4179879dynamic, WANG2023109102, hammouamri2022mitigating, yao2022glif}, training method improvements \cite{kheradpisheh2022spiking, meng2022training, 10005101, brainsci13020168, ZHAN2023110193, lucas2022entrenamiento, chen2023training, duan2022temporal, yang2022training, xiao2022online, zheng2022label}, medical diagnosis\cite{feng2022building, electronics11121889}, network pruning\cite{ijcai2021-236, liu2022dynsnn, chen2022state, kim2022lottery, chen2023a}, neural architecture search\cite{na2022autosnn, kim2022neural}, neuromorphic data augmentation\cite{li2022neuromorphic}, natural language processing\cite{zhu2023spikegpt}, object detection/tracking for DVS/frame data\cite{cordone2022object, barchid2022spiking, xiang2022spiking}, odor recognition\cite{xiong2021odor}, optical flow estimation with DVS data\cite{cuadrado2023optical}, reinforcement learning for controlling\cite{liu2021human, chen2022deep, qin2022low}, and semantic communication\cite{arxiv.2210.06836}. Fig.~\ref{figure: framework}\textbf{f} shows parts of these adoptions. Its widespread adoption by the community marks SpikingJelly as one of the most commonly used spiking deep learning frameworks.

\begin{figure}
	\centering
	%	\subfloat[]{\includegraphics[width=0.49\textwidth,trim=0 360 680 0,clip]{./fig/neuron.pdf}}
	%	\subfloat[]{\includegraphics[width=0.49\textwidth,trim=0 0 0 0,clip]{./fig/lif.pdf}}
	%	
	%	
	%	\vspace{-0.4cm}
	%	\subfloat[]{\includegraphics[width=0.4\textwidth,trim=20 630 280 20,clip]{./fig/uml_class.pdf}}
	%	\subfloat[]{\includegraphics[width=0.35\textwidth,trim=0 0 0 0,clip]{./fig/surrogate_function.pdf}}
	%	\subfloat[]{\includegraphics[width=0.25\textwidth,trim=20 260 670 20,clip]{./fig/surrogate.pdf}}
	%	
	%	\subfloat[]{\includegraphics[width=0.3\textwidth,trim=140 620 270 48,clip]{./fig/latency_encoder.pdf}}
	%	\subfloat[]{\includegraphics[width=0.3\textwidth,trim=30 10 0 26,clip]{./fig/events_example3.pdf}}
	%	\begin{minipage}[t]{0.4\linewidth}
		%		\centering
		%		\subfloat[]{\includegraphics[width=1.\textwidth,trim=10 680 520 10,clip]{./fig/events_frames3.pdf}}
		%		
		%		\subfloat[]{\includegraphics[width=1.\textwidth,trim=170 760 100 30,clip]{./fig/dataset_class.pdf}}
		%	\end{minipage}

	\includegraphics[width=0.94\textwidth,trim=20 262 556 20,clip]{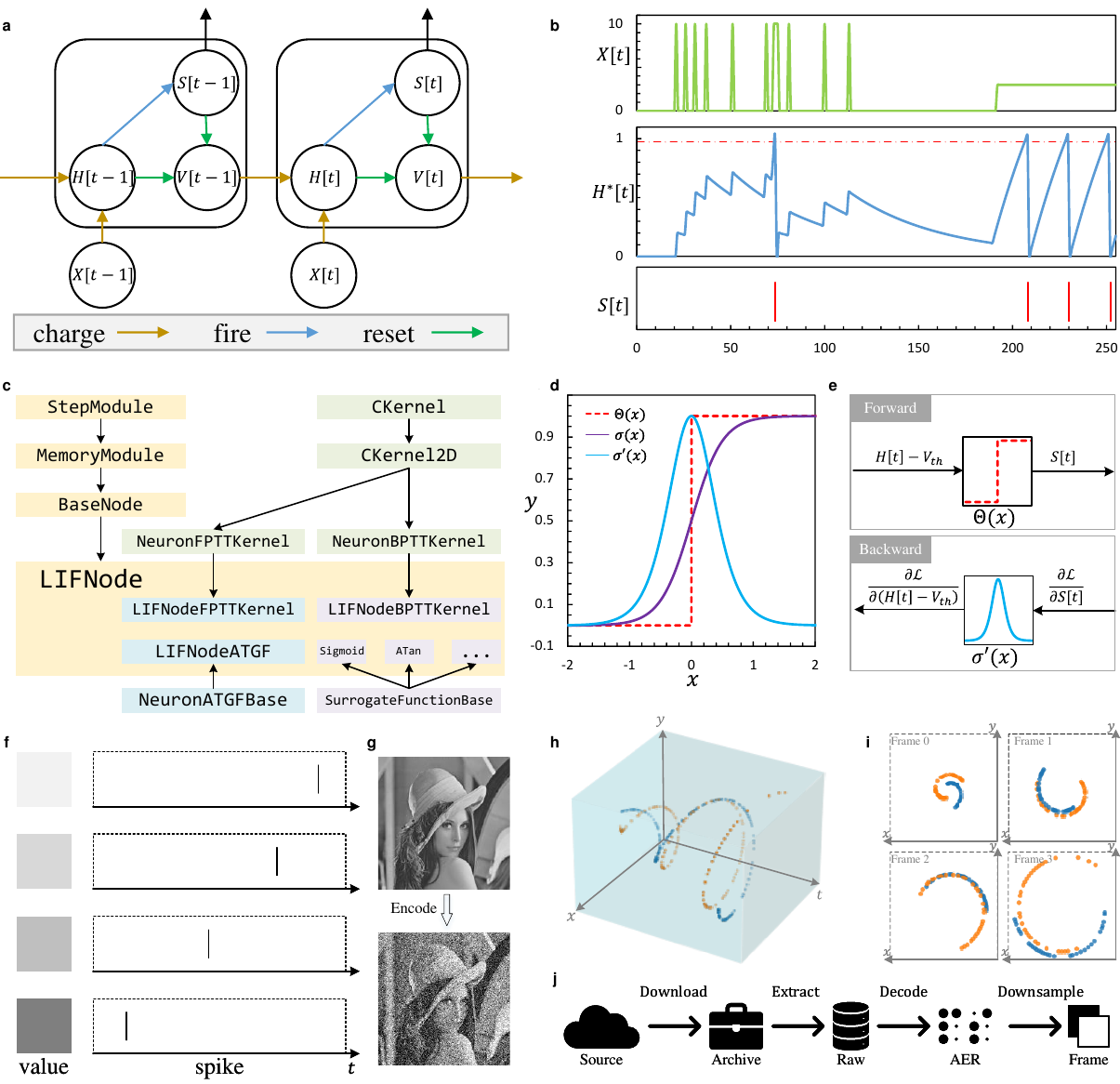}
	\caption{\textbf{Component modules contained in SpikingJelly.} 
		\textbf{a}. The general discrete-time spiking neuron model described by (Eqs.~(\ref{eq discrete neuronal charge}, \ref{eq discrete neuronal fire}, \ref{eq discrete neuronal reset})): the neuronal charge, fire, and reset equations. 
		\textbf{b}. Simulation of an LIF neuron. The input $X[t]$ is 0 or 10 when $0 \leq t <128$, 0 when $128 \leq t <196$, and 3 when $196 \leq t < 256$. Here $\{H^{*}[t]\}$ records $H[t]$ at all time steps and includes $V[t]$ when $S[t]=1$. 
		\textbf{c}. The inheritance relationships of an LIF neuron.
		\textbf{d}. An example of the sigmoid surrogate function $\sigma(x) = \frac{1}{1 + \exp (-\alpha x)}$, which is an approximator of the Heaviside function $\Theta(x)$. 
		\textbf{e}. The principle of the surrogate gradient method that uses $\Theta(x)$ during forward propagation to generate discrete binary spikes, while using $\sigma'(x)$ during backward propagation to obtain continuous float gradients. 
		\textbf{f}. The latency encoder that encodes larger values, which are shown by darker squares, for spikes with earlier firing times.
		\textbf{g}. Illustration of the input image and the output spikes of the Poisson encoder.
		\textbf{h}. Visualization of synthetic events with two polarities and \textbf{i}. the frames downsampled from these events.
		\textbf{j}. The workflow of dataset processing.
	}
	\label{figure: components}
\end{figure}

\begin{figure}
	\centering
	
	\subfloat[]{\includegraphics[width=0.9\textwidth,trim=0 240 90 0,clip]{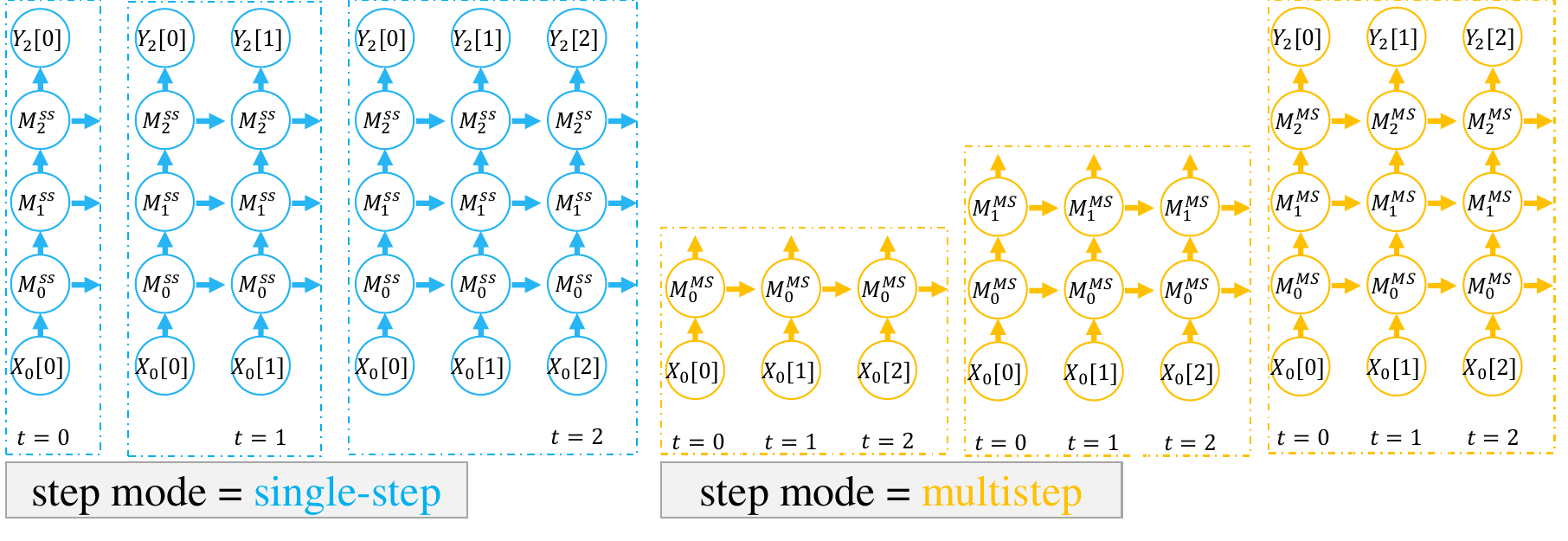}}
	%\subfloat[]{\includegraphics[width=0.6\textwidth,trim=12 260 580 64,clip]{./fig/forward2.pdf}}
	\vspace{-0.3cm}
	
	\subfloat[]{\includegraphics[width=0.3915\textwidth,trim=140 730 340 50,clip]{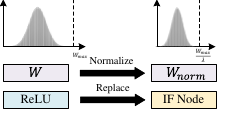}}
	\subfloat[]{\includegraphics[width=0.2115\textwidth,trim=0 0 0 0,clip]{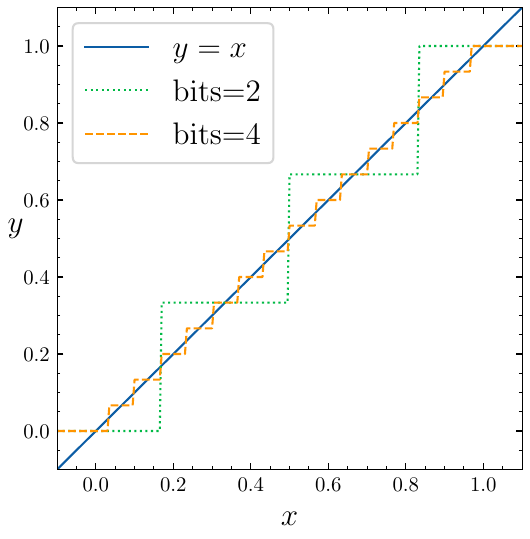}}
	\subfloat[]{\includegraphics[width=0.297\textwidth,trim=30 670 370 34,clip]{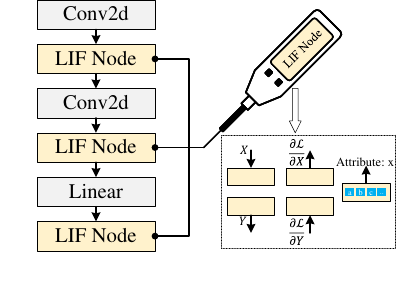}}
	\vspace{-0.8cm}
	
	\subfloat[]{\includegraphics[width=0.315\textwidth,trim=50 630 380 48,clip]{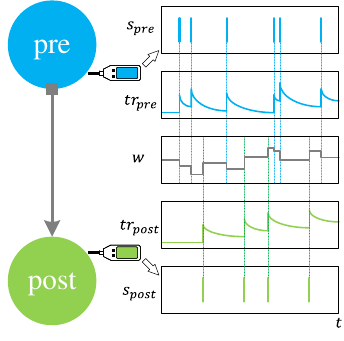}}
	\subfloat[]{\includegraphics[width=0.315\textwidth,trim=20 8 2 40,clip]{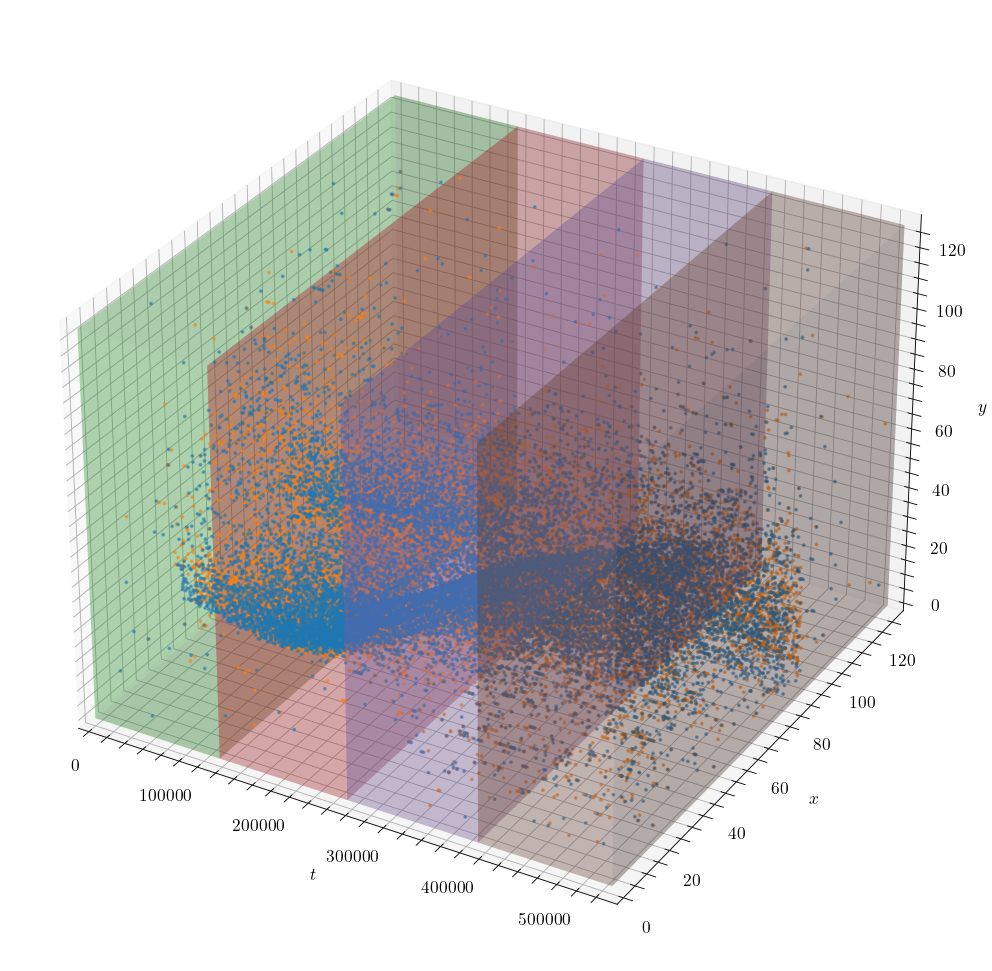}}
	\subfloat[]{\includegraphics[width=0.315\textwidth,trim=20 8 2 40,clip]{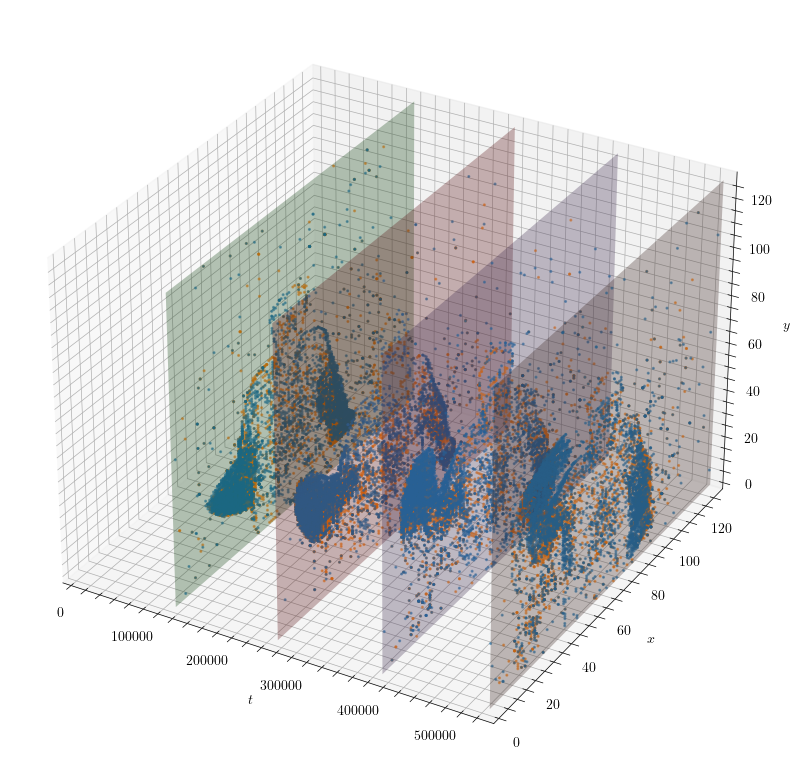}}

\caption{\textbf{Functional modules in SpikingJelly.}
	\textbf{a}. Schematic plot demonstrating the use of a step-by-step or layer-by-layer propagation pattern to simulate a network with 3 layers and 3 time steps. $X_{i}[t]$ is the input of the $i$-th layer at time step $t$, and $Y_{i}[t]$ is the output of the $i$-th layer at time step $t$. $M^{ss}_{i}$ denotes the $i$-th layer in the single-step mode, while $M^{ms}_{i}$ denotes the $i$-th layer in the multistep mode. The complete computational graphs of the two propagation patterns are identical but constructed in different orders.
	\textbf{b}. Functions of ANN2SNN that normalize weights and replace the rectified linear unit (ReLU) with an integrate-and-fire (IF) neuron layer.
	\textbf{c}. The quantizer maps the input in the range $[0, 1]$ to the nearest fixed-point, the number of which is controlled by the bits.
	\textbf{d}. The monitor works like a probe, recording the input, output, input gradients, output gradients, and attributes of the assigned module.
	\textbf{e}. The STDP learner uses a monitor to record presynaptic spikes and presynaptic spikes, computes traces, and updates weights.
	\textbf{f}. Diagram showing the process of slicing events from the DVS Gesture dataset to 4 bins and integrating them to 4 frames (shown in \textbf{g}).
	}
	\label{figure: functional module}
\end{figure}

\subsection*{Core Modules of SpikingJelly}
The key factor that enables SpikingJelly to achieve a combination of flexibility, efficiency, and completeness is the design philosophy of its modules. Some of the core components and functional modules are shown in Fig.~\ref{figure: components} and Fig.~\ref{figure: functional module}, respectively.

Figs.~\ref{figure: components}\textbf{a-e} illustrate a spiking neuron and its components in SpikingJelly. As Fig.~\ref{figure: components}\textbf{a} shows, SpikingJelly uses three discrete-time equations, which are charge, fire, and reset equations, to describe the behaviors of spiking neurons. $X[t], H[t], S[t]$, and $V[t]$ are the input, membrane potential after charging, spikes, and membrane potential after resetting at time step $t$, respectively. Fig.~\ref{figure: components}\textbf{b} shows the response of the LIF neuron in SpikingJelly for a given stimulus.
Fig.~\ref{figure: components}\textbf{c} shows the inheritance of the LIF neuron class. By inheriting the parent class, the LIF neuron can be easily implemented with only a few lines of code.
Fig.~\ref{figure: components}\textbf{d} shows the Heaviside function $\Theta(x)$, sigmoid surrogate function $\sigma(x) = \frac{1}{1 + \exp (-\alpha x)}$ (where $\alpha$ is a hyperparameter for controlling the shape), and surrogate gradient $\sigma'(x)$, which are key attributes of the spiking neuron shown in Fig.~\ref{figure: components}\textbf{c}. Fig.~\ref{figure: components}\textbf{e} illustrates how surrogate learning is implemented in SpikingJelly. $\Theta(x)$ is applied during the forward propagation to generate spikes, and $\sigma'(x)$ is applied during the backward propagation to calculate gradients.
Figs.~\ref{figure: components}\textbf{f}, \textbf{g} visualize two typical spiking encoders in SpikingJelly. The latency encoder shown in Fig.~\ref{figure: components}\textbf{f} encodes larger values (darker color) to earlier spikes. The Poisson encoder shown in Fig.~\ref{figure: components}\textbf{g} encodes the input value to spikes with a firing probability obeying the Poisson distribution.
Fig.~\ref{figure: components}\textbf{h} shows synthetic events with two polarities, and the frames downsampled by SpikingJelly are shown in Fig.~\ref{figure: components}\textbf{i}. The workflow used to process datasets in SpikingJelly is shown in Fig.~\ref{figure: components}\textbf{j}.

The modules in SpikingJelly have the \textit{step\_mode} attribute to decide whether to apply single-step or multistep forwarding. A module in the single-step mode receives $X[t]$ and outputs $Y[t]$, which are data at a single time step. In contrast, a module in the multistep mode receives $X = \{X[0], X[1], ..., X[T-1]\}$ and outputs $Y = \{Y[0], Y[1], ..., Y[T-1]\}$, which are sequences with data at many time steps.
Correspondingly, the SNN with all modules in a single-step or multistep mode follows the step-by-step or layer-by-layer propagation pattern, respectively, which is illustrated in Fig.~\ref{figure: functional module}\textbf{a}. The difference between these propagation patterns is the order in which the computational graphs are constructed. More specifically, the step-by-step propagation pattern is a depth-first-search (DFS), while the layer-by-layer pattern is a breadth-first-search (BFS). The propagation patterns in SpikingJelly are designed to satisfy specific user intentions (see \textit{Distinction of Propagation Patterns} in the supplementary materials). The step-by-step approach is more flexible for building recurrent connections and is suitable for ANN2SNN since its memory consumption is not proportional to the number of time steps $T$ during inference. The layer-by-layer pattern has the advantage of efficiency optimized by merging time steps into batch dimensions for stateless modules and fusing kernels for stateful modules in SpikingJelly.
Fig.~\ref{figure: functional module}\textbf{b} shows the ANN2SNN conversion functions, which normalize the weights and replace the rectified linear units (ReLUs) with integrate-and-fire (IF) neuron layers.
Fig.~\ref{figure: functional module}\textbf{c} shows the $k$-bit quantizer with $k=2, 4$, which quantizes the input $x\in(0, 1)$ to the nearest fixed-point value $y$. Quantizers support quantization-aware training, which quantizes weights during training. Note that the gradient of the quantizer is zero almost everywhere, hence the surrogate gradient method is employed.
Fig.~\ref{figure: functional module}\textbf{d} shows an example of using the monitor to record data. The monitor works like a probe, collecting data from prescribed modules. Inputs, outputs, input gradients, output gradients, and attributes can be recorded to meet the main data collection requirements.
Fig.~\ref{figure: functional module}\textbf{e} illustrates the STDP learner based on the trace method \cite{morrison2008phenomenological}, which uses two monitors to record pre/postsynapse spikes. Then, the traces are updated, and the weights of the synapses are modified.
Figs.~\ref{figure: functional module}\textbf{f}, \textbf{g} visualize the event slicing and downsampling operations. In Fig.~\ref{figure: functional module}\textbf{f}, a sample from the DVS Gesture dataset is sliced into four bins, which are shown in four cuboids with different colors. Then the events in each cuboid are accumulated, and four frames are generated in Fig.~\ref{figure: functional module}\textbf{g}. More details of modules in SpikingJelly can be found in the \textit{Materials and Methods} section.

\begin{figure}
	\centering
	\includegraphics[width=1.\textwidth,trim=20 272 20 110,clip]{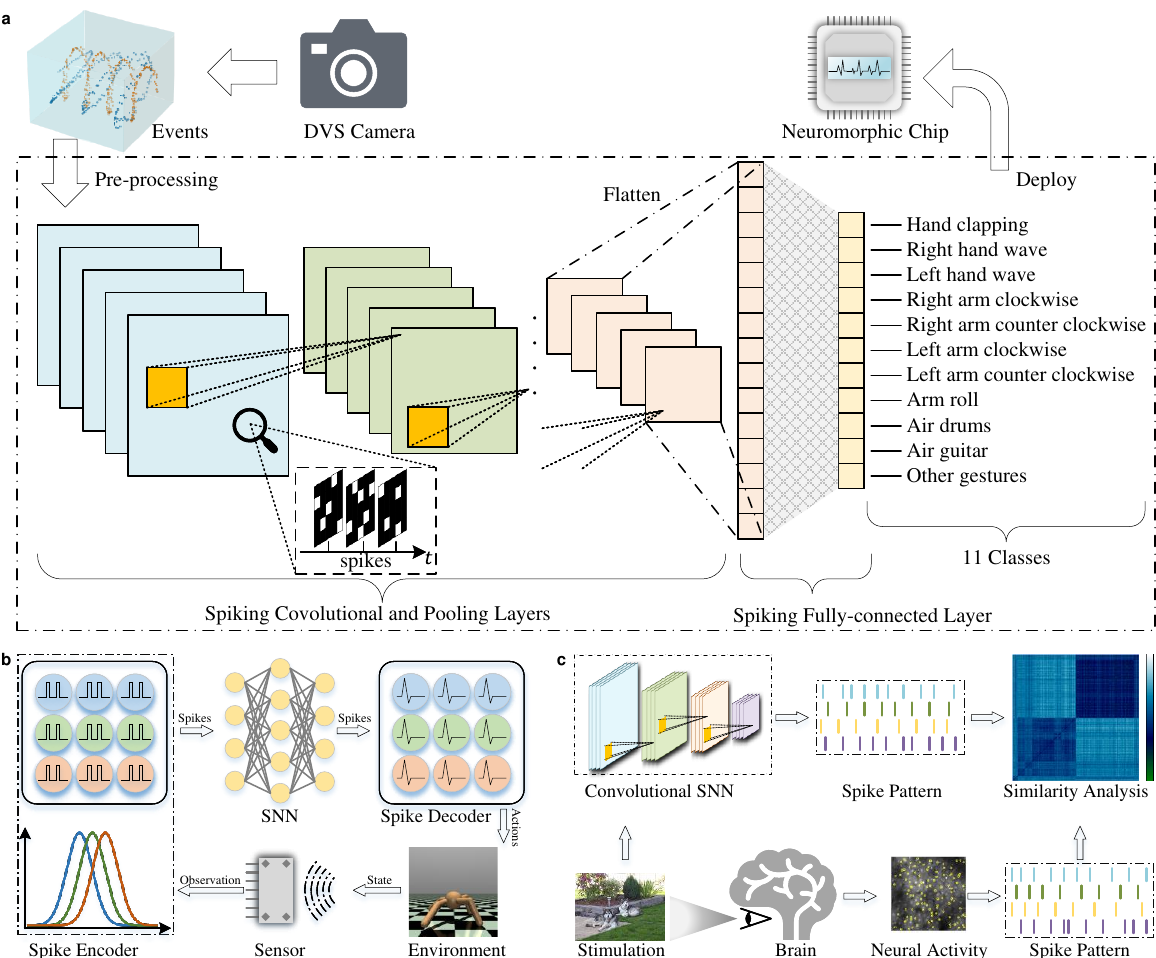}
	%	\subfloat[]{\includegraphics[width=1.\textwidth,trim=500 180 540 190,clip]{./fig/classify_dvsg_example.pdf}}
	%	
	%	\subfloat[]{\includegraphics[width=0.5\textwidth,trim=10 430 570 18,clip]{./fig/rl_example.pdf}}
	%	\subfloat[]{\includegraphics[width=0.5\textwidth,trim=570 590 110 610,clip]{./fig/neural_similarity.pdf}}
	
	\caption{\textbf{Typical applications of SpikingJelly.} \textbf{a}. The full-stack solution consists of preprocessing the raw events derived from the DVS Gesture dataset to tensors, building and training a deep SNN for classification, and deploying the SNN to the neuromorphic Loihi chip for inference. \textbf{b}. Pipeline for reinforcement learning and control. The spike encoder transforms the observation obtained from the state of the environment in the OpenAI Gym into spike trains, and then the spike trains are converted into continuous values in the spike decoder after SNN processing to represent the actions. \textbf{c}. Workflow of measuring neural similarity between the biovisual system and SNNs. The same visual stimuli are fed to the biovisual system and the pretrained spiking neural network to obtain neural representations of the two systems, and then the brain-likeness of the model is assessed by measuring the representational similarity.}
	\label{figure: cases}
\end{figure}

\subsection*{Typical Applications}
As a full-stack toolkit for spiking deep learning, SpikingJelly involves almost all SNN applications.
For simplicity, we demonstrate three typical scenarios that utilize SpikingJelly in Fig.~\ref{figure: cases}. 
The first case shown in Fig.~\ref{figure: cases}\textbf{a} is an example of training a deep SNN to classify the neuromorphic DVS Gesture dataset. First, raw events from the DVS camera are preprocessed to tensors via the \textit{dataset} subpackage in SpikingJelly. Then, we use the \textit{layer}, \textit{neuron}, and \textit{exchange} subpackages to build and train a deep convolutional SNN for classifying the DVS Gesture dataset. The spikes flow through the convolutional layers, as shown in Fig.~\ref{figure: cases}\textbf{a}. After training, we use the \textit{exchange} subpackage to deploy the SNN on the neuromorphic Loihi chip for inference purposes. The second case shown in Fig.~\ref{figure: cases}\textbf{b} is an example of using a deep SNN to solve the continuous control tasks from the OpenAI gym\cite{1606.01540}. First, the observation obtained from the state of the environment is encoded by different Gaussian receptive fields composed of population neurons \cite{BOHTE200217}, which effectively convert float values to spike trains. After processing, the discrete spike trains are converted to float values, which are in fact the membrane potentials of subsequent layers of non-spiking neurons, to represent the action. Each module in the network can be implemented by the \textit{layer} and \textit{neuron} subpackages. The third case shown in Fig.  \ref{figure: cases}\textbf{c} is an example of modeling the biological visual cortex by the deep SNN. Firstly, we build and train the deep SNN on the ImageNet classification task by the \textit{layer} and \textit{neuron} subpackages. Then, to obtain the neural representations of the visual cortex and the model, the same visual stimuli are fed to the biological visual system and the pretrained SNNs. Finally, the performance of SNNs in modeling the visual cortex can be quantitatively analyzed by calculating the resulting representation similarity. Source codes and experimental results for these applications are provided in the supplementary materials.

\section*{Discussion}

Although SNNs outperform ANNs in terms of biological plausibility and power efficiency, their applications were restricted to neuroscience rather than computational science due to the lack of available learning methods. With the introduction of deep learning methods, the performance of SNNs has been greatly improved, making spiking deep learning a new research hot spot. However, this emerging research area faces the dilemma that the classical software frameworks focus on neuroscience rather than deep learning, while new frameworks have not been developed. SpikingJelly is designed to satisfy the booming research interests of spiking deep learning (see \textit{Statistical Trends} in the supplementary materials).

% convenient APIs
Spiking deep learning is an emerging interdisciplinary field where researchers should be well versed in both neuroscience and deep learning. However, researchers who specialize in one research area may not be familiar with the other, which is consistent with our developers' experience when answering questions and participating in discussions posed by users in GitHub. To mitigate the learning and the cost of use, SpikingJelly provides brief and convenient APIs. Classic models and frequently-used training scripts are also included. With a few lines of code, users can easily build various types of SNNs and run their models even if they are not seasoned developers aware of the underlying implementation. This design philosophy of SpikingJelly frees users from painstaking coding operations when implementing more creative work. 

Complex and diverse spiking neurons and synapses are the core components of SNNs. Modifying neurons and synapses by mimicking the biological neural system \cite{SNN-IIR, fang2021incorporating, ijcai2021-236}  or referring experience from deep ANNs \cite{hu2020spiking, SEWResNet} is a practicable method for improving spiking deep learning. Researchers want to modify existing modules to define new classes of spiking modules by changing certain functions and properties. Researchers also expect that they should only need to write a few codes to make large changes to the behavior and performance of a model. Such a research paradigm is supported by SpikingJelly's flexible APIs. Most of the modules in SpikingJelly are created by inheriting from their parents, overriding functions, and adding/deleting new attributes, which also provides a perfect reference for researchers to define new modules.

Deep learning typically uses datasets with massive numbers of samples and large-scale models \cite{Goodfellow-et-al-2016}. A larger number of training epochs is also employed to achieve better performance \cite{https://doi.org/10.48550/arxiv.2110.00476}. All the above characteristics are computationally intensive and hold for spiking deep learning. Moreover, due to the additional temporal dimension, deep SNNs have a higher level of computational complexity than deep ANNs. Thus, the simulation efficiency of deep SNNs is critical, especially for the recent research progress in terms of evaluating deep SNNs with more than 50 layers on the ImageNet dataset containing 1.28 million images has been a widely used performance baseline \cite{10.3389/fnins.2019.00095, han2020rmp, pmlr-v139-li21d, hu2020spiking, SEWResNet}. Computational efficiency is emphasized in the design of SpikingJelly. The simulation process using SpikingJelly benefits from its infrastructure, PyTorch, which enables CPU acceleration with OpenMP/MKL and GPU acceleration with cuBLAS/CUDNN. Advanced acceleration with fusion operations is introduced by merging dimensions, semiautomatically generated CUDA kernels, and just-in-time (JIT) compiling, which brings extremely high training/inference efficiency. Equipped with these acceleration methods, SpikingJelly achieves state-of-the-art simulation speed, which frees researchers from wasting too much time waiting for lengthy deep SNN training processes.

% applications
Deep learning methods boost the performance of SNNs and make SNNs practical for use in real-life tasks \cite{tavanaei2019deep}. With the full-stack solution provided by SpikingJelly for building, training, and deploying SNNs, the boundaries of deep SNNs have been extended from toy dataset classification to applications with practical utility, including human-level performance classification, network deployment, and event data processing. Beyond the classic machine learning tasks, several frontier applications of deep SNNs have also been reported, including a spike-based neuromorphic perception system consisting of calibratable artificial sensory neurons \cite{yuan2022calibratable}, a neuromorphic computing model running on memristors \cite{doi:10.1063/5.0096643} and the design of an event-driven SNN hardware accelerator \cite{li2021feas}. All the above evidence indicates that the advent of SpikingJelly will accelerate the boom of the spiking deep learning community. The future development scheme of SpikingJelly will keep track of the advances in neuromorphic computing, and the rapid symbiotic growth of SpikingJelly and the community will be witnessed.

\section*{Materials and Methods}
In this section, we first introduce the key modules of SpikingJelly and their design concepts. For clarity of exposition, we categorize modules into two categories: component modules and functional modules. Component modules are essential components of SNNs and are frequently used to define networks. Functional modules are modules or functions that simulate an SNN, accelerate its simulation process, record specific data, or modify variables. Then, the acceleration methods in SpikingJelly are discussed. 

\subsection*{Component Modules}

\textbf{Neuron Model.} Spiking neurons are the key components of SNNs. Neuroscientists prefer to use spiking neurons with complex neuronal dynamics, e.g., the Hodgkin-Huxley model \cite{hodgkin1952quantitative} and the Izhikevich model \cite{izhikevich2003simple}, for high biological plausibility. When applying deep learning in SNNs, simplified spiking neuron models, including IF and LIF models, are more commonly used due to their straightforward neuronal dynamics and compact hyperparameters, which reduce the computational complexity and workload levels of researchers fine-tuning the training process of the network.

In general, the behavior of spiking neurons can be described by three processes: neuronal charging, neuronal firing, and neuronal resetting. Neuronal charging is also denoted as subthreshold dynamics, which can be defined as one or many ODEs. For example, the neuronal charging equation for an LIF neuron can be formulated as
\begin{equation}
	\tau_{m} \frac{\mathrm{d}V(t)}{\mathrm{d}t} = -(V(t) - V_{rest}) + X(t),
	\label{eq continuous LIF neuronal charge}
\end{equation}
where $\tau_{m}$ is the membrane time constant, $V(t)$ is the membrane potential at time $t$, $V_{rest}$ is the resting potential, and $X(t)$ is the input current. The spiking neuron fires a spike $S(t) = 1$ when its membrane potential $V(t)$ crosses the threshold $V_{th}$, and remains silent ($S(t) = 0$) if $V(t) < V_{th}$, which can be described as
\begin{equation}
	S(t) = \Theta(V(t) - V_{th}),
	\label{eq continuous neuronal fire}
\end{equation}
where $\Theta(x)$ is the Heaviside step function defined by $\Theta(x) = 1$ for $x \ge 0$ and $\Theta(x) = 0$ for $x < 0$. 
If the neuron fires a spike, it will be reset. There are two types of resets, hard reset and soft reset. The hard reset, which resets $V(t)$ to $V_{reset}$, is commonly used in directly trained SNNs due to its better experimental performance \cite{ledinauskas2020training}, while the soft reset, which subtracts $V_{th}$ if the neuron fires, is adopted in ANN2SNN due to its lower theoretical fitting error for ReLU activations\cite{Bodo2017Conversion}. The neuronal reset can be formulated as
\begin{equation}
	\lim_{\Delta t \rightarrow 0^{+}} V(t + \Delta t) = \begin{cases}
		V_{reset}, ~hard~reset\\
		V(t) - V_{th}, ~soft~reset
	\end{cases}. \label{eq continuous neuronal reset}
\end{equation}

To simulate a spiking neuron, the continuous-time differential equations can be approximated by discrete-time differential equations. In practice, the simplest one-order Eulerian method is used due to its low computational cost. For example, the discrete-time version of Eq.~(\ref{eq continuous LIF neuronal charge}) is 
\begin{equation}
	\tau_{m}(V[t] - V[t-1]) = -(V[t-1] - V_{rest}) + X[t].
	\label{eq discrete LIF neuronal charge}
\end{equation}

The neuronal charging equation is specific to different neurons, while the neuronal firing and neuronal resetting equations can be shared. Extending from our previous research\cite{fang2021incorporating}, SpikingJelly uses three discrete-time equations to describe spiking neurons:
\begin{align}
	H[t] &= f(V[t-1], X[t]), \label{eq discrete neuronal charge}\\
	S[t] &= \Theta(H[t] - V_{th}), \label{eq discrete neuronal fire}\\
	V[t] &= \begin{cases}
		H[t] \cdot (1 - S[t]) + V_{reset} \cdot S[t], ~hard~reset\\
		H[t] - V_{th} \cdot S[t], ~soft~reset
	\end{cases}. \label{eq discrete neuronal reset}
\end{align}
To avoid confusion, we use $H[t]$ to represent the membrane potential after charging but before firing, and $V[t]$ to represent the membrane potential after resetting. Eq.~(\ref{eq discrete neuronal charge}) is the neuronal charging equation, and $f$ is specific to a certain neuron. Eq.~(\ref{eq discrete neuronal fire}) is the neuronal firing equation, and Eq.~(\ref{eq discrete neuronal reset}) is the neuronal reset equation.
Fig.~\ref{figure: components}\textbf{a} shows the general discrete-time neuron model described by these three equations.

Fig.~\ref{figure: components}\textbf{b} shows the simulation of an LIF neuron built by SpikingJelly with $\tau_{m}=50, V_{th}=1$, and $V_{reset}=0$. The input $X[t]$ is 0 or 10 when $0 \leq t <128$, 0 when $128 \leq t <196$, and 3 when $196 \leq t < 256$. Here $\{H^{*}[t]\}$ not only records $H[t]$ at all time steps but also includes $V[t']$ for those time steps $t'$ when the neuron fires a spike ($S[t']=1$), which shows both fires and resets more clearly than only visualizing $H[t]$ or $V[t]$ alone.

Fig.~\ref{figure: components}\textbf{c} takes an LIF neuron as the example and illustrates how SpikingJelly simplifies the module building process via multilevel inheritance. The \textit{StepModule} is a module that can work in both single/multistep modes with a single tensor or a sequence input, and it is the base class of most modules in SpikingJelly (see \textit{Design of the Step Module} in the supplementary materials). Based on the \textit{StepModule}, the \textit{MemoryModule} adds hidden states as extra attributes, making it suitable as the base class for stateful modules, including neurons and synapses in SNNs. Then, the general discrete-time neuron \textit{BaseNode} can be easily designed by extending the \textit{MemoryModule} with neuronal firing and resetting operations obeying Eqs.~(\ref{eq discrete neuronal fire}), and (\ref{eq discrete neuronal reset}), respectively, and defining an empty neuronal charging function obeying Eq.~(\ref{eq discrete neuronal charge}) for child classes to complete. Finally, we build the \textit{LIFNode} inherited from the \textit{BaseNode} by completing the neuronal charging function Eq.~(\ref{eq discrete neuronal charge}) and utilizing an extra membrane potential parameter $\tau_{m}$. Beyond the naive PyTorch implementation, SpikingJelly offers additional optional CUDA kernels for accelerating spiking neurons. The forward CUDA kernel \textit{LIFNodeFPTTKernel}, the backward CUDA kernel \textit{LIFNodeBPTTKernel}, the autograd function \textit{LIFNodeATGF} and the surrogate function of the LIF neuron are also extended through multilevel inheritance with several lines of code changes. Such a module design paradigm enhances the extensibility of SpikingJelly and reduces the workload required for researchers to develop new models. When researchers want to define a new kind of module, they only need to inherit the corresponding base class and complete or override some functions.

\hspace*{\fill}

\noindent\textbf{Surrogate Gradients.} The great success of deep learning models is largely due to their multiple processing layers trained by backpropagation and gradient descent \cite{deep-learning-nature}. However, the derivative of the Heaviside function $\Theta(x)$ introduced by the spike trigger mechanism in Eq.~(\ref{eq discrete neuronal fire}) is $+\infty$ at $x=0$ and 0 at $x \neq 0$, which blocks the gradient propagation process. To apply the gradient-based deep learning method in SNNs, the surrogate gradient method has been proposed \cite{lee2016training, STBP, shrestha2018slayer, lee2020enabling}, which uses the derivative of the surrogate function $\sigma(x)$ to define the derivative of $\Theta(x)$ during backpropagation. Commonly used surrogate functions are approximators of $\Theta(x)$, which have similar but smoother shapes; an example is the sigmoid surrogate function $\sigma(x) = \frac{1}{1 + \exp (-\alpha x)}$ shown in Fig.~\ref{figure: components}\textbf{d}, where $\alpha$ is a hyperparameter used to control the shape.

As a surrogate function is used in the spike trigger mechanism, SpikingJelly packages it as a component of the spiking neuron. As shown in Fig.~\ref{figure: components}\textbf{e}, during forward propagation, the neuron uses $S[t] = \Theta(H[t] - V_{th})$ to generate discrete binary spikes $S[t] \in \{0, 1\}$. During backward propagation, the gradient is calculated by $\frac{\partial \mathcal{L}}{\partial (H[t] - V_{th})} = \frac{\partial \mathcal{L}}{\partial S[t]} \cdot \sigma'(H[t] - V_{th})$, where $\mathcal{L}$ is the loss.

\hspace*{\fill}

\noindent\textbf{Encoder.} The spiking encoder is a unique module used in SNNs that, encodes nonbinary input data into spikes. Fig.~\ref{figure: components}\textbf{f} illustrates the latency encoder in SpikingJelly, which can be formulated as 
\begin{align}
	S[t] &= 
	\begin{cases}
		1, t = t_{f}, \\
		0, t \neq t_{f},\\
	\end{cases}\\
	t_{f} &= [ (T_{max} - 1) \cdot(1 - x)],
\end{align}
where $S[t]$ is the output spike at time step $t$, $t_{f}$ is the firing time step, $[\cdot]$ is the rounding function,  $T_{max}$ is the latest firing time step, and $x$ is the input value. The latency encoder encodes a larger $x$ into an early firing time step $t_{f}$, indicating that a larger stimulus should cause a faster response, as has been observed in the retinal pathway \cite{gollisch2008rapid}. 

The Poisson encoder is another commonly used encoder in SNNs that, encodes inputs $x \in [0, 1]$ to Poisson events $\{s[t]\}$ with 
\begin{align}
	P(\sum_{t=0}^{T-1}s[t] = k) = \Pr(X=k) = \frac{(Tx)^{k}\exp(-Tx)}{k!}, k \in \{0,1,...,T\}.
\end{align}
For simplicity, an approximate implementation is used, which is provided in SpikingJelly as
\begin{align}
	P(S[t] = 1) = x.
\end{align}
Fig.~\ref{figure: components}\textbf{g} shows an example of an input image and the output spikes of the Poisson encoder. We find that the output spikes preserve most of the information from the original image. Other commonly used encoders, including weighted phase \cite{kim2018deep} and Gaussian tuning encoders\cite{BOHTE200217}, are also available in SpikingJelly.

\hspace*{\fill}

\noindent\textbf{Neuromorphic Datasets.} Neuromorphic datasets are frequently used as benchmarks for evaluating the temporal information processing abilities of SNNs \cite{10.3389/fnins.2021.608567}. SpikingJelly provides commonly used datasets, including ASL-DVS \cite{Bi_2019_ICCV}, CIFAR10-DVS \cite{10.3389/fnins.2017.00309}, DVS Gesture \cite{Amir_2017_CVPR}, ES-ImageNet \cite{10.3389/fnins.2021.726582}, HARDVS\cite{wang2022hardvs}, N-Caltech101 \cite{10.3389/fnins.2015.00437}, N-MNIST \cite{10.3389/fnins.2015.00437}, Nav Gesture \cite{10.3389/fnins.2020.00275}, and Spiking Heidelberg Digits (SHD) \cite{shd} (see \textit{Neuromorphic Datasets} in the supplementary materials). New datasets can be easily integrated by inheriting the dataset base and completing some processing functions.

Fig.~\ref{figure: components}\textbf{h} visualizes a series of events with two polarities drawn in blue and green. The raw data collected from the sensor is stored in a specific format, e.g., AEDAT, which requires a specific binary decoding method. To reduce the cost of use, decoding methods for different datasets are implemented in SpikingJelly. Correspondingly, the outputs of the datasets in SpikingJelly are in the NumPy format, which is compatible with all Python applications. The number of events in a sample can be in the millions and can hardly be handled directly by the network. Widely used subsampling methods integrate events into frames. Fig.~\ref{figure: components}\textbf{i} shows the 4 frames integrated from the events in Fig.~\ref{figure: components}\textbf{h}. SpikingJelly also provides frequently-used integration methods.

Fig.~\ref{figure: components}\textbf{j} shows the process flow of the datasets in SpikingJelly, which includes downloading from the source, extracting archives, decoding the raw data to an AER Python dictionary in the NumPy format, and downsampling events to frames. To add new datasets, the user only needs to inherit the base class and implement the "download", "extract" and "decode" functions, while another workload is handled by the base accelerated by multithreading, which also reveals the superior extensibility of SpikingJelly.

\subsection*{Functional Modules}

\noindent\textbf{Step Modes and Propagation Patterns.} In SpikingJelly, each module has a variable attribute called \textit{step\_mode}, which controls whether the module uses the single-step mode or multistep mode (see \textit{Design of the Step Module} in the supplementary materials). In the single-step mode, the module receives $X[t]$ and outputs $Y[t]$, which are data at a single time step. The input sequence for the $0$-th layer is denoted as $X_{0} = \{X_{0}[0], X_{0}[1], ..., X_{0}[T-1]\}$. When an SNN is stacked with $L$ single-step modules $\{M^{ss}_{0}, M^{ss}_{1}, ..., M^{ss}_{L-1}\}$, the SNN is simulated in the step-by-step propagation pattern, which is shown in Alg.~\ref{algo pps}\textbf{a}.

\begin{algorithm}
	\caption{Propagation patterns}\label{algo pps}
	\textbf{(a)~Propagation pattern: step-by-step}
	
	\textbf{Require:} an SNN stacked by $L$ single-step modules $\{M^{ss}_{0}, M^{ss}_{1}, ..., M^{ss}_{L-1}\}$, the input sequence for the $0$-th layer $X_{0} = \{X_{0}[0], X_{0}[1], ..., X_{0}[T-1]\}$
	
	Create an empty list $Y_{L-1} = \{ \}$
	
	\algorithmicfor{~$t \gets 0, 1, ... T-1$}
	
	~~~~\algorithmicfor{~$l \gets 0, 1, ... L-1$}
	
	~~~~~~~~$X_{l+1}[t] = Y_{l}[t] =  M^{ss}_{l}(X_{l}[t])$
	
	~~~~Append $Y_{L-1}[t]$ in $Y_{L-1} = \{Y_{L-1}[0], Y_{L-1}[1], ..., Y_{L-1}[t-1]\}$
	
	Output $Y_{L-1} = \{Y_{L-1}[0], Y_{L-1}[1], ..., Y_{L-1}[T-1]\}$
	
	\hrulefill
	
	\textbf{(b)~Propagation pattern: layer-by-layer}
	
	\textbf{Require:} an SNN stacked by $L$ multistep modules $\{M^{ms}_{0}, M^{ms}_{1}, ..., M^{ms}_{L-1}\}$,the input sequence for the $0$-th layer $X_{0} = \{X_{0}[0], X_{0}[1], ..., X_{0}[T-1]\}$
	
	\algorithmicfor{~$l \gets 0, 1, ... L-1$}
	
	~~~~$X_{l+1} = Y_{l} =  M^{ms}_{l}(X_{l})$
	
	Output $Y_{L-1} = \{Y_{L-1}[0], Y_{L-1}[1], ..., Y_{L-1}[T-1]\}$
\end{algorithm}

%\begin{algorithm}
%	\caption{Propagation pattern: step-by-step}\label{algo step-by-step}
%	\textbf{Require:} an SNN stacked by $L$ single-step modules $\{M^{ss}_{0}, M^{ss}_{1}, ..., M^{ss}_{L-1}\}$, the input sequence for $0$-th layer $X_{0} = \{X_{0}[0], X_{0}[1], ..., X_{0}[T-1]\}$
%	
%	Create an empty list $Y_{L-1} = \{ \}$
%	
%	\algorithmicfor{~$t \gets 0, 1, ... T-1$}
%	
%	~~~~\algorithmicfor{~$l \gets 0, 1, ... L-1$}
%	
%	~~~~~~~~$X_{l+1}[t] = Y_{l}[t] =  M^{ss}_{l}(X_{l}[t])$
%	
%	~~~~Append $Y_{L-1}[t]$ in $Y_{L-1} = \{Y_{L-1}[0], Y_{L-1}[1], ..., Y_{L-1}[t-1]\}$
%	
%	Output $Y_{L-1} = \{Y_{L-1}[0], Y_{L-1}[1], ..., Y_{L-1}[T-1]\}$
%\end{algorithm}
%
%
%\begin{algorithm}
%	\caption{Propagation pattern: layer-by-layer}\label{algo layer-by-layer}
%	\textbf{Require:} an SNN stacked by $L$ multistep modules $\{M^{ms}_{0}, M^{ms}_{1}, ..., M^{ms}_{L-1}\}$,the input sequence for $0$-th layer $X_{0} = \{X_{0}[0], X_{0}[1], ..., X_{0}[T-1]\}$
%	
%	\algorithmicfor{~$l \gets 0, 1, ... L-1$}
%	
%	~~~~$X_{l+1} = Y_{l} =  M^{ms}_{l}(X_{l})$
%	
%	Output $Y_{L-1} = \{Y_{L-1}[0], Y_{L-1}[1], ..., Y_{L-1}[T-1]\}$
%\end{algorithm}

When switching the \textit{step\_mode} to the multistep mode, the module receives a sequence $X = \{X[0], X[1], ..., X[T-1]\}$ and outputs a sequence $Y = \{Y[0], Y[1], ..., Y[T-1]\}$. Correspondingly, the SNN built with $L$ multistep modules $\{M^{ms}_{0}, M^{ms}_{1}, ..., M^{ms}_{L-1}\}$ can be simulated in a layer-by-layer manner, as Alg.~\ref{algo pps}\textbf{b} shows.

Fig.~\ref{figure: functional module}\textbf{a} shows how identical computational graphs are built in different orders by step-by-step and layer-by-layer propagation patterns. 
Considering that there are two dimensions, the time step and depth, in the computational graphs of SNNs, we can find that the step-by-step and layer-by-layer propagation patterns are DFS and BFS processes for traversing the computational graph, respectively.
The network can switch between two propagation patterns easily by changing the \textit{step\_mode} attribute of each layer in SpikingJelly. The chosen propagation mode is determined by the user's intent. The layer-by-layer pattern has the advantage of efficiency because the calculation over time steps can be executed in parallel for stateless layers and the fusion operations can be implemented for stateful layers, which will be discussed in the \textit{Acceleration} section. However, when using the layer-by-layer pattern, the inputs $X[t]$ at all time steps $t$ must be given simultaneously, which is impossible when $X[t+1]$ depends on $X[t]$, e.g., when we use the recurrent connections between layers. In such cases, the step-by-step pattern is preferred due to its flexible usage.

\hspace*{\fill}

\noindent\textbf{Conversion Methods.} In addition to surrogate gradient training, another way to obtain high-performance SNNs is through ANN-to-SNN conversion, also known as ANN2SNN. The conversion process is based on the fact that a rate-coded SNN contains an equivalence relation between the firing rate and the activation of the ANN~\cite{cao2015spiking, ding2021optimal} under the conditions that the ANN should use ReLU activations and average pooling. Based on this fact, one can transform a trained ANN into a corresponding SNN.

To provide a convenient resolution to the ANN2SNN problem, SpikingJelly has a \textit{Converter}  module that normalizes the weight of the ANN and converts the ANN to an SNN by replacing the ReLUs with IF neuron layers, which is shown in Fig.~\ref{figure: functional module}\textbf{b}. The neuronal charging function of the IF neuron is 
\begin{equation}
	H[t] = V[t-1] + X[t],
	\label{eq discrete IF neuronal charge}
\end{equation}
which has not decay and can approximate the ReLU in ANNs by firing rates.
The implementation in SpikingJelly is based on the modeling of postsynaptic potential~\cite{deng2021optimal, bu2021optimal, bu2022optimized} so that the average postsynaptic potential is equal to the activation of the original ANN. We provide a \textit{VoltageHook} module to record the maximum (or some percentile) of the hidden features as the scale of the postsynaptic current for the conversed SNN. The activation is then altered by the IF neuron layer to accomplish the conversion process. Furthermore, to perform fast inference (i.e., inference with few time steps), we provide a plug-and-play optimized membrane potential initialization\cite{bu2021optimal} in SpikingJelly.

\hspace*{\fill}

\noindent\textbf{Quantizer.} To perform inference with SNNs on resource-constrained hardware, including field-programmable gate arrays (FPGAs), neuromorphic chips, and mobile phones, network quantization is a necessary technique for reducing storage and computational costs. For example, a network quantized to 8 bits has a $4\times$ reduction in its model size and memory bandwidth requirements and up to $4\times$ throughput when compared with a 32-bit model. SpikingJelly provides a quantizer for quantization-aware training that can be used to quantize weights and neuronal dynamics during training. Fig.~\ref{figure: functional module}\textbf{c} shows a typical $k$-bit quantizer in SpikingJelly, which maps an input $x$ in the range $[0, 1]$ to the nearest fixed-point $y = q(x) = \frac{[(2^{k} - 1) \cdot x]}{2^{k} - 1}$, where $[\cdot]$ is the rounding function. Note that $q'(x)$ is zero almost everywhere, and SpikingJelly also uses the surrogate method to redefine its gradients.

\hspace*{\fill}

\noindent\textbf{Monitor.} The demands for recording data, e.g., the firing rates of spiking neuron layers, are frequently required by researchers. To satisfy the demand for recording data, SpikingJelly provides five general purpose monitors, which can monitor the input/output during forward/backward propagation and the attributes of specific layers. As Fig.~\ref{figure: functional module}\textbf{d} shows, a monitor acts like a probe, which is inserted into layers of the user-defined type and records the data with a custom transform $\lambda$. For example, if we are interested in the firing rates of all spiking neurons in an SNN, we can set the \textit{OutputMonitor}  with the layer type as the spiking neuron layer and the custom transform as $\lambda(S)=\frac{1}{T}\sum_{t=0}^{T-1}S[t]$, where $T$ is the number of time steps and $S[t]$ is the spike at time step $t$. With this monitor, we can easily record the firing rate of each spiking neuron layer.

\hspace*{\fill}

\noindent\textbf{STDP Learner.} Combining the local unsupervised STDP learning process and the global supervised surrogate gradient has become a promising learning method for deep SNNs \cite{wu2022brain}, and it can be implemented by the STDP learner in SpikingJelly. In the minimal example with two neurons shown in Fig.~\ref{figure: functional module}\textbf{e}, the STDP learner in SpikingJelly employs the "all-to-all" STDP with the trace method \cite{morrison2008phenomenological}, in which all the pre- and postsynaptic spike pairs are considered.
It uses monitors to record presynaptic spikes $s_{pre}$ and postsynaptic spikes $s_{post}$. Then the traces $tr_{pre}[t], tr_{post}[t]$ are updated as
\begin{align}
	tr_{pre}[t] &= tr_{pre}[t - 1] - \frac{tr_{pre}[t-1]}{\tau_{pre}} + s_{pre}[t], \\
	tr_{post}[t] &= tr_{post}[t - 1] - \frac{tr_{post}[t-1]}{\tau_{post}} + s_{post}[t],
\end{align}
where $\tau_{pre}, \tau_{post}$ are the time constants of presynaptic and postsynaptic traces, respectively. Then the weight $w$ is updated by the traces as 
\begin{align}
	\Delta w[t] = F_{post}(w[t]) \cdot tr_{pre}[t] \cdot s_{post}[t] - F_{pre}(w[t]) \cdot tr_{post}[t] \cdot s_{pre}[t],
\end{align}
where $F_{pre}$ and $F_{post}$ are custom user-defined functions for controlling the amplitudes of synapse changes. In SpikingJelly, $\Delta w$ can be added to the gradient $\frac{\partial \mathcal{L}}{\partial w}$, indicating that the STDP learner can work together with the gradient descent method and various kinds of optimizers, including momentum stochastic gradient descent (SGD) and adaptive moment estimation (Adam)\cite{kingma2014adam}.

\hspace*{\fill}

\noindent\textbf{Event Downsampling Methods.} The event-to-frame downsampling method is widely used in deep SNNs with various kinds of slicing and integration methods. Considering these diverse usages, SpikingJelly supports both predefined and user-defined custom slicing and integration methods. A commonly used downsampling method is to integrate events with a fixed time duration $\Delta T$, which can be formulated as

\begin{align}
	\textbf{F}[j, p, y, x] = \sum_{t_{0} + j \cdot \Delta T \leq t_{i} < t_{0} + (j + 1) \cdot \Delta T} \mathcal{I}_{p, x, y}(p_{i}, y_{i}, x_{i}),
\end{align}
where $\mathcal{I}_{p, y, x}(p_{i}, y_{i}, x_{i})$ is an indicator function and equals 1 only when $(p, y, x) = (p_{i}, y_{i}, x_{i})$. Fig.~\ref{figure: functional module}\textbf{f} visualizes the four slices obtained by the fixed time durations of events from a sample in the DVS Gesture dataset, and the four integrated frames are shown in Fig.~\ref{figure: functional module}\textbf{g}.

However, the fixed time duration-based integration generates frames with different frame numbers because the durations of the samples in neuromorphic datasets are not identical. Another frequently used method provided in SpikingJelly is fixed frame number-based integration\cite{fang2021incorporating}, which can be formulated as
\begin{align}
	\textbf{F}[j, p, y, x]  = \sum_{i = j_{l}}^{j_{r} - 1} \mathcal{I}_{p, x, y}(p_{i}, y_{i}, x_{i}), \label{eq: integrate event split by number}
\end{align}
where $j_{l}$ and $j_{r}$ are the indices created by a specific splitting method. For example, splitting by event number is performed as follows:
\begin{align}
	j_{l}  &= \lfloor \frac{N}{T} \rfloor \cdot j,\\
	j_{r}  &= \begin{cases}
		\lfloor \frac{N}{T} \rfloor \cdot (j + 1), & j <  T - 1, \\
		N, & j = T - 1,\\
	\end{cases}
\end{align}
where $\lfloor \cdot \rfloor$ is the floor operation, and $N$ is the number of events. Splitting by event duration is performed as:
\begin{align}
	\Delta T & = \lfloor\frac{t_{N-1} - t_{0}}{T}\rfloor, \\
	j_{l} & = \mathop{\arg\min}\limits_{k} \{t_{k} \mid t_{k} \geq t_{0} + \Delta T \cdot j\}, \\
	j_{r} & = \begin{cases} 
		\mathop{\arg\max}\limits_{k} \{t_{k} \mid t_{k} < t_{0} + \Delta T \cdot (j + 1)\} + 1, & j <  T - 1 \cr N, & j = T - 1
	\end{cases}.
\end{align}
Note that fixed frame number-based integration can be used only once all events have arrived, which makes it unsuitable for a real-time system whose event durations are unpredictable. 

User-defined custom slicing and integration methods are also supported in SpikingJelly; they are implemented as callable functions and are used in the initialized args for the dataset class.

\subsection*{Acceleration Modules}\label{sec: Acceleration}

\noindent\textbf{Principle of Acceleration.} SpikingJelly accelerates the simulation of SNNs via two methods, which are the parallelization of sequential computations for stateless layers and the fusion of CUDA kernels for stateful layers.

\begin{figure}
	\centering
	%	\subfloat[]{\includegraphics[width=0.4\textwidth,trim=100 570 210 40,clip]{./fig/seq_to_ann_forward.pdf}}
	%	\subfloat[]{\includegraphics[width=0.4\textwidth,trim=20 528 170 20,clip]{./fig/stateless_speed.pdf}}
	%	\vspace{-0.8cm}
	%	
	%	\subfloat[]{\includegraphics[width=0.4\textwidth,trim=20 520 110 30,clip]{./fig/neuron_speed.pdf}}
	%	\subfloat[]{\includegraphics[width=0.5\textwidth,trim=20 630 140 10,clip]{./fig/api_example6.pdf}}
	%	\vspace{-0.8cm}
	%	
	%	\subfloat[]{\includegraphics[width=0.9\textwidth,trim=280 190 520 90,clip]{./fig/cuda_code.pdf}}
	%	\vspace{-0.6cm}
	\includegraphics[width=0.78\textwidth,trim=18 156 18 18,clip]{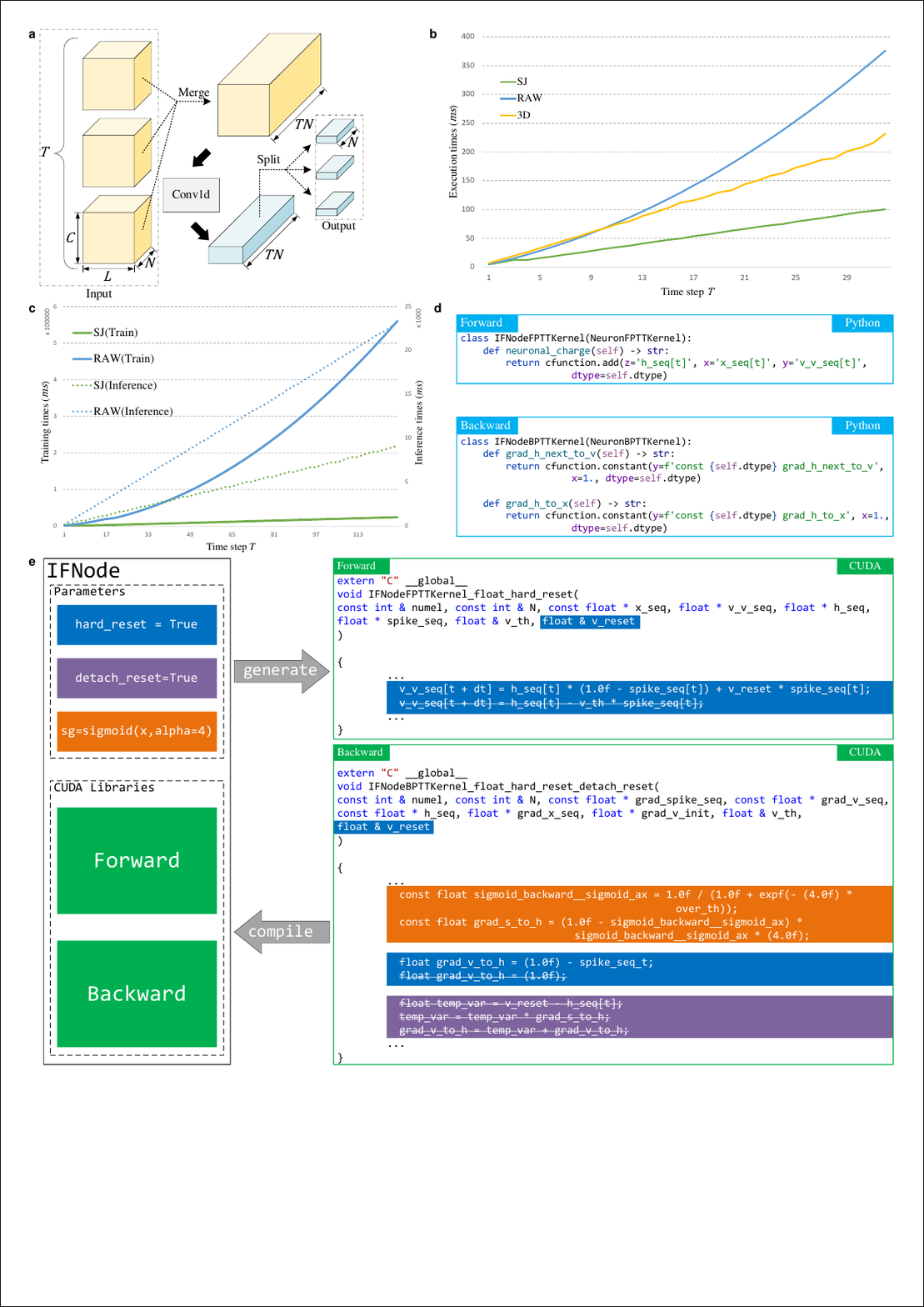}
	
\caption{\textbf{Acceleration in SpikingJelly.} 
	\textbf{a}. The use of \textit{seq\_to\_ann\_forward} for wrapping stateless layers and merging the time step dimension into the batch dimension to compute in parallel over time steps.
	\textbf{b}. Comparison of execution times for different implementations of applying 2D convolution on sequential data. ``SJ'' is SpikingJelly's method that merges the batch and time step dimensions, ``RAW'' is the plain for-loop for time steps, and ``3D'' regards the time step dimension as the depth dimension of the 3D convolution. 
	\textbf{c}. Comparison between the execution times of the LIF neuron in SpikingJelly (``SJ") accelerated by fused operations and the plain for-loop implemented by PyTorch (``RAW"). 
	\textbf{d}. Example of easy implementation of forward and backward kernels for an IF neuron with inheritance.
	\textbf{e}. An example of the IF neuron that shows the complete workflow of CUDA neurons in SpikingJelly. The parameters in each neuron are marked with different colors. Correspondingly, the CUDA codes governed by these parameters are also highlighted with the same colors. Strikethroughs such as \sout{\textit{abc}} represent deletions of the code \textit{abc}. 
}
	\label{figure: accelceration}
\end{figure}

Stateless layers, including convolutional and linear layers, are widely used in SNNs. Similar to stateful layers, these layers also receive a sequence $X = \{X[0], X[1], ..., X[T-1]\}$ with a shape of $(T, N, ...)$ as input. The difference between stateless and stateful layers is that the computation of $Y[t]$ in stateless layers only depends on $X[t]$. Thus, the for-loop concerning the time steps is not necessary since the computation has no temporal dependence. SpikingJelly provides the \textit{SeqToANNContainer} or its functional formulation \textit{seq\_to\_ann\_forward} to wrap stateless layers or their forward propagations. The \textit{SeqToANNContainer} merges the time step dimension into the batch dimension by reshaping the input tensor from a shape of $(T, N, ...)$ to $(T \cdot N, ...)$ before sending the input sequence to the stateless layer to execute the computation over all time steps in parallel, and splits the outputs into a sequence $Y = \{Y[0], Y[1], ..., Y[T-1]\}$ with the original sequence length. Fig.~\ref{figure: accelceration}\textbf{a} shows an example of using the \textit{SeqToANNContainer} to wrap a 1-D convolutional layer \textit{Conv1d}, whose input has a shape of $(N, C, L)$, where $C$ is the number of channels and $L$ is the input size. To avoid the for-loop concerning the time steps, using an $(n+1)$-D convolution to implement the $n$-D convolution is also a practicable method and has been applied in some frameworks. The $(n+1)$-D convolution method uses the time step dimension as the last dimension, and the input sequence has a shape of $(N, C, ..., T)$. Then, the $n$-D convolution imposed on the $(n+1)$-D input sequence is implemented by the $(n+1)$-D convolution with weight and stride values of 1 in the time step dimension. Fig.~\ref{figure: accelceration}\textbf{b} compares the execution times of different implementations that apply 2D convolution with 128 channels, a kernel size of 3, and a stride of 1 on the input with a shape of $(T, N, C, H, W) = (T, 16, 128, 64, 64)$, which is a common convolutional operation for deep SNNs. In Fig.~\ref{figure: accelceration}\textbf{b}, ``SJ'' is SpikingJelly's method that merges the batch and time step dimensions, ``RAW'' is the plain for-loop conducted for the time steps, and ``3D'' is using the time step dimension as the depth dimension of the 3-D convolution, which is the $(n+1)$-D convolution method. The results show that ``SJ'' is the best method because its execution time increases much more slowly than those of the other two methods with increasing time steps.

For stateful layers such as spiking neurons, $Y[t]$ depends on not only $X[t]$ but also the hidden states $H[t-1]$, which successively depend on $X[t-1], X[t-2], ... X[0]$. Thus, the for-loop concerning time steps is inevitable. However, when using the naive for-loop implemented in PyTorch, one or a series of CUDA kernels are invoked at each time step. The frequent launching of many small CUDA kernels wastes time on calling overhead including memory access time and kernel launch time, which decreases the computational efficiency of the network. To accelerate the stateful layers, SpikingJelly fuses many small kernels with a few large kernels, which avoids the calling overhead. Fig.~\ref{figure: accelceration}\textbf{c} compares the execution times of the training and inference processes of the LIF neuron in multistep mode implemented by the CUDA kernel fusion method (``SJ'') with the plain for-loop implemented by PyTorch (``RAW''). The results show that ``SJ'' is much faster than ``RAW''. It can also be found that as $T$ increases, the execution times of both neurons increase linearly during the inference process. However, the execution time of ``RAW" grows quadratically during training, while that of ``SJ" still increases slowly with $T$. For more details about the effect of acceleration methods in SpikingJelly, refer to \textit{Ablation Study of Acceleration Methods} in the supplementary materials.

Based on the above two acceleration methods, it is recommended to train deep SNNs with the layer-by-layer propagation pattern for higher efficiency. As Alg.~\ref{algo pps}\textbf{b} shows, each module receives sequence data, and then the parallelization of the sequential computations for stateless layers and the fusion of CUDA kernels for stateful layers can be employed. However, the spatial complexity of the layer-by-layer propagation pattern is $\mathcal{O}(T \cdot N)$, even during inference. When $T$ is large, e.g., hundreds and thousands of time steps in an ANN2SNN inference case, the layer-by-layer propagation pattern cannot be used due to memory restrictions. In such a case, the step-by-step propagation pattern can be used for inference due to its spatial complexity of $\mathcal{O}(N)$, which is not proportional to $T$.

\hspace*{\fill}

\noindent\textbf{Semi-automatic CUDA Code Generation.} The fusion of operations into one CUDA kernel has the highest computational efficiency but involves a massive workload for developers. Standard practices include writing CUDA code, wrapping CUDA functions in C++ interfaces, and binding C++ to Python via pybind. To simplify the development process, SpikingJelly uses CuPy\cite{nishino2017cupy} to implement CUDA kernels. After defining a raw CUDA kernel in a Python string, CuPy wraps and compiles the string to make a CUDA binary library, which is cached and reused in subsequent runs. The introduction of CuPy avoids the trivialities of binding between CUDA, C++ and Python, and helps users focus on a pure Python development environment.

A notable advantage of using CuPy is that CUDA codes are written in Python strings, indicating that we can easily control CUDA codes in a Python environment. Combined with the superior extensibility of SpikingJelly, this characteristic greatly simplifies the process of creating a CUDA kernel. As mentioned in Fig.~\ref{figure: components}\textbf{c}, the forward kernel and backward kernel for a specific neuron class are extended from the neuron kernel base class with a relatively small number of code changes (see \textit{Details of the Neuron Kernel} in the supplementary materials). According to Eq.~(\ref{eq discrete neuronal charge}), Eq.~(\ref{eq discrete neuronal fire}), and Eq.~(\ref{eq discrete neuronal reset}) for forward propagation, the backward propagation process is defined as
\begin{align}
	\frac{\mathrm{d} \mathcal{L}}{\mathrm{d} H[t]} &=\frac{\partial \mathcal{L}}{\partial S[t]}\frac{\mathrm{d} S[t]}{\mathrm{d} H[t]} + \Bigg(\frac{\partial \mathcal{L}}{\partial V[t]}+\frac{\mathrm{d} \mathcal{L}}{\mathrm{d} H[t+1]}\frac{\mathrm{d} H[t+1]}{\mathrm{d} V[t]}\Bigg)\frac{\mathrm{d} V[t]}{\mathrm{d} H[t]},\\
	\frac{\mathrm{d} \mathcal{L}}{\mathrm{d} X[t]} &= \frac{\mathrm{d} \mathcal{L}}{\mathrm{d} H[t]}\frac{\mathrm{d} H[t]}{\mathrm{d} X[t]},
\end{align}
where $\frac{\mathrm{d} S[t]}{\mathrm{d} H[t]}$ is defined by the surrogate function, $\frac{\mathrm{d} H[t+1]}{\mathrm{d} V[t]}, \frac{\mathrm{d} H[t]}{\mathrm{d} X[t]}$ is defined by Eq.~(\ref{eq discrete neuronal charge}), while the other parts are generic for different kinds of neurons. Thus, the neuron kernel base class in SpikingJelly is an incomplete CUDA kernel with vacant functions for the inheritor to complete, which includes Eq.~(\ref{eq discrete neuronal charge}) for the forward propagation process and $\frac{\mathrm{d} H[t+1]}{\mathrm{d} V[t]}, \frac{\mathrm{d} H[t]}{\mathrm{d} X[t]}$ for the backward propagation process. For example, when we use an IF neuron whose neuronal charging function and backward propagation process are defined as 
\begin{align}
	&H[t] = V[t-1] + X[t], \\
	&\frac{\mathrm{d} H[t+1]}{\mathrm{d} V[t]} = 1,\\
	&\frac{\mathrm{d} H[t]}{\mathrm{d} X[t]} = 1,
\end{align}
then the forward propagation through time (FPTT) kernel \textit{IFNodeFPTTKernel} and the backward propagation through time (BPTT) kernel \textit{IFNodeBPTTKernel} are defined as shown in Fig.~\ref{figure: accelceration}\textbf{d}. The FPTT kernel implements the neuronal dynamics of the IF neuron including Eqs.~(\ref{eq discrete IF neuronal charge}), (\ref{eq discrete neuronal fire}) and (\ref{eq discrete neuronal reset}) at $T$ time steps in a single CUDA kernel. Correspondingly, the BPTT kernel implements the backward propagation of the neuronal dynamics at all time steps in a single CUDA kernel (See Alg.~\ref{alg: IF BPTT} in the supplementary materials).
We find that completing these empty functions requires only a few lines of code.

Fig.~\ref{figure: accelceration}\textbf{e} shows the complete workflow of CUDA neurons in SpikingJelly by taking an IF neuron as the example. Parameters in the neuron are marked with different colors. Correspondingly, the CUDA codes governed by these parameters are also highlighted with the same color. For example, when \textit{hard\_reset} is true, the Python class performs the following operations, which are also shown by the blue highlights in Fig.~\ref{figure: accelceration}\textbf{e}:
\begin{itemize}
	\item add \textit{float \& v\_reset} in the forward and backward kernels' arguments;
	
	\item keep the hard reset commands \textit{v\_v\_seq[t + dt] = h\_seq[t] * (1.0f - spike\_seq[t]) + v\_reset * spike\_seq[t]} and delete the sort reset commands \textit{v\_v\_seq[t + dt] = h\_seq[t] - v\_th * spike\_seq[t]} in the forward kernel;
	
	\item keep the backward hard reset commands \textit{float grad\_v\_to\_h = (1.0f) - spike\_seq\_t} and delete the backward soft reset commands \textit{float grad\_v\_to\_h = (1.0f)} in the backward kernel.
\end{itemize}

When \textit{detach\_reset} is true, $S[t]$ in Eq.~(\ref{eq discrete neuronal reset}) is detached from the backward computation\cite{Zenke2020.06.29.176925}, and then the corresponding CUDA codes are deleted, as the purple highlights in Fig.~\ref{figure: accelceration}\textbf{e} show. The CUDA codes used by surrogate learning to define $\frac{\mathrm{d} S[t]}{\mathrm{d} H[t]}$ are generated by the surrogate function class, as the orange highlights in Fig.~\ref{figure: accelceration}\textbf{e} show. After the CUDA code generation process, CuPy compiles these CUDA codes to the CUDA library, and forward/backward computation can be accelerated by fusion operations within the large CUDA kernel.

\hspace*{\fill}

\noindent\textbf{JIT Acceleration.} JIT is an alternative method for accelerating a network, which can compile Python codes into high-performance C++/CUDA programs while running. Some simple elementwise operations, e.g., the neuronal firing (Eq.~(\ref{eq discrete neuronal fire})) and neuronal resetting (Eq.~(\ref{eq discrete neuronal reset})) operations, can be wrapped into one CUDA kernel by JIT. As a common optimization method, JIT has lower performance but also a lower development cost than specific optimization methods, such as manually writing CUDA codes. Thus, SpikingJelly uses semiautomatically generated CUDA kernels for complex operations such as forward and backward kernels through time steps with gradients for spiking neurons and uses JIT to accelerate other simple functions such as the neuronal dynamics at one time step. A typical example is shown in Fig.~\ref{figure: accelceration}\textbf{c}, in which the inference process of ``SJ'' is accelerated by JIT.

\bibliography{scibib}

\bibliographystyle{ScienceAdvances}

\subsection*{Acknowledgements}
The authors gratefully acknowledge contributors including Yifan Huang, Liuzhenghao LYU, Liutao Yu, Ruijie Zhu, Quanyu Pu, Yumin Ye, Haonan Qiu, Dongyang Ma, Ismail Khalfaoui-Hassani, Yichen Pan, Mingqing Xiao, Chen Sun, Yaoyu Zhu, Huanhuan Gao, Hengyu Zhang, Man Yao from GitHub and the OpenIntelligence (OpenI) community for codes commits, OpenI for bonus and server support, Zhengyu Ma and Peng Cheng Cloud Brain for computing resources, the Intel Neuromorphic Research Community for remote access to Intel Loihi, Yanyu Lin, Xueke Zhu, Wenjie Lin from Peng Cheng Laboratory and engineers from Lynxi for deployment on Lynxi KA200, Tong Bu, Xinhao Luo and Siyu Ding for deployment on Intel Loihi, Donghyun Lee and Zidong Zhou for explorations about CUDA accelerations of quantized math operations.
\subsubsection*{Funding}
This work is supported by grants from the National Natural Science Foundation of China (62027804, 61825101, 62088102, 62236009 and 62176003), the major key project of the Peng Cheng Laboratory (PCL2021A13), and Beijing Natural Science Fundation (JQ21015).

\subsection*{Author Contributions}
Yonghong Tian led the writing of the paper and proposed the initial idea of building SpikingJelly, and supervised the whole project.
Guoqi Li took part in the design of the structure of this paper, revised its contents, contributed to exchange modules, datasets, CUDA accelerations and popularized SpikingJelly in the community.
Wei Fang led the development of SpikingJelly and wrote this paper. Yanqi Chen took part in the design of the structure of SpikingJelly and contributed codes concerning complex neurons. Jianhao Ding contributed codes concerning ANN2SNN, and took part in the development of architectures of deep SNN models. Zhaofei Yu and Timothée Masquelier took part in designing the datasets, stateful modules and propagation patterns, and popularized SpikingJelly in the community. Ding Chen contributed codes concerning reinforcement learning and the application of reinforcement learning control. Liwei Huang contributed codes concerning encoders and the application of neural similarity. Huihui Zhou took part in the design of the structure of SpikingJelly and applied for bonus and computing resources. All authors reviewed the manuscript.

\subsection*{Competing Interests}
The authors declare no competing interests.

\subsection*{Data and Materials Availability}
The source codes of SpikingJelly can be found at its GitHub page: \url{https://github.com/fangwei123456/spikingjelly}.
Source codes for the experiment in Fig.~\ref{figure: framework}\textbf{d} and the typical applications in Fig.~\ref{figure: cases} are available at \url{https://doi.org/10.5281/zenodo.8310901}.

\section*{Supplementary Information}
\subsection*{Subpackages of Deep Learning}
To simplify the usage of SpikingJelly, coordinate different programming styles, promote extensibility, and reduce the cost of code reuse, the deep learning sections in SpikingJelly are elaborated into four parts: \textit{Components}, \textit{Functions}, \textit{Acceleration}, and \textit{Networks}. Table~\ref{tab: sub-packages} shows all these subpackages and the typical modules/functions in each part.

\textit{Components} contain essential modules including spiking neurons, layers, encoders, and surrogate functions, all of which are components of SNNs. For example, the Parametric LIF (PLIF) neurons \cite{fang2021incorporating} with learnable $\tau_{m}$ is implemented in the \textit{neuron} subpackage, whose neuronal charging function is
\begin{align}
	H[t] &= V[t - 1] + \tau_{m}(a) \cdot (X[t] - (V[t-1] - V_{reset})), \\
	\tau_{m}(a) &= \frac{1}{1 + \exp (-a)},
\end{align}
where $a$ is the learnable parameter that controls $\tau_{m}(a)$.

\textit{Functions} provide functions for training, simulating, analyzing, converting, quantizing, and deploying SNNs. For example, the FPTT online training algorithm for SNNs \cite{yin2021accurate} is implemented in the \textit{functional} subpackage, which is shown in Alg.~\ref{algo fptt online training}. Some modules in \textit{Functions} are the functional formulations of the corresponding modules in \textit{Components}, e.g., \textit{multi\_step\_forward} in \textit{Functions} and \textit{MultiStepContainer} in \textit{Components}. With these functional formulations, the user can reserve the module definition while modifying the forward propagation process by overriding the \textit{forward} function. This characteristic is especially suitable for model reuse because overriding \textit{forward} has no influence on the loading weights. For example, an ANN with trained weights can be easily modified to an SNN by overriding the \textit{forward} functions of the activation layers. 

\begin{algorithm}
	\footnotesize
	\caption{FPTT online training algorithm}\label{algo fptt online training}
	
	\textbf{Require:} 
	
	~~~~ the network $f_{\mathbf{W}}$ with parameters $\mathbf{W}$ and hidden states $\mathbf{H}[-1]$
	
	~~~~ the optimizer for computing gradients with learning rate $\eta$
	
	~~~~ the input sequence at all time steps $\{X[t]\}, t=0,1,...,T-1$
	
	~~~~ the target sequence at all time steps $\{O[t]\}, t=0,1,...,T-1$
	
	~~~~ the loss function $l$ for computing loss at each time step
	
	~~~~ the gradient scale $\alpha$
	
	~~~~ the running average of parameters $\bar{\mathbf{W}}$

	initialize $\frac{\partial L[-1]}{\mathbf{W}[0]} = 0, \mathbf{W}[0] = \bar{\mathbf{W}}[0] = \mathbf{W}$
	
	\algorithmicfor{~$t \gets 0, 1, ..., T - 1$}
	
	~~~~~~ detach all hidden states $\mathbf{H}[t-1]$ to avoid BPTT
	
	~~~~~~ $Y[t] = f_{\mathbf{W}[t]}(X[t], \mathbf{H}[t-1])$
	
	~~~~~~ $L[t] = l(Y[t], O[t]) + \frac{\alpha}{2}||\mathbf{W}[t] - \bar{\mathbf{W}}[t] - \frac{1}{2\alpha} \cdot \frac{\partial L[t-1]}{\partial \mathbf{W}[t]}||^{2}$

	~~~~~~ update $\mathbf{W}[t]$ by the optimizer with gradients $\frac{\partial L[t]}{\partial \mathbf{W}[t]}$

	~~~~~~ $\bar{\mathbf{W}}[t + 1] = \frac{1}{2} (\bar{\mathbf{W}}[t] + \mathbf{W}[t + 1]) - \frac{1}{2\alpha} \mathbf{W}[t + 1]$

\end{algorithm}

\textit{Acceleration} provides extra optional functions for accelerating the simulation processes of SNNs via semiautomatically generated CUDA kernels, which exploits the efficiency of low-level programming languages and reduces the development cost through the code generation technique. Some optimized layers using binary characteristics to reduce memory consumption are also integrated into \textit{Acceleration}.

Based on the above sections, \textit{Networks} provide classic and large-scale network structures. Classic structures such as regular SNNs are used in the tutorial to provide a low-cost learning process for new users and help them get started with SpikingJelly quickly. Large-scale network structures, including the PLIF Net \cite{fang2021incorporating}, Spiking VGG, Spiking ResNet, and SEW ResNet \cite{SEWResNet} are provided with training scripts, which can be employed for model reuse. 
%For example, the $l$-th block in the SEW ResNet provided in the subpackage \textit{model} is defined as
%\begin{align}
%	O^{l}[t] = g(\mathrm{SN}(\mathcal{F}^{l}(S^{l}[t])), S^{l}[t]),
%\end{align}
%where $\mathcal{F}^{l}$ represents the stacked convoltuional layers, $S^{l}[t]$ is the input spike and $O^{l}[t]$ is the output for the $l$-th block at the time step $t$, $\mathrm{SN}$ is the spiking neuron layer, and $g$ is the spike-element-wise function.

\begin{table}[]
	\resizebox{\linewidth}{!}
	{
		\begin{tabular}{|l|l|l|l|}
			\hline
			\textbf{Section}              & \textbf{Supackage} & \textbf{Description}                              & \textbf{Typical Module/Function}           \\ \hline
			\multirow{4}{*}{Components}   & neuron               & spiking neurons                                   & \makecell[l]{LIF neuron\\Parametric LIF neurons \cite{fang2021incorporating} with learnable $\tau_{m}$}                                 \\ \cline{2-4} 
			& layer                & synapses and containers                           & \makecell[l]{stateful synapses\\multi dimensional attention \cite{10032591}}                        \\ \cline{2-4} 
			& encoding             & spiking encoders                                   & \makecell[l]{Poisson encoder\\weighted phase encoder \cite{kim2018deep}}                            \\ \cline{2-4} 
			& surrogate            & surrogate gradient functions                      & \makecell[l]{sigmoid surrogate function\\arctan surrogate function}                \\ \hline
			\multirow{4}{*}{Functions}    & ann2snn              & conversion functions for ANN2SNN                  & \makecell[l]{conv/bn fusion function\\maximum activation normalization}           \\ \cline{2-4} 
			& functional           & forward, loss and setting functions               & \makecell[l]{reset states of the whole network\\FPTT online training\cite{yin2021accurate}}          \\ \cline{2-4} 
			& monitor              & monitors for recording data from specific layers  & \makecell[l]{attribute monitor for recoding layers' attributes\\output monitor for recoding layers' outputs} \\ \cline{2-4} 
			& learning             & biologically plausible learning rules               & \makecell[l]{STDP learner\\modulated STDP learner with eligibility trace\cite{florian2007reinforcement}}                               \\ \cline{2-4}
			& quantize             & quantizers               & \makecell[l]{rounding function with surrogate gradients\\$k$ bits quantizer}                               \\ \cline{2-4} 
			& exchange             & conversion modules for neuromorphic computation chips             & \makecell[l]{simplified IF neuron\\packing synapses and neurons into Lava blocks}                               \\ \hline
			\multirow{2}{*}{Acceleration} & auto cuda        & CUDA kernels for acceleration                     & \makecell[l]{sigmoid surrogate gradient function\\CUDA kernel for the BPTT of LIF neurons}         \\ \cline{2-4} 
			& spike op             & storage optimization layers for spike operations & \makecell[l]{float-to-char spike CUDA conversion function\\spike convolutional layer}                  \\ \hline
			\multirow{2}{*}{Networks}     & model                & predefined classic SNNs                          & \makecell[l]{Spiking ResNet \\ SEW ResNet \cite{SEWResNet}}                                \\ \cline{2-4} 
			& example             & friendly examples with tutorials                  & \makecell[l]{tutorials about training spiking CNNs\\tutorials about training large-scale deep SNNs}        \\ \hline
		\end{tabular}
	}
	\caption{\textbf{Introduction of subpackages in SpikingJelly.}}
	\label{tab: sub-packages}
\end{table}

\subsection*{Using Examples}
Building and training SNNs can be easily implemented with SpikingJelly, as will be shown in the next two examples. For more examples, the tutorial on the SpikingJelly homepage is recommended.

\subsubsection*{Build a Recurrent SNN}
\begin{figure}
	\centering
	\subfloat[]{\includegraphics[width=1.\textwidth,trim=100 620 90 70,clip]{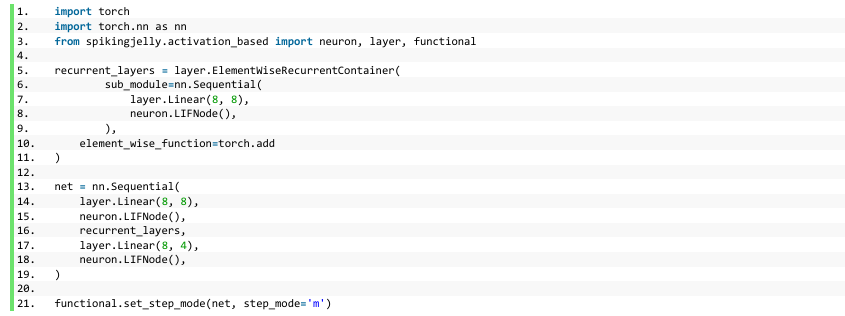}}
	
	\subfloat[]{\includegraphics[width=1.\textwidth,trim=10 700 180 20,clip]{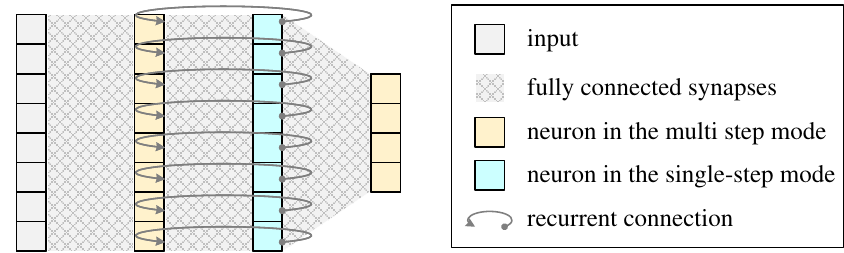}}
	\caption{The example of building an SNN with recurrent connections. \textbf{a}. Source codes. \textbf{b}. The network structure.}
	\label{figure: example of recurrent SNN}
\end{figure}

Fig.~\ref{figure: example of recurrent SNN} shows the example of building an SNN with recurrent connections. Fig.~\ref{figure: example of recurrent SNN}\textbf{a} is the source codes of this example. In \textbf{lines 5-11}, the recurrent structure is composed of a fully connected layer $FC$ and an LIF neuron layer $LIF$ wrapped by the element-wise recurrent container with the element-wise function as the addition function. Denote the external input and output at the time step $t$ for the recurrent structure as $X[t]$ and $Y[t]$, then the forward propagation of this recurrent structure can be formulated as 
\begin{align}
	Y[t] = LIF(FC(X[t] + Y[t-1])).
\end{align}
It is worth noting that the element-wise function of an element-wise recurrent container can be freely defined by the user. In \textbf{lines 13-19}, we build the SNN with two feedforward layers before and after the recurrent structure, respectively, and the structure of the SNN is shown in Fig.~\ref{figure: example of recurrent SNN}\textbf{b}.

Note that the recurrent structure can only work in the single-step mode because the actual input $X[t] + Y[t-1]$ at all time steps can not be given at $t=0$. In \textbf{line 21} of Fig.~\ref{figure: example of recurrent SNN}\textbf{b}, we use the \textit{set\_step\_mode} function to set the step mode of all modules in the SNN to the multistep mode, which will not take effect for the $FC$ and $LIF$ in the recurrent structure. These modules still use the single-step mode. But other modules in the SNN, including the recurrent container, will use the multistep mode. The hybrid step modes in this SNN result in an interesting situation, in which we simulate this SNN with both propagation modes employed. The recurrent part (the modules inside the recurrent container) uses the step-by-step mode. The feedforward part of the SNN uses the layer-by-layer propagation pattern, and the acceleration methods based on the multistep mode such as the fusion of time step and batch dimensions and the fused CUDA kernel can still be exploited.

\begin{figure}
	\centering
	\includegraphics[width=1.\textwidth,trim=100 270 90 70,clip]{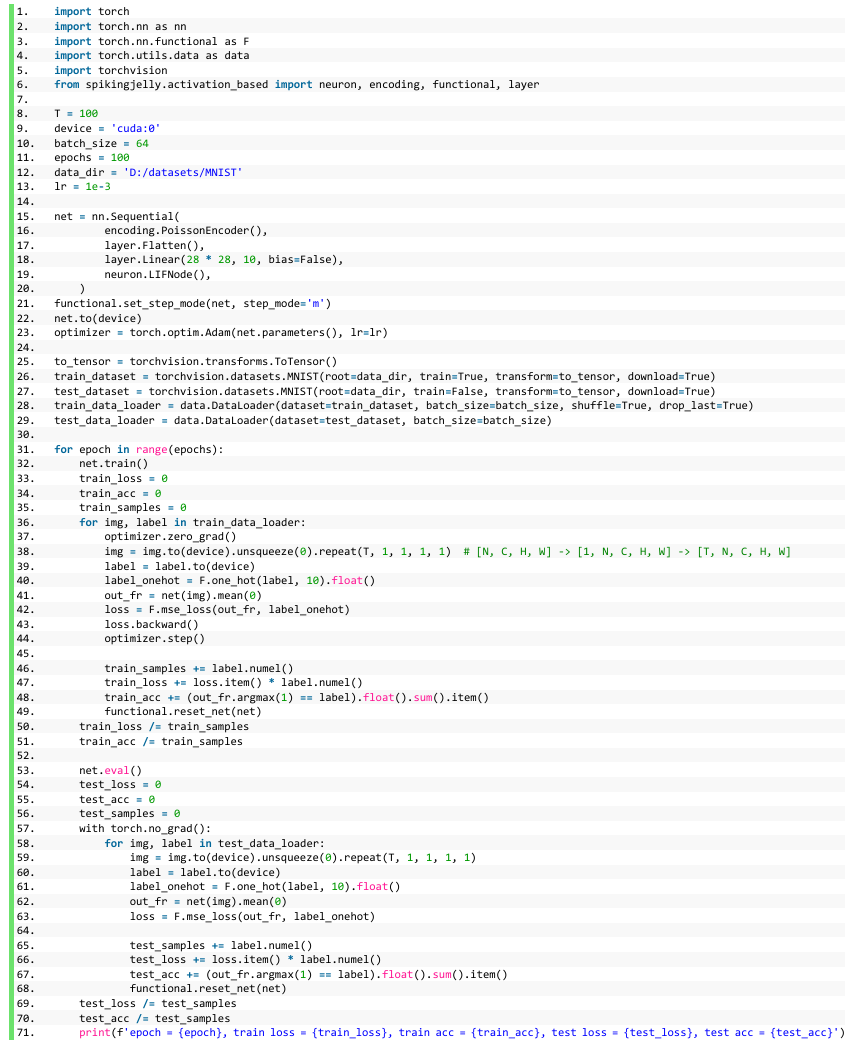}
	\caption{The example of building and training an SNN for classifying the MNIST dataset}
	\label{figure: example of train an SNN}
\end{figure}

\subsubsection*{Train an SNN for the Classification Task}
Fig.~\ref{figure: example of train an SNN} shows the code of using SpikingJelly to build and train an SNN for classifying the MNIST dataset. The code will be familiar to those who are proficient in using PyTorch. In \textbf{lines 8-13}, we define the hyper-parameters including the number of time steps $T$, the device, the batch size, the number of training epochs, the dataset path, and the learning rate. In \textbf{lines 15-20}, we define the SNN as composed of a Poisson encoder, a flatten layer, a fully connected layer, and an LIF neuron layer. A Poisson encoder is employed to transform the floating input values into spikes. The flatten layer is used to reshape input with a shape $(..., C, H, W)$ to a shape $(..., C \cdot H \cdot W)$, where $C, H, W$ are the number of channels, height, and width. In \textbf{line 21}, we set all modules in the SNN to use the multistep mode. In \textbf{line 22}, we move the SNN to the designative device. Note that SpikingJelly is fully compatible with CPUs and GPUs, and the device can also be \textit{cpu}. In this example, we set the device as \textit{cuda:0}, which is the $0$-th GPU in the training environment. In \textbf{lines 23-29}, the Adam optimizer, train, and test dataset loaders are defined, which is the standard practice of training networks. In \textbf{lines 31-71}, the training loop is defined. In \textbf{line 32}, we set the SNN to the training mode, which is a concept from PyTorch obeyed in SpikingJelly. Modules such as batch normalization layer and dropout layer have different behaviors during training and inference, and need to be distinguished by the training/inference mode. Thus, \textbf{line 32} and \textbf{line 53} are the mode settings for training and inference. In \textbf{line 33-35}, we initialize variables to record train loss, train accuracy, and samples during training. In \textbf{lines 36-49}, the mini batch gradient descent over the train set is implemented. In \textbf{line 38}, the original image has a shape of $(N, C, H, W)$, which does not have a time step dimension. We first unsqueeze it to the shape $(1, N, C, H, W)$, and then repeat it on the time step dimension with $T$ repeats to the shape $(T, N, C, H, W)$. In \textbf{line 41}, the input sequence is sent to the SNN and the output sequence has a shape of $(T, N, 10)$. We use the firing rate of the output sequence to compute the loss and obtain the classification result. By this, we mean the output at the time step dimension. In \textbf{line 42}, the loss is defined as the mean squared error between the firing rate and the label in the one-hot format. In \textbf{lines 43-44}, we call the backward propagation to compute gradients and update parameters by the optimizer. Note that the surrogate learning method is used in this process, which is implemented inside the spiking neuron layer. In \textbf{line 48}, the classification result is regarded as the index with the highest firing rate, as the \textit{argmax} used in code. The inference procedures defined in \textbf{lines 53-70} are similar to the train procedures.

This simple SNN achieves 92.9\% test accuracy. For more details about training and results, please refer to the "Single Fully Connected Layer SNN to Classify MNIST" tutorial on SpikingJelly's homepage. This example is a simplified version of the tutorial.

\subsection*{Design of the Step Module}
\begin{figure}
	\centering
	\includegraphics[width=1.\textwidth,trim=20 430 200 20,clip]{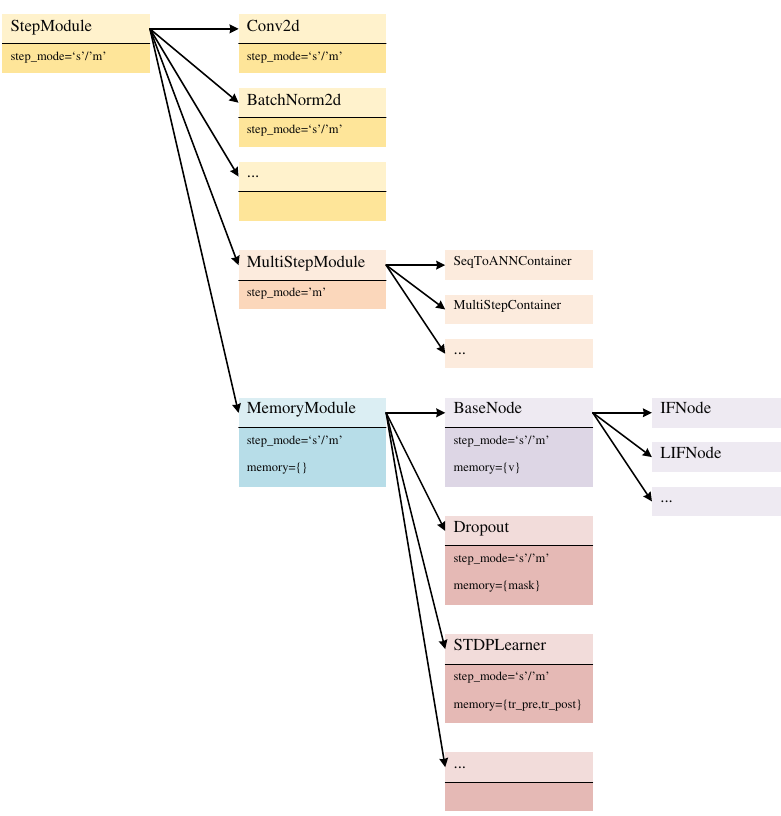}
	\caption{\textbf{Inheritance map of the \textit{StepModule}.}
		The \textit{StepModule} is the core base class, whose \textit{step\_mode} attribute controls whether the module uses the single-step mode for data with a shape of $(N, ...)$ or the multistep mode for data with a shape of $(T, N, ...)$. Inherited from the \textit{StepModule}, stateless modules such as \textit{Conv2d} can be easily implemented with support for both step modes. By fixing \textit{step\_mode='s'}, the child class \textit{MultiStepModule} is used as the base for those modules that can only work in the multistep mode. By adding the \textit{memory} attribute for storing hidden states such as membrane potentials for spiking neurons, the \textit{MemoryModule} is implemented as the base for stateful modules.
	}
	\label{figure: step module and memory module}
\end{figure}

The \textit{StepModule} is the core base class in SpikingJelly, whose attribute \textit{step\_mode} controls the step mode. Fig.~\ref{figure: step module and memory module} shows the inheritance map of the \textit{StepModule}. When \textit{step\_mode = 's'}, the module works in the single-step mode and uses data with a shape of $(N, ...)$. When \textit{step\_mode = 'm'}, the module works in the multistep mode and uses data with a shape of $(T, N, ...)$. Inherited from the \textit{StepModule}, stateless modules such as \textit{Conv2d} are implemented with support for two step modes, and the multistep forward process is accelerated by merging the time step dimension into the batch dimension. Note that when all modules in the SNN are based on the \textit{StepModule}, the SNN can easily switch between the step-by-step and layer-by-layer propagation patterns, which satisfies flexible simulation like training with a small $T$ in a layer-by-layer manner to achieve faster speed and lower memory consumption, while inference is performed with a large $T$ in a step-by-step manner.

The \textit{MultiStepModule} is inherited from the \textit{StepModule} by fixing \textit{step\_mode = 'm'} to force itself to run in the multistep mode. Modules such as the \textit{SeqToANNContainer} are based on the \textit{MultiStepModule}.

Stateful modules such as spiking neurons are widely used in SNNs. Different from the PyTorch style, which regards hidden states as parts of the inputs and outputs, SpikingJelly stores hidden states inside a module, which is implemented by a \textit{MemoryModule}. Based on the \textit{StepMode}, the \textit{MemoryModule} adds an extra Python dictionary \textit{memory} to store hidden states. For example, the \textit{memory} in the neuron base \textit{BaseNode} stores \textit{v}, which is the membrane potential $V[t]$ of the spiking neuron. Another example is that the dropout layer in the SNNs implemented by \textit{Dropout} in SpikingJelly uses a time-invariant mask \cite{lee2020enabling}, which is also regarded as an attribute of \textit{memory}. 

\subsection*{Distinction of Step Modes and Propagation Patterns}
The step mode controls the behavior of propagation for a module. By setting the \textit{step\_mode} values of all modules as \textit{"s"} (single-step) or \textit{"m"} (multistep), the SNN can run in a step-by-step or layer-by-layer propagation pattern. Thus, the difference between the two step modes is largely reflected in the distinction of the propagation patterns. Suppose that the number of layers of an SNN is $L$, the number of time steps is $T$, the batch size is $N$, and the $i$-th layer in the SNN takes $d_{s}[i]$ to process data of a single time step, and takes $d_{m}[i]$ to process data of all time steps. $d_{s}[i]$ and $d_{m}[i]$ are also the latency of the $i$-th layer in two step modes. Denote $X[t]$ as the input and $Y_{L-1}[t]$ as the output of the SNN at the time step $t$.

\subsubsection*{Memory Consumption}
\begin{figure}
	\centering
	\includegraphics[width=1.\textwidth,trim=4 270 16 8,clip]{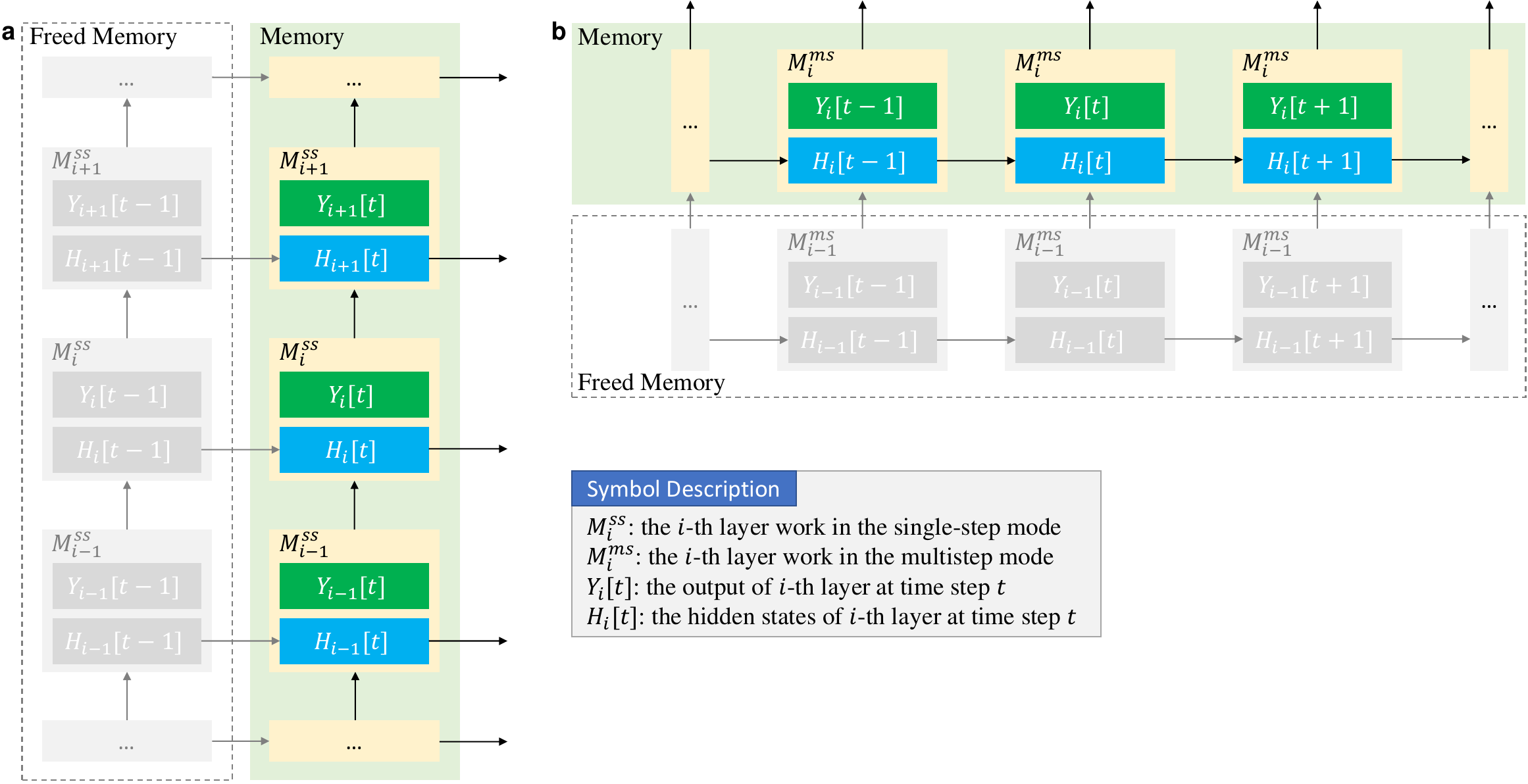}
	\caption{\textbf{The spatial inference complexities of two propagation patterns.} Suppose that the number of layers of an SNN is $L$ and that the number of time steps is $T$.
		\textbf{a}. When using step-by-step inference and propagating at time step $t$, the data at time step $t-1$ in the memory can be freed. The spatial complexity is $\mathcal{O}(N \cdot L)$ for the step-by-step propagation pattern. 
		\textbf{b}. When using layer-by-layer inference and propagating to the $i$-th layer, the data of the $(i-1)$-th layer in the memory can be freed. The spatial complexity is $\mathcal{O}(N \cdot T)$ for the layer-by-layer propagation pattern. 
	}
	\label{figure: propagation patterns memory}
\end{figure}

When training the SNN by BPTT, the spatial complexities of the two propagation patterns are both $\mathcal{O}(N \cdot T\cdot L)$ because their complete computing graphs are identical, as shown in Fig.~\ref{figure: functional module}. 
When performing inference with an SNN, the spatial complexity is different due to the locality of data and computation. 
As Fig.~\ref{figure: propagation patterns memory}\textbf{a} shows, when using step-by-step inference and propagation at time step $t$, the computations are localized at a single time step, and then all data, including the outputs and hidden states of all layers at time step $t-1$, can be discarded. Correspondingly, the spatial inference complexity of the step-by-step propagation pattern is $\mathcal{O}(N \cdot L)$, which is not proportional to $T$. In this way, the step-by-step propagation pattern is suitable for inference simulations with many time steps, such as ANN2SNN.
When using layer-by-layer inference and propagating to the $i$-th layer, as shown in Fig.~\ref{figure: propagation patterns memory}\textbf{b}, the computations are localized in a single layer, and then all data of the previous layers can be discarded. Thus, the spatial inference complexity of the layer-by-layer propagation pattern is $\mathcal{O}(N \cdot T)$, which is appropriate for performing inference with extremely deep SNNs.

\subsubsection*{Latency}

When the $i$-th layer works in the single-step mode, it requires $d_{s}[i]$ to process the input $Y_{i-1}[t]$ and output $Y_{i}[t]$. When all layers use the single-step mode and the SNN runs in a step-by-step propagation pattern, the latency between the input $X[0]$ and the output $Y_{L-1}[T-1]$ is 
\begin{align}
	D_{s} = T \cdot \sum_{i=0}^{L-1} d_{s}[i].
\end{align}
When the $i$-th layer works in the multistep mode, it requires $d_{m}[i]$ to process the input $Y_{i-1} = \{Y_{i-1}[0], Y_{i-1}[1], ..., Y_{i-1}[T-1]\}$ and output $Y_{i} = \{Y_{i}[0], Y_{i}[1], ..., Y_{i}[T-1]\}$. If the devices only support serial computing, then $d_{m}[i] = T \cdot d_{s}[i]$.

When running on GPUs, $d_{m}[i]$ can be optimized. If the $i$-th layer is stateless, computation in all time steps is parallel in SpikingJelly, and $d_{m}[i] \approx d_{s}[i]$ in an ideal situation, which is the memory reading and writing has not reached the bottleneck, and the number of concurrent threads does not exceed the maximum parallel processing capacity of the GPU.
If the $i$-th layer is stateful, the computation across time steps can be fused into a large CUDA kernel in the multistep mode. Compared with the single-step mode which calls small CUDA kernels in each time step, the multistep mode has much lower calling overhead and correspondingly has faster running speed, resulting in $d_{m}[i] << T \cdot d_{s}[i]$ when $T$ is large. When all layers use the multistep mode and the SNN runs in a layer-by-layer propagation pattern, the latency between the input $X[0]$ and the output $Y_{L-1}[T-1]$ is 
\begin{align}
	D_{m} = \sum_{i=0}^{L-1} d_{m}[i].
\end{align} 
Due to the fact that $d_{m}[i] \leq T \cdot d_{s}[i]$ almost always holds, we can find that $D_{m} \leq D_{s}$.

It is worth noting that the above latency analysis does not consider the latency of the input sequence itself. The layer-by-layer propagation pattern requires inputs at all time steps, indicating that the latency in these tasks is
\begin{align}
	D_{m}' = D_{m} + t_{X[T-1]} - t_{X[0]},
\end{align} 
where $t_{X[0]}$ is the time when $X[0]$ arrives and $t_{X[T-1]}$ is the time when $X[T-1]$ arrives. 

\begin{figure}
	\centering
	\includegraphics[width=1.\textwidth,trim=00 0 0 0,clip]{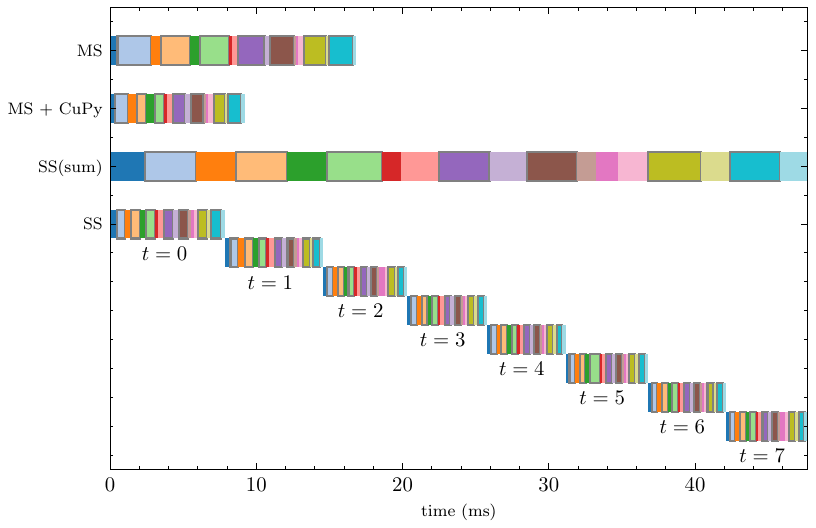}
	\caption{\textbf{Latency of each layer during the forward propagation in an SNN.}
		Each layer is represented by a unique color.
		Spiking neuron layers are labeled with gray edge lines, and other layers without gray edges are stateless. 
		``MS'' is using the multistep mode, ``CuPy'' is enabling the CuPy backend, ``SS'' is using the single-step mode. ``SS (sum)'' is a merged plot from ``SS'' that sums the latency of each layer in all time steps.}
	
	\label{figure: latency forward}
\end{figure}

\begin{figure}
	\centering
	\includegraphics[width=1.\textwidth,trim=00 0 0 0,clip]{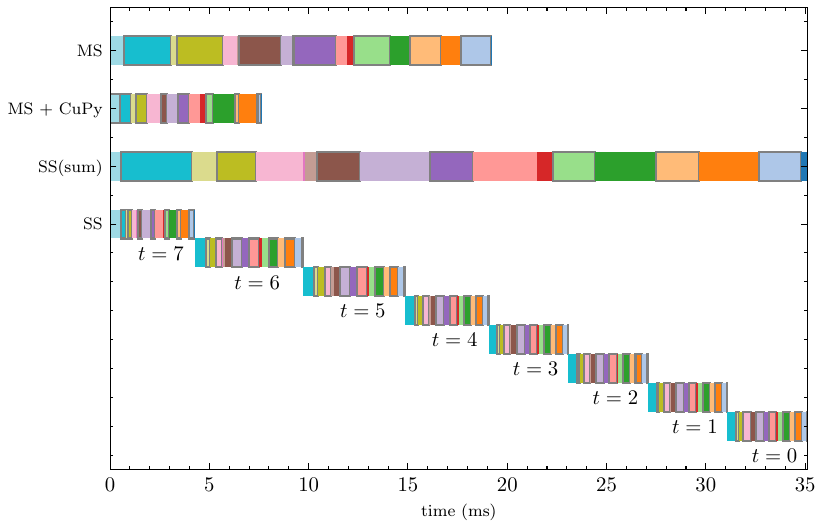}
	\caption{\textbf{Latency of each layer during the backward propagation in an SNN.}
		Refer to Fig.~\ref{figure: latency forward} for more details about legends.
	}
	
	\label{figure: latency backward}
\end{figure}

To prove our analysis, we build a deep SNN with a commonly used structure $\{\{Conv2d-IF\} * 2 - MaxPool\} * 2 - Flatten -\{ FC - IF\} * 2 - Loss$, where $Conv2d$ is the convolutional layer, $IF$ is the IF neuron layer, $MaxPool$ is the max pooling layer, $*2$ represents 2 repeated modules, $Flatten$ is the flatten layer, $FC$ is the fully connected layer, and $Loss$ is the module to compute loss. We trained this SNN with $T=8$ time steps on the GPU. We measured the used time of the forward and backward propagations of each layer during training. With these data, we plot the latency of each layer, which is shown in Fig.~\ref{figure: latency forward} and Fig.~\ref{figure: latency backward}. These two figures clearly show that when using the multistep mode, the latency is much slower than the single-step mode, with or without CuPy enabled. Comparing ``MS'' and ``SS (sum)'', we can find that the latencies of stateless layers in the multistep mode are much smaller than those in the single-step mode, which is caused by the merging of time step and batch dimensions. Comparing ``MS + CuPy'' with ``MS'', we can find that the latencies of spiking neurons are largely reduced by the CuPy backend, which uses a large CUDA kernel to fuse operations. Although the spiking neurons without using the CuPy backend in ``MS'' and ``SS'' and the stateless layers in ``MS + CuPy'' and ``MS'' have the same behavior, their latencies are still different. This can be explained by the fact that when running tasks named A and B sequentially, if the efficiency of task A is low, the GPU will need more time for post-processing such as memory allocation and resetting caches and streams. Then the efficiency of task B is also slowed down. Thus, layers with the same behavior are still faster when the other layers in the SNN are running in a more efficient mode.
As a conclusion, this experiment verifies that $d_{m}[i] \leq T \cdot d_{s}[i]$, which results in $D_{m} \leq D_{s}$.

\subsubsection*{Conclusions of Step Modes and Propagation Patterns}
The step modes of modules in SNNs control the propagation patterns. Two patterns are in different orders of generating the computational graph, and the outputs of SNNs in the two patterns do not differ. But different orders affect speed, memory consumption, and latency.

Switching propagation patterns in SpikingJelly is very easy, which is done by changing the step modes of all modules. The users can choose the step mode for specific purposes. In most cases, the multistep mode is highly recommended for faster speed. When inferring SNNs and the memory is limited, the multistep mode can be used when $L$ is large, while the single-step mode can be used when $T$ is large. 

Some specific networks, tasks, and hardware are not compatible with both patterns. There exist SNNs that can only work in the layer-by-layer propagation pattern, such as the SNN using time-to-first-spike coding \cite{lew2022time, 10.5555/3437539.3437564}, event-driven backpropagation \cite{zhu2022training}, and attention SNNs \cite{10032591}. These SNNs involve operation on data at all time steps and thus can only be simulated in the layer-by-layer pattern. In real-world tasks, the input $X[t]$ is usually generated in some intervals, and $T$ can be infinity in some online tasks which requires the network to output $Y[t]$ as soon as possible when it receives $X[t]$. The SNN with recurrent connections \cite{rao2022long, yin2021accurate} use $Y[t-1]$ as part of inputs at the time step $t$, which also makes it impossible to get input data at all time steps.
In these cases, only the step-by-step pattern can be used. To the best of our knowledge, the behavior of most event-driven neuromorphic chips is closer to the step-by-step pattern that each event is processed and routed between cores, while some heterogeneous chips \cite{kim2023c, chang202373} also use the layer-by-layer pattern. In general, both propagation patterns are widely used in the spiking deep learning community, and they are all well-supported by SpikingJelly.

\subsection*{Examples of Implementing Complex Neurons}
Benefiting from multilevel inheritance, users can define complex neurons based on base neurons with little effort, as we illustrate with two implementations of complex neurons.

\subsubsection*{Implementation of the Adaptive-Leaky Integrate-and-Fire Neuron}
The first example is implementing the Adaptive-Leaky Integrate-and-Fire (ALIF) neuron \cite{bellec2018long}, which extends the LIF neuron with the threshold dynamics: 
\begin{align}
	V_{th}[t] &= B^{0} + \beta \cdot B[t], \label{eq: ALIF threshold dynamics 1} \\
	B[t + 1] &= \rho \cdot B[t] + (1 - \rho) \cdot S[t], \label{eq: ALIF threshold dynamics 2}
\end{align}
where $V_{th}[t]$ is threshold and $B[t]$ is the dynamic offset at the time step $t$. $B^{0}$ is the baseline threshold, and $\beta, \rho$ are parameters that control the threshold dynamics.

\begin{figure}
	\centering
	\subfloat[]{\includegraphics[width=1.\textwidth,trim=100 610 90 70,clip]{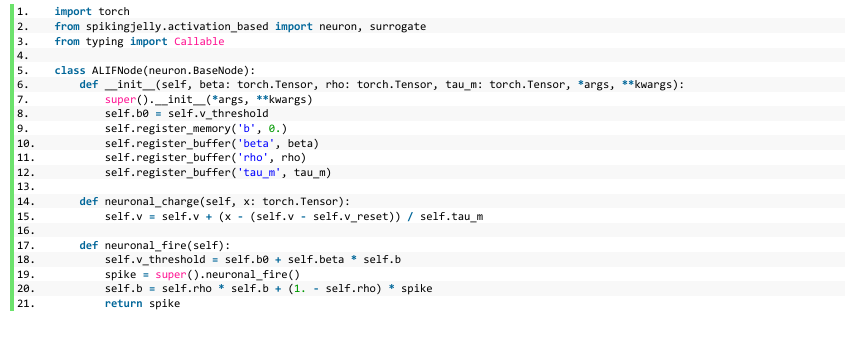}}
	
	\subfloat[]{\includegraphics[width=0.8\textwidth,trim=0 0 0 0,clip]{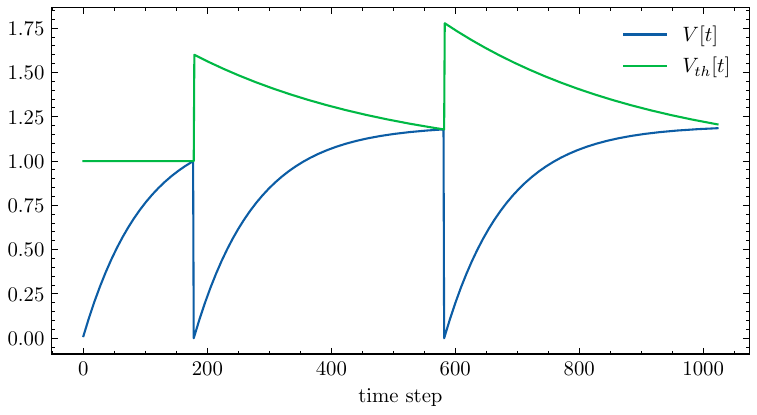}}
	\caption{The example of implementing the ALIF neuron. \textbf{a}. Source code. \textbf{b}. $V[t]$ and $V_{th}[t]$ of the ALIF neuron when given a constant input.}
	\label{figure: ALIF example}
\end{figure}
As Fig.~\ref{figure: ALIF example}\textbf{a} shows, we implement the ALIF neuron as follows. 
In \textbf{line 5}, we first inherit the base neuron.
In \textbf{line 6}, the constructor function is defined. Extra parameters including $\beta, \rho$ are put in front of arguments, while the parameters of the base class such as $V_{th}, V_{reset}$, the surrogate function, and the step mode are defined in \textit{*args, **kwargs}.
In \textbf{line 7}, we call the constructor function of the base class. After that, $V_{th}, V_{reset}$ will be initialized.
Thus, in \textbf{line 8}, we use the initial threshold as $B^{0}$.
In \textbf{lines 9-12}, we add $B[t], \beta, \rho, \tau_{m}$ to the member variables. $B[t]$ is a state and we register it as a memory variable, which we have introduced in the section \textit{Design of the Step Module}.
$\beta, \rho, \tau_{m}$ do not change with time steps, and we register them as buffers, which are concepts of PyTorch that are the non-learnable parameters of a module. Note that $\beta, \rho, \tau_{m}$ are all tensors rather than float values, and they can be element-wise, channel-wise, or layer-wise as long as their shapes are correctly set to satisfy the broadcast mechanism in PyTorch. For example, if this neuron layer is put after a convolutional layer whose output shape is $(N, C, H, W)$, then we can set the initial value of $\beta, \rho, \tau_{m}$ with the shape $(C, H, W), (C, 1, 1)$ or $(1)$ for element-wise, channel-wise or layer-wise neuronal dynamics. This is also the reason why we do not inherit from the LIF neuron from SpikingJelly, which defines a layer-wise $\tau_{m}$. Also, \textbf{lines 9-12} are a good example of how native LIF neurons in SpikingJelly can be modified to be element-wise or channel-wise with little effort.
In \textbf{lines 14-15}, we define the neuronal charging function according to Eq.~(\ref{eq discrete LIF neuronal charge}).
In \textbf{lines 17-21}, we define the neuronal firing function. The threshold is generated according to Eq.~(\ref{eq: ALIF threshold dynamics 1}). The firing can then be done by calling the firing function inherited from the base node. After firing, the dynamic offset $B[t]$ is also updated by $S[t]$ according to Eq.~(\ref{eq: ALIF threshold dynamics 2}). 
In the end, the implementation is completed in just 21 lines of code.

Fig.~\ref{figure: ALIF example}\textbf{b} shows $V[t]$ and $V_{th}[t]$ curves of the ALIF neuron when given a constant input. It can be found that $V_{th}[t]$ increases instantaneously after each firing and decreases exponentially, which demonstrates the adaptative threshold dynamic of the ALIF neuron that is able to avoid both too high and too low firing rates by adjusting the threshold.

\subsubsection*{Implementation of the Izhikevich Neuron}
The second example is the implementation of the Izhikevich neuron, whose neuronal charging function is
\begin{align}
	H[t] = V[t-1] + \frac{1}{\tau_{m}} \cdot (X[t] + a_{0} \cdot (V[t-1] - V_{rest}) \cdot (V[t-1] - V_{c}) - W[t-1]), \label{eq: Izhikevich charge}
\end{align}
where $\tau_{m}$ is the membrane time constant, $V_{rest}$ is the reset potential, $a_{0}$ is the excitability parameter, $V_{c}$ is the critial potential, and $W[t]$ is the recovery parameter. The Izhikevich neuron will update $W[t]$ after charging as  
\begin{align}
	W_{pre}[t] = W[t-1] + \frac{1}{\tau_{w}} \cdot (a \cdot (V[t-1] - V_{rest}) -  W[t-1]),\label{eq: Izhikevich adapt}
\end{align}
where $\tau_{w}$ is the recovery time constant, and $a$ is the recovery update parameter. We use $W_{pre}[t]$ to represent the recovery parameter after updating but before resetting. Similar to the membrane potential, the recovery parameter is reset after firing, which is 
\begin{align}
	W[t] = W_{pre}[t] + b \cdot S[t], \label{eq: Izhikevich w reset}
\end{align}
where $b$ is the recovery reset parameter.

\begin{figure}
	\centering
	\includegraphics[width=1.\textwidth,trim=100 400 90 70,clip]{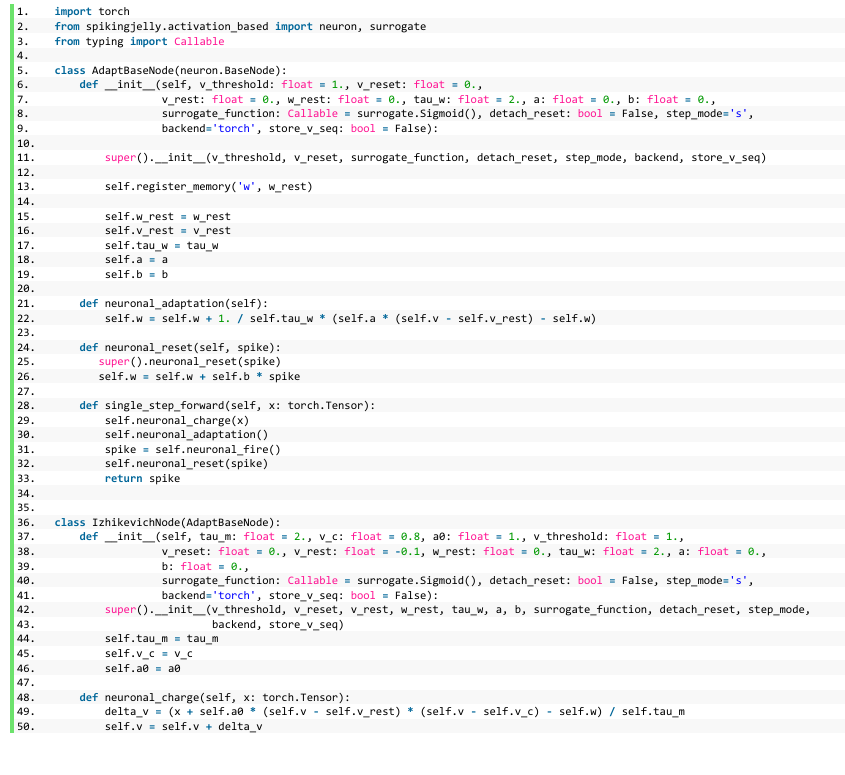}
	
	\caption{The code example of implementing the Izhikevich neuron.}
	\label{figure: Izhikevich example}
\end{figure}

Fig.~\ref{figure: Izhikevich example} shows the code example of implementing the Izhikevich neuron.
In \textbf{lines 5-33}, we first extend the base neuron to the adaptative base neuron. 
More specifically, in \textbf{lines 5-9}, we inherit the base neuron and add extra arguments including $\tau_{w}, W_{rest}, a, b$ to the constructor function.
In \textbf{line 13}, we set $W[t]$ as the memory state because it is a state variable, and its default value is $W_{rest}$.
In \textbf{lines 15-19}, we add these extra parameters as the member variables.
In \textbf{lines 21-22}, we add the neuronal adaptation function according to Eq.~(\ref{eq: Izhikevich adapt}).
In \textbf{lines 24-26}, we add the update of $W[t]$ in the neuronal resetting function.
In \textbf{lines 28-33}, we override the forward function with insert the neuronal adaptation function in \textbf{line 30} after the neuronal charging function.

Implementing the Izhikevich neuron does not require much effort after completing the adaptation of the base neuron. 
In \textbf{lines 36-46}, we inherit the adaptative base neuron and add extra parameters of the Izhikevich neuron including $\tau_{m}, V_{c}$ and $a_{0}$.
In \textbf{lines 48-50}, we define the neuronal charging function according to Eq.~(\ref{eq: Izhikevich w reset}).
Finally, the implementation was completed in just 50 lines of code.

\begin{figure}
	\centering
	\includegraphics[width=0.8\textwidth,trim=0 0 0 0,clip]{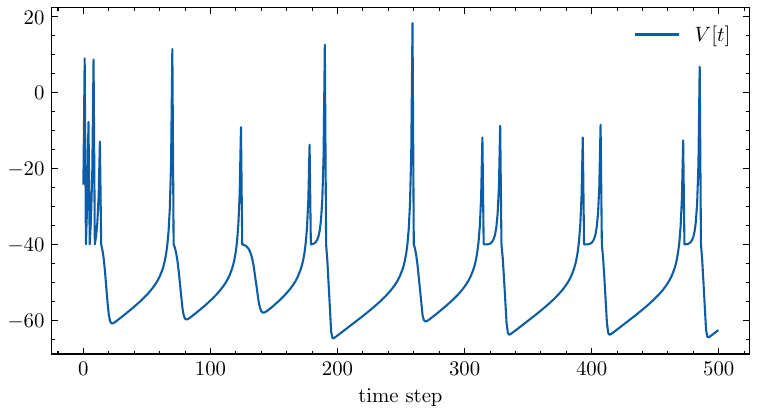}
	
	\caption{The chattering dynamic modeled by the Izhikevich neuron.}
	\label{figure: Izhikevich example curve}
\end{figure}

Fig.~\ref{figure: Izhikevich example curve} shows the chattering dynamic modeled by Izhikevich neurons, which resemble the vivo recordings of the pyramidal neuron of cat visual cortex. Chattering neurons are also known as fast rhythmic bursting (FRB) neurons, which generate high-frequency repetitive bursts in response to injected depolarizing currents. The hyper-parameters are $\tau_{m}=50, V_{c}=-40, a_{0}=1.5, V_{th}=25, V_{reset}=-40, W_{rest}=0, \tau_{w}=100, a=\frac{100}{3}, b=150$, and the input is constant $X[t]=800$.

\subsection*{Examples of Implementing New Learning Rules}
In this section, we will show how to use SpikingJelly to implement new learning rules, which also demonstrates the superior extensibility and flexibility of SpikingJelly.

\subsubsection*{Neuronal Dynamics}
\begin{figure}
	\centering
	\includegraphics[width=1.\textwidth,trim=100 500 90 70,clip]{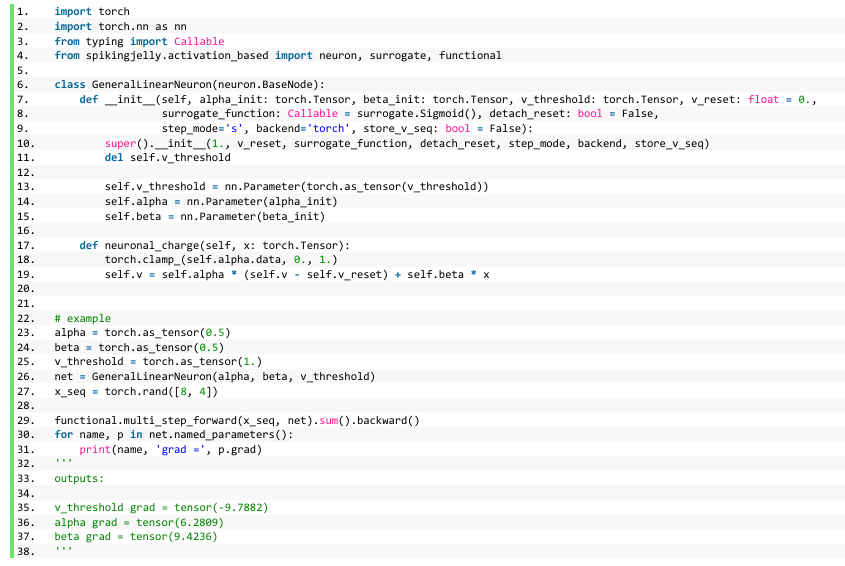}
	\caption{The example of implementing the general linear neuron with learnable neuronal dynamics.}
	\label{figure: learning rule example1}
\end{figure}

The learning methods incorporating learnable neuronal dynamics to enhance the performance of deep SNNs are frequently reported in spiking deep learning research \cite{fang2021incorporating, yao2022glif, ponghiran2022spiking}. Implementing spiking neurons with learnable parameters is straightforward in SpikingJelly. Suppose we want to implement a general linear neuron whose neuronal charging function is
\begin{align}
	H[t] = \alpha \cdot (V[t-1] - V_{reset}) + \beta \cdot X[t], \label{eq: general linear neuron neuronal charge}
\end{align} 
where $\alpha$ and $\beta$ are learnable parameters. We also want to set the threshold $V_{th}$ to be learnable. This neuron can be implemented as the codes shown in Fig.~\ref{figure: learning rule example1}.

In \textbf{lines 6-9}, we first inherit the base neuron and add extra arguments \textit{alpha\_init, beta\_init} as the initial value of $\alpha, \beta$. 
In \textbf{lines 10-11}, we call the constructor function of the base neuron, and then we delete the attribute \textit{v\_threshold} because its default data type is float, which can not be learnable.
In \textbf{lines 13-15}, we set three learnable parameters, which are $V_{th}, \alpha$ and $\beta$. These parameters are wrapped by \textit{torch.nn.Parameter}, then they can be optimized by the gradient descent automatically.
In \textbf{lines 17-19}, we define the neuronal charging function according to Eq.~(\ref{eq: general linear neuron neuronal charge}). We clamp $\alpha$ to be in the range $[0, 1]$ to make sure that the neuron can not charge itself without inputs.
In \textbf{lines 22-31}, we test this implementation. We give a random input sequence, apply a forward and a backward propagations, then check the parameters' gradients. 
In \textbf{lines 32-38}, the gradients are shown, indicating that this neuron can work normally with learnable $\alpha, \beta, V_{th}$.

\subsubsection*{Surrogate Functions}

\begin{figure}
	\centering
	\includegraphics[width=1.\textwidth,trim=100 476 90 70,clip]{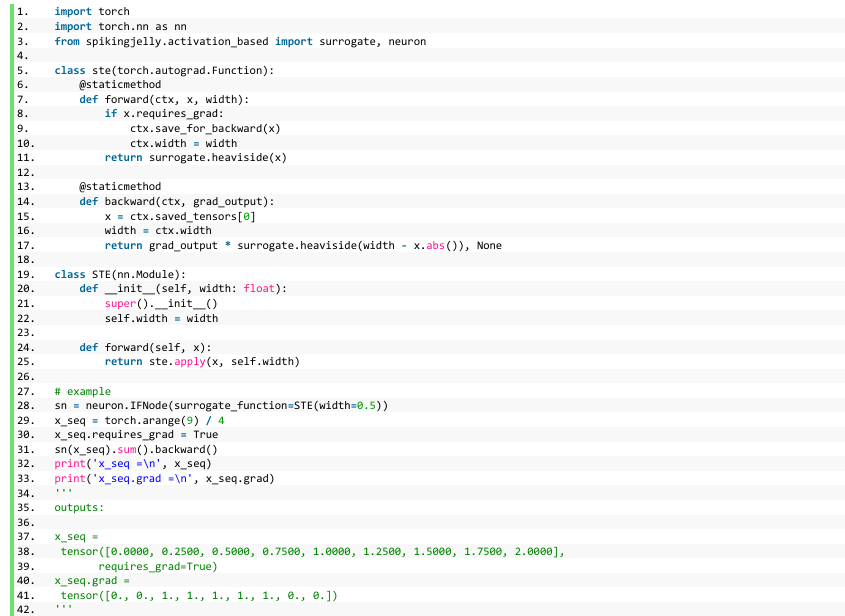}
	\caption{The example of implementing the custom Straight Through Estimator (STE) surrogate function.}
	\label{figure: learning rule example3}
\end{figure}

Fig.~\ref{figure: learning rule example3} shows the implementation of the Straight Through Estimator (STE) \cite{bengio2013estimating}, which is the most commonly used surrogate gradient function in quantized neural networks.

In \textbf{line 5-17}, we define the forward and the backward function of the STE as
\begin{align}
	y &= \Theta(x), \\
	\frac{\partial \mathcal{L}}{\partial x} &= \frac{\partial \mathcal{L}}{\partial y} \cdot \Theta(w - |x|), 
\end{align}
where $x$ is the input, $y$ is the output, $\mathcal{L}$ is the loss and $w$ is the width parameter.
In \textbf{line 19-25}, we wrap the STE surrogate function as a module.
In \textbf{line 28}, we create an IF neuron layer with the STE as the surrogate function.
In \textbf{line 29-33}, we give inputs with 9 elements to the IF neuron layer, apply the backward, and check the gradient of inputs.
Note that we only run one time step, and the gradient can be easily computed as $\frac{\partial \mathcal{L}}{\partial X_{i}} = \Theta(w - |X_{i} - V_{th}|)$, where $w=0.5, V_{th} = 1, X_{i} = \frac{i - 1}{4}$. 
The gradients are shown in \textbf{line 34-42}, which are identical to the calculation formula.
\subsubsection*{Gradient-based Optimizers}
We take the code of the Gradient Rewiring (Grad R) algorithm \cite{ijcai2021-236} as an example of how to implement the gradient-based optimizer in SpikingJelly. 

Grad R is a pruning algorithm that maintains a pretraining-free pipeline to minimize the additional computational burden. Grad R serves as a plug-and-play PyTorch optimizer in SpikingJelly, which is based on PyTorch's official implementation of Adam optimizer with minor modifications. Note that all modules in SpikingJelly follow the style of PyTorch, resulting in a simple replacement of the origin optimizer for researchers that would enable them to directly prune their SNNs during training.

\begin{figure}
	\centering
	\includegraphics[width=1.\textwidth,trim=100 689 90 70,clip]{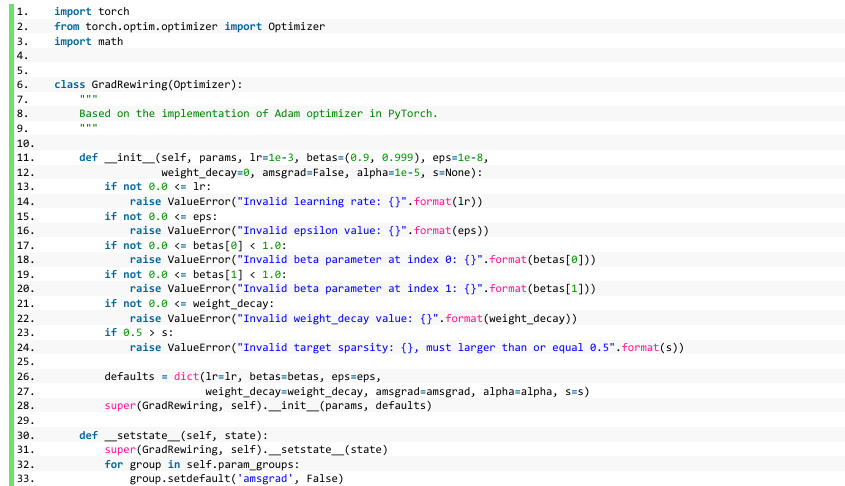}
	\caption{The parameter definition part of code example of implementing the Grad R optimizer.}
	\label{figure: learning rule example4-1}
\end{figure}

The code for Grad R optimization is depicted in Figs.~\ref{figure: learning rule example4-1}, \ref{figure: learning rule example4-2}.
Aside from basic hyperparameters for training such as learning rate and weight decay, the \textit{GradRewiring} class introduces a hyperparameter named $s$ as target sparsity and $\alpha$ as the penalty term of sparse prior, as shown in \textbf{lines 11-12} of Fig.~\ref{figure: learning rule example4-1}. These hyperparameters will be added to the state dictionary along with the other normal training hyperparameters in \textbf{lines 26-27} of Fig.~\ref{figure: learning rule example4-1}.

Fig.~\ref{figure: learning rule example4-2} shows the key parts of the pruning algorithm. \textbf{Line 37} starts a walkthrough of the network parameters. \textbf{Line 46} extracts all hyperparameters and some states of Adam. Since Grad R must keep the sign of each weight component, the initial sign $s$ will be recorded in \textbf{line 60} when initializing the optimizer (\textbf{lines 49-66}) and fixed in each training iteration in \textbf{line 68}. The strength $\theta$ is a learnable parameter that describes the magnitude of corresponding hidden weights during training. The relation between actual weight $W$, initial sign $s$, and hidden weight $\theta$ during inference is defined through a ReLU mapping in \textbf{line 108} in the following formulation
\begin{align}
	s & =\mathrm{Sign}(W[0]),\\
	W[k] & =s \cdot \max(0, \theta[k]),
\end{align}
where $W[k], \theta[k]$ are actual weight and hidden weight at $k$-th training iteration. 
Normally, the above equation implies the backpropagation path from the loss function to $W[k]$, and then to $\theta[k]$. Grad R defines this process in five parts:

1.	Compute the step size of a typical Adam update for sparse weight in \textbf{lines 71-94}.

2.	Generate dense weight in \textbf{line 96} using the initial sign $s$ and hidden weight $\theta$.

3.	Apply sparse prior to $\theta$ in \textbf{lines 102-103}.

4.	Generate updated hidden weight in \textbf{line 105}.

5.	Get sparse weight in \textbf{line 108} using ReLU mapping.

\noindent The aforementioned sparse prior of $\theta$ takes the form
\begin{align}
	p(\theta)=\frac{\alpha}{2}\exp(-\alpha|\theta -\mu|),
\end{align}
where $\mu$ is computed through target sparsity $s$ and $\alpha$ as 
\begin{align}
	\mu=\frac{1}{\alpha}\ln(2-2s).
\end{align}

\begin{figure}
	\centering
	\includegraphics[width=1.\textwidth,trim=100 164 90 310,clip]{./fig/grad_r.pdf}
	\caption{The core algorithm part of code example of implementing the Grad R optimizer.}
	\label{figure: learning rule example4-2}
\end{figure}

\subsubsection*{Local Learning Rules}
\begin{figure}
	\centering
	\includegraphics[width=1.\textwidth,trim=100 315 90 70,clip]{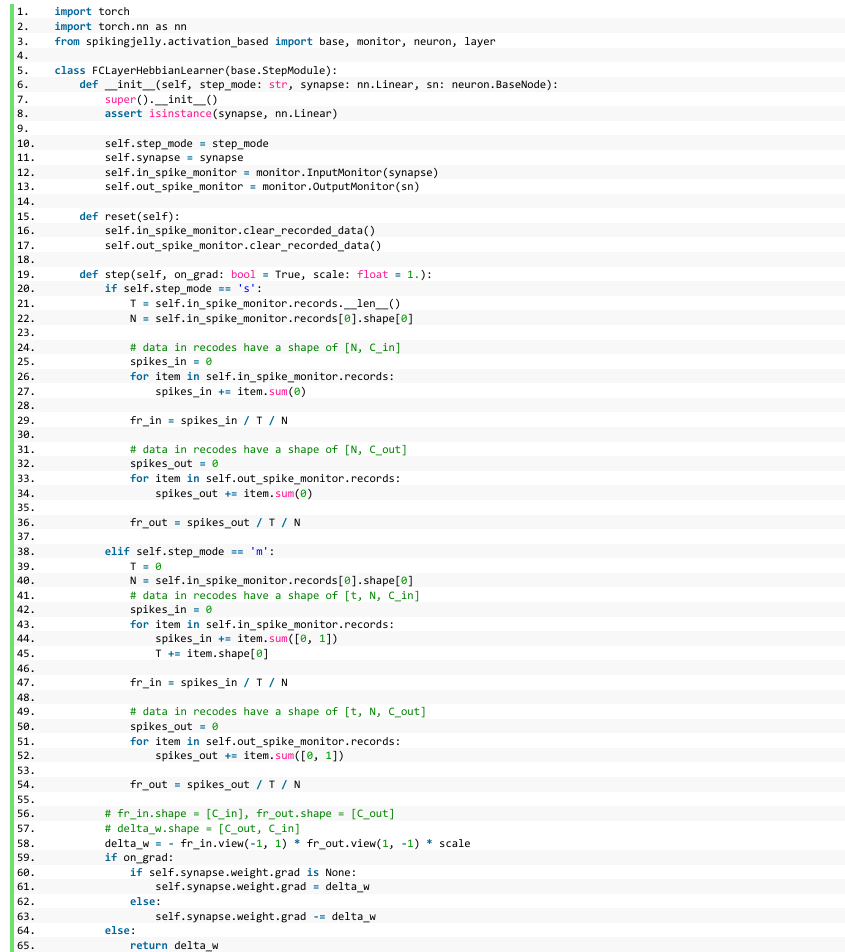}
	\caption{The example of implementing the Hebbian learner.}
	\label{figure: learning rule example2}
\end{figure}

Local learning rules such as STDP can be efficiently implemented with monitors in SpikingJelly. For simplicity, we take the implementation of the Hebbian learning rule for the fully connected layer as the example, which is shown in Fig.~\ref{figure: learning rule example2}.

In \textbf{lines 5-7}, we define the constructor function, whose arguments are the step mode, the synapse, and the spiking neuron.
In \textbf{line 8}, we check the synapse to make sure that it is the fully connected layer.
In \textbf{lines 10-13}, we set the attributes and monitors, which are used to record input spikes and output spikes.
In \textbf{lines 15-17}, we define the reset function, which clears recorded data in monitors.
In \textbf{line 19}, we define the \textit{step} function, which computes $\Delta W$ of the weight $W$.
We compute the firing rates of input spikes and output spikes for the single-step mode in \textbf{lines 20-36} and for the multistep mode in \textbf{lines 38-54}.
In \textbf{lines 56-58}, we generate $\Delta W$ as 
\begin{align}
	\Delta W[i][j] = scale \cdot fr_{pre}[i] \cdot fr_{post}[j],
\end{align}
where $fr_{pre}[i]$ is the firing rate of the $i$-th input spikes, $fr_{post}[j]$ is the firing rate of the $j$-th output spikes, and $scale$ is the scale factor. This generation is efficiently implemented by the broadcast mechanism.
In \textbf{lines 59-65}, we add $\Delta W$ on the gradient of the weight of the fully connected layer when \textit{on\_grad} is true. Otherwise, this function returns $\Delta W$. 

Fig.~\ref{figure: learning rule example2 useage} shows the usage of the Hebbian learner implemented in Fig.~\ref{figure: learning rule example2}. We set $V_{th}=0.1$ for easy firing of the output neurons in \textbf{line 74}. 
The Hebbian learner is not based on autograd mechanism and we disable autograd in \textbf{line 77}.
In \textbf{lines 78-79}, we give random input spikes to the SNN. Note that the Hebbian learner has been recording data during the forward propagation of the SNN.
In \textbf{line 81}, we call the \textit{step} function to get and print $\Delta W$, which is shown in \textbf{lines 85-88}.

\begin{figure}
	\centering
	\includegraphics[width=1.\textwidth,trim=100 148 90 535,clip]{./fig/learning_rule_example2.pdf}
	\caption{The useage of the Hebbian learner implemented in Fig.~\ref{figure: learning rule example2}.}
	\label{figure: learning rule example2 useage}
\end{figure}

\subsection*{Ablation Study of Acceleration Methods}
\begin{figure}
	\centering
	\includegraphics[width=1.\textwidth,trim=30 500 576 20,clip]{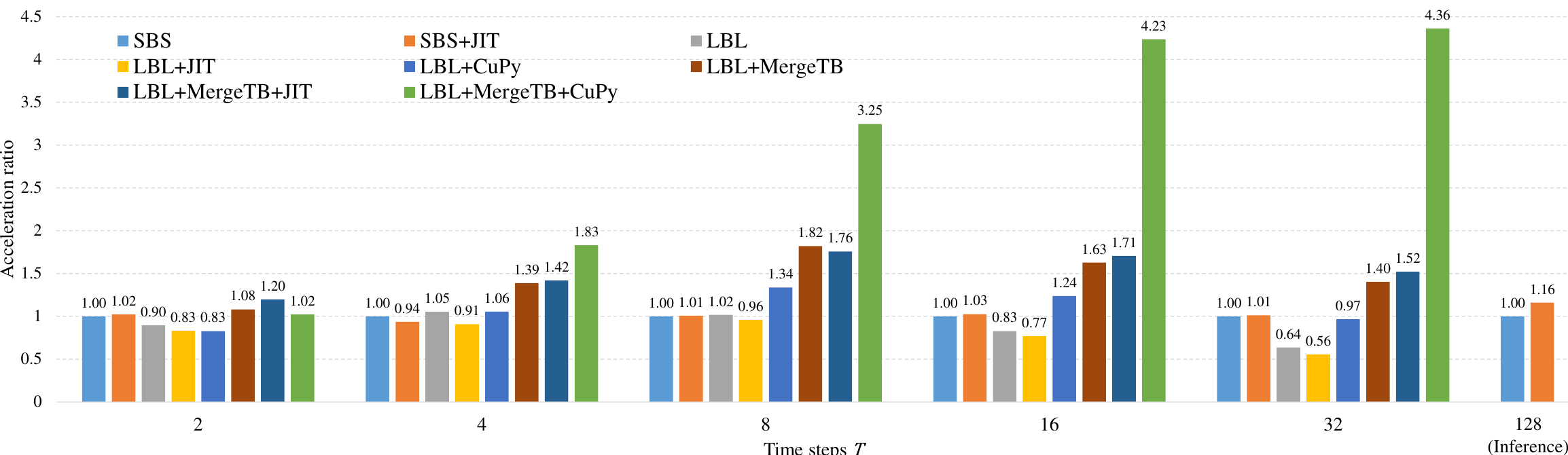}
	\caption{\textbf{Acceleration ratios of using different methods to the naive implementation (``SBS'') as the baseline.} ``SBS'' is using the single-step mode resulting in the step-by-step propagation pattern, ``LBL'' is using the multistep mode resulting in the layer-by-layer propagation pattern, ``CuPy'' is using the CuPy backend for spiking neurons, ``JIT'' is using the JIT for spiking neurons, and ``MergeTB'' is merging the time step dimension to the batch dimension for the stateless layers. Note the large time steps inference ($T=128$) only use the single-step mode, which is a typical usage of ANN2SNN.
	}
	\label{figure: acceleration ablation}
\end{figure}

To illustrate the effects of each acceleration method in SpikingJelly, we design ablation experiments on the Spiking ResNet-18 with $T=2, 4, 8, 16, 32$ for surrogate training and $T=128$ for inference, which is also identical to the experiment option in Fig.~\ref{figure: framework}\textbf{d}. We summarize the experiment results shown in Fig.~\ref{figure: acceleration ablation} as follows:

\begin{itemize}
	\item The acceleration of JIT is obvious in inference for the step-by-step pattern. But its effect varies from training to training. Its acceleration ratio is slight for the step-by-step pattern and even slows down the speed in the layer-by-layer pattern. However, when the layer-by-layer pattern enables the merging of the time step dimension to the batch dimension for the stateless layers, the JIT can accelerate by about 10\%.
	
	\item ``MergeTB'' and ``CuPy'' are critical for the layer-by-layer pattern. They accelerate the stateless and stateful layers, respectively. Using ``CuPy'' is slower than using ``JIT'' only when $T$ is small and the calling overhead of tiny kernels is minor.
	
	\item Using both ``MergeTB'' and ``CuPy'' in the layer-by-layer pattern can get the fastest simulation speed for training in most cases. 
	
\end{itemize}

In SpikingJelly, by setting all modules in the multistep mode and using the CuPy backend, ``LBL + MergeTB + CuPy'' is enabled and the training efficiency is maximized.

\subsection*{Comparison of SNN Frameworks}

\subsubsection*{Overview of Frameworks}
Tab.~\ref{tab: comparison between frameworks} shows a neat comparison between SNN frameworks, which can be regarded as a supplement of Fig.~\ref{figure: framework}\textbf{e}.

\textbf{Module Complexity} is high in NEURON, NEST, and Brian2, which aim to simulate actual biological neural systems in neuroscience. For example, neurons in NEURON are described by numerous parameters such as somatic properties, morphological properties, and ion channels. Neuronal dynamics also involve many ODEs.
Nengo and BindsNet achieve a balance between neuroscience and computer science with models of intermediate complexity. While SpikingJelly, Norse, and SNNTorch use neurons and synapses with low complexity. For example, the refractory period of LIF neurons  is considered in Nengo, but dismissed in SpikingJelly due to the fast simulation speed and high accuracy during training.

\textbf{GPU Support} is determined by the \textbf{GPU backend}. GPUs were not the lead role of high-performance computing (HPC) when the traditional frameworks including NEURON, NEST, Brian2, and Nengo were proposed. As a result, these frameworks do not natively support GPUs. New subpackages are developed, which are shown in \textbf{GPU backend} in Tab.~\ref{tab: comparison between frameworks}, to enhance the parent frameworks. However, these subpackages can not override all the functions of the parent frameworks, resulting in the imperfect support for GPUs.
BindsNet, SpikingJelly, Norse and SNNTorch are based on PyTorch. Almost all modules and functions in PyTorch are GPU-compatible, and distributed training on multiple GPUs is well supported. As a result, these frameworks can fully exploit the GPU to speed up the simulation. Moreover, SpikingJelly introduces the CuPy backend, which uses CUDA-level acceleration and is much more efficient than naive PyTorch-based modules.

\textbf{Neuromorphic Support} may not be the main scope of some frameworks. SpikingJelly integrates a full-stack solution for processing neuromorphic datasets and provides nine datasets. Norse integrates only one dataset, while SNNTorch provides three.
Only Brian2, Nengo, and SpikingJelly offer subpackages for neuromorphic chips. Remarkably, NengoLoihi is the most developed package for Loihi, giving Nengo the best compatibility with Loihi.

\textbf{Synaptic Plasticity} refers to biologically plausible learning rules such as STDP to modify the synapse weights, which are the strengths of Brian2, Nengo, and BindsNet. Note that although NEURON does not have built-in plasticity, it is easy to implement these rules by NEURON, as suggested by the moderators in the NEURON forum. In deep SNNs, gradient-based or conversion optimization methods are the main roles to modify synapse weights. Thus, synaptic plasticity is often neglected in spiking deep learning frameworks, of which only SpikingJelly implements biologically plausible learning rules.

\textbf{Deep Learning} is the main concern of SpikingJelly, Norse, and SNNTorch. All of them support surrogate learning methods, and SpikingJelly also integrates ANN2SNN methods. It is worth noting that Nengo also supports ANN2SNN through the subpackage NengoDL. However, it approximates ReLU in ANNs using the response of LIF neurons, which may not be compatible with most ANN2SNN studies that use IF neurons. 

\textbf{Forum} includes the number of posts in the forum (if the framework has) and issues as well as pull requests in GitHub. \textbf{Publication} is the number of academic publications. \textbf{Forum} and \textbf{Publication} represent the number of users and academic researchers respectively, which shows the activity of the framework community. As \textbf{Release Date} shows, classical frameworks including NEURON, NEST, Brian2, and Nengo have been developing for many years and enjoy great reputations with a large scale community. BindsNet, SpikingJelly, Norse, and SNNTorch are young generations with smaller community sizes than the classical frameworks. Fortunately, their community is growing rapidly due to the growing interest in spiking machine learning. For example, although SpikingJelly was launched in 2019, it accumulated 94+ publications in three years, which is even close to the number of publications in the classical framework.

\begin{table}[]
	\scalebox{0.5}{
		\begin{tabular}{c|c|c|c|c|c|c|c|c|c|}
			\cline{2-10}
			& Module Complexity & GPU Support & GPU Backend                                                         & Neuromorphic Support                                                                          & Synaptic Plasticity & Deep Learning                                                              & Forum                                                                       & Publication & Release Date\\ \hline
			\multicolumn{1}{|c|}{NEURON}       & $\bigstar$$\bigstar$$\bigstar$$\bigstar$$\bigstar$                     & $\bigstar$$\bigstar$          & CoreNEURON                                                          &                                                                                               &                     &                                                                            & 18000+  posts                                                                & 2672+       & 1984 \\ \hline
			\multicolumn{1}{|c|}{NEST}         & $\bigstar$$\bigstar$$\bigstar$                       & $\bigstar$           & NEST GPU                                                            &                                                                                               & $\bigstar$$\bigstar$$\bigstar$                 &                                                                            & 1200+ issues/prs                                                            & 668+          & 1994 \\ \hline
			\multicolumn{1}{|c|}{Brian2}       & $\bigstar$$\bigstar$$\bigstar$$\bigstar$                      & $\bigstar$$\bigstar$          & Brian2GENN                                                          & Brian2Loihi                                                                                   & $\bigstar$$\bigstar$$\bigstar$                 &                                                                            &\makecell[c]{3000+ posts\\800+ issues/prs} & 100+         & 2013 \\ \hline
			\multicolumn{1}{|c|}{Nengo}        & $\bigstar$$\bigstar$                        & $\bigstar$$\bigstar$$\bigstar$         & \makecell[c]{NengoOCL\\ TensorFlow} & NengoLoihi                                                                                    & $\bigstar$$\bigstar$                  & ANN2SNN                                                                    & \makecell[c]{7200+ posts\\800+ issues/prs} & 100+        & 2003 \\ \hline
			\multicolumn{1}{|c|}{BindsNet}     & $\bigstar$$\bigstar$                        & $\bigstar$$\bigstar$$\bigstar$$\bigstar$        & PyTorch                                                             &                                                                                               & $\bigstar$$\bigstar$                  &                                                                            & 630+ issues/prs                                                             & 90+          & 2018 \\ \hline
			\multicolumn{1}{|c|}{SpikingJelly} & $\bigstar$                         & $\bigstar$$\bigstar$$\bigstar$$\bigstar$$\bigstar$       & \makecell[c]{PyTorch\\CuPy}        & \makecell[c]{9 datasets\\    LavaExchange\\  LynxiExchange} & $\bigstar$                   & \makecell[c]{Surrogate Learning\\ANN2SNN}& 390+ issues/prs                                                             & 94+          & 2019 \\ \hline
			\multicolumn{1}{|c|}{Norse}        & $\bigstar$                         & $\bigstar$$\bigstar$$\bigstar$$\bigstar$        & PyTorch                                                             &                                                                                               1 dataset &                     & Surrogate Learning                                                         & 360+ issues/prs                                                             & 3+          & 2019 \\ \hline
			\multicolumn{1}{|c|}{SNNTorch}     & $\bigstar$                         & $\bigstar$$\bigstar$$\bigstar$$\bigstar$        & PyTorch                                                             &                                                                                               3 datasets &                     & Surrogate Learning                                                         & 170+ issues/prs                                                             & 45+          & 2020\\ \hline
		\end{tabular}
	}
	\caption{Comparison between SNN frameworks}
	\label{tab: comparison between frameworks}
\end{table}

\subsubsection*{Difference between SpikingJelly/Norse/SNNTorch}
In addition to the comparisons in Tab.~\ref{tab: comparison between frameworks}, there are still some delicate diversities between three deep learning frameworks.

\textbf{States} SpikingJelly stores states such as the membrane potential $V[t]$ inside the module. A module is a stateful module if it has states, or a stateless module otherwise. For example, the LIF neuron is a stateful module and the linear layer is a stateless module in SpikingJelly. Instead, Norse considers the states as part of the input/output of the module. SNNTorch provides an option to control whether to store states or not. Storing states inside a module brings convenience for building and training SNNs, since the states are managed by the module itself. From an external perspective, these stateful modules only require inputs from the last layer and send outputs to the next layer, which can work with stateless modules easily, e.g., they can be wrapped into \textit{torch.nn.Sequential} in PyTorch to build deep SNNs. In Norse, modules whose states are not stored inside require an extra wrapper \textit{norse.torch.SequentialState}.  The only additional operation for a stateful module with stored states inside itself is that they need to be reset before the next simulation with a new input, which can be easily done by applying a reset function on the whole network. 

\textbf{Step Mode} Modules in SpikingJelly can be switched between the single-step and the multistep, which process data at a single time step or many time steps. Based on step modes, the step-by-step propagation and the layer-by-layer propagation patterns are derived, which can be used flexibly for different cases. In Norse and SNNTorch, the concepts of step modes and propagation patterns are not explicitly defined. Most modules in Norse use the multistep mode, while few modules use the single-step mode. The wrapper \textit{Lift} is provided in Norse to wrap single-step modules such as stateless convolutional layers from PyTorch to help them work in the multistep mode. \textit{Lift} in Norse has the same behavior as \textit{MultiStepContainer} in SpikingJelly, which is used to wrap a single-step stateful module to work in the multistep mode.
While SpikingJelly uses \textit{SeqToANNContainer} to wrap stateless layers with merging time step and batch dimensions, which is much faster than \textit{Lift} and \textit{MultiStepContainer}. The default propagation pattern in Norse is layer-by-layer, but can also be step-by-step when using modules in the single-step mode. In SNNTorch, all modules work in the single-step mode and the propagation pattern is step-by-step.

\subsection*{Details of the Neuron Kernel}
\begin{figure}
	\raggedright
	\subfloat[]{\includegraphics[width=0.42\textwidth,trim=5 590 820 770,clip]{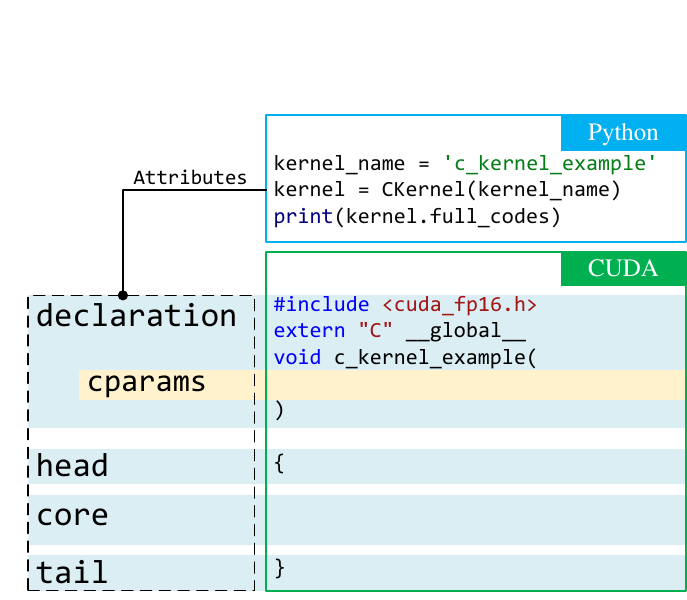}}
	\subfloat[]{\includegraphics[width=0.42\textwidth,trim=345 590 480 770,clip]{./fig/ckernel.pdf}}
	
	\subfloat[]{\includegraphics[width=0.75\textwidth,trim=5 180 600 1060,clip]{./fig/ckernel.pdf}}
	\caption{\textbf{CUDA kernels generated by Python classes \textit{CKernel}, \textit{CKernel1D} and \textit{CKernel2D}.}
		\textbf{a}. \textit{CKernel} has attributes \textit{declaration}, \textit{head}, \textit{core}, and \textit{tail}. \textit{declaration} is generated from the attribute \textit{cparams}, which is an empty Python dictionary in \textit{CKernel} by default.
		\textbf{b}. \textit{CKernel1D} is extended from \textit{CKernel} for processing 1D tensors with elementwise operations. The default values in \textit{cparams} are \textit{numel}, which is the number of elements.
		\textbf{c}. based on \textit{CKernel}, \textit{CKernel2D} is designed for processing 2D tensors with an extra for-loop. The default values in \textit{cparams} are \textit{numel}, which is the number of all elements, and \textit{N}, which is the number of elements at one time step.
		For more delicate coding control, the extra attributes \textit{pre\_core} and \textit{post\_core} are introduced to add CUDA codes before and after \textit{core}, respectively.
	}
	
	\label{figure: ckernel}
\end{figure}

\begin{figure}
	\raggedright
	\includegraphics[width=1\textwidth,trim=2 440 220 20,clip]{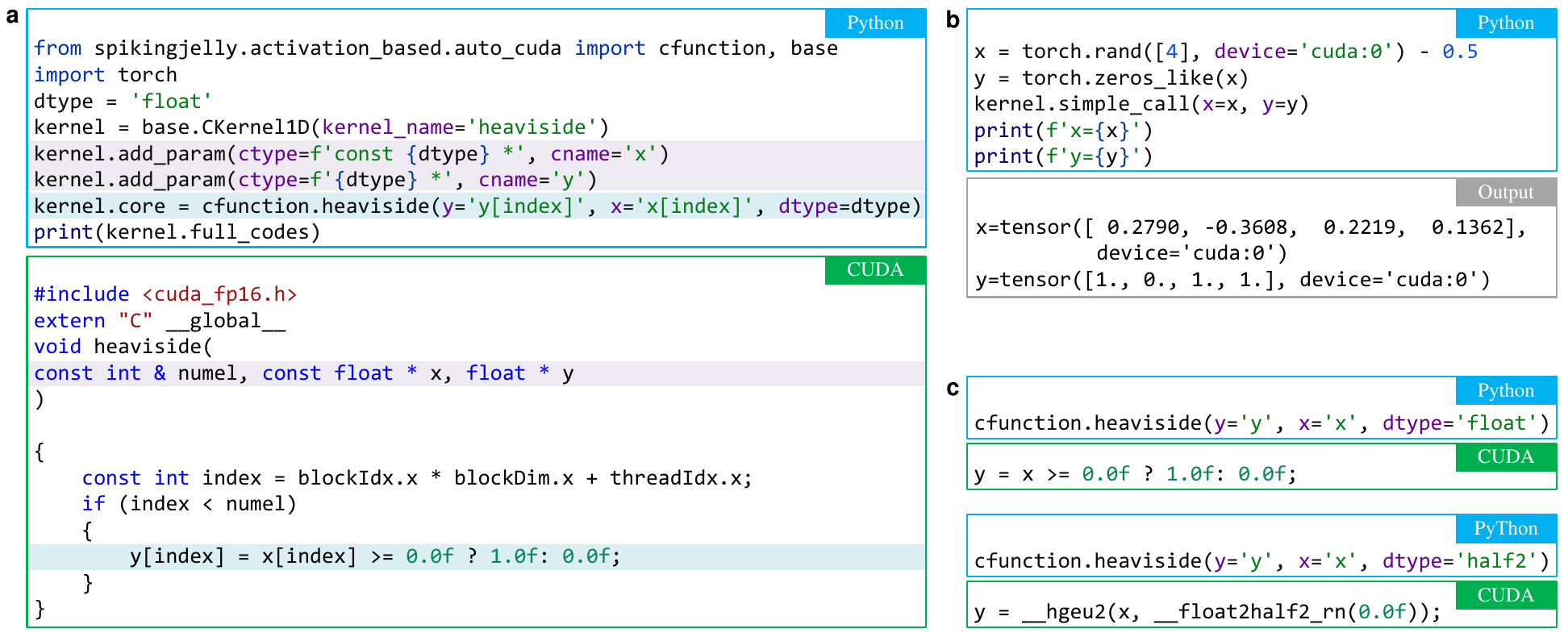}

	\caption{\textbf{Examples that utilize \textit{CKernel1D}.}
		\textbf{a}. The Python code implementing the Heaviside function and the resulting CUDA code. The user only needs to define the function arguments via \textit{add\_function} and the core codes by setting the \textit{core} attribute. These Python codes are colored, and they control the corresponding CUDA codes with the same color. 
		\textbf{b}. An example of executing a CUDA kernel from \textbf{a} in a Python environment.
		\textbf{c}. An example of \textit{cfunction.heaviside} with different data types and their generated CUDA codes. \textit{cfunction} provides Python functions to generate CUDA codes that support both the \textit{float} and \textit{half2} data types. By switching the \textit{dtype}, the user can easily obtain CUDA codes for different data types.
	}
	
	\label{figure: ckernel1d example}
\end{figure}

\begin{algorithm}
	\footnotesize
	\caption{BPTT of the IF neuron}
	
	\textbf{Require:} 
	
	~~~~ the batch size $N$ \tcp{N}
	
	~~~~ the number of time steps $T$ \tcp{numel = $N \cdot T$}
	
	~~~~ the gradients of output spikes at all time steps $\{\frac{\partial \mathcal{L}}{\partial S[t]}\}, t=0,1,...,T-1$ \tcp{grad\_spike\_seq}
	
	~~~~ the gradient of membrane potential after resetting at the last time step $\frac{\partial \mathcal{L}}{\partial V[T-1]}$  \tcp{grad\_v\_seq}
	
	~~~~ the membrane potentials after charging at all time steps $\{H[t]\}, t=0,1,...,T-1$  \tcp{h\_seq}
	
	~~~~ the threshold potential $V_{th}$ \tcp{v\_th}
	
	~~~~ the reset potential $V_{reset}$ \tcp{v\_reset}
	
	\textbf{Outputs:} 
	
	~~~~ the gradient of inputs at all time steps $\{\frac{\mathrm{d} \mathcal{L}}{\mathrm{d} X[t]}\}, t=0, 1,..., T-1$ \tcp{grad\_x\_seq}
	
	~~~~ the gradient of the initial membrane potential after resetting $\frac{\mathrm{d} \mathcal{L}}{\mathrm{d} V[-1]}$ \tcp{grad\_v\_init}

	$\frac{\mathrm{d} \mathcal{L}}{\mathrm{d} H[T]} = 0$ \tcp{grad\_h = 0}

	\algorithmicfor{~$t \gets T-1, T-2, ..., 0$}
	
	~~~~~~ $\frac{\mathrm{d} H[t+1]}{\mathrm{d} V[t]} = 1$ \tcp{grad\_h\_next\_to\_v}
	
	~~~~~~ $\frac{\mathrm{d} H[t]}{\mathrm{d} X[t]} = 1$ \tcp{grad\_h\_to\_x}
	
	~~~~~~ $\frac{\mathrm{d} S[t]}{\mathrm{d} H[t]} = \sigma'(H[t] - V_{th})$ \tcp{surrogate gradient}
	
	~~~~~~ $\frac{\mathrm{d} V[t]}{\mathrm{d} H[t]} = 1 - S[t] + (-H[t] + V_{reset}) \cdot \frac{\mathrm{d} S[t]}{\mathrm{d} H[t]}$
	
	~~~~~~ $\frac{\mathrm{d} \mathcal{L}}{\mathrm{d} H[t]} = \frac{\partial \mathcal{L}}{\partial S[t]} \cdot \frac{\mathrm{d} S[t]}{\mathrm{d} H[t]} + (\frac{\partial \mathcal{L}}{\partial V[t]} + \frac{\mathrm{d} \mathcal{L}}{\mathrm{d} H[t+1]} \cdot \frac{\mathrm{d} H[t+1]}{\mathrm{d} V[t]}) \cdot \frac{\mathrm{d} V[t]}{\mathrm{d} H[t]}$
	
	~~~~~~ $\frac{\mathrm{d} \mathcal{L}}{\mathrm{d} X[t]} = \frac{\mathrm{d} \mathcal{L}}{\mathrm{d} H[t]} \cdot \frac{\mathrm{d} H[t]}{\mathrm{d} X[t]}$

	$\frac{\mathrm{d} \mathcal{L}}{\mathrm{d} V[-1]} = \frac{\mathrm{d} \mathcal{L}}{\mathrm{d} H[0]} \cdot \frac{\mathrm{d} H[0]}{\mathrm{d} V[-1]}$
	\label{alg: IF BPTT}
\end{algorithm}

\begin{figure}
	\centering
	\includegraphics[width=1.\textwidth,trim=26 40 280 40,clip]{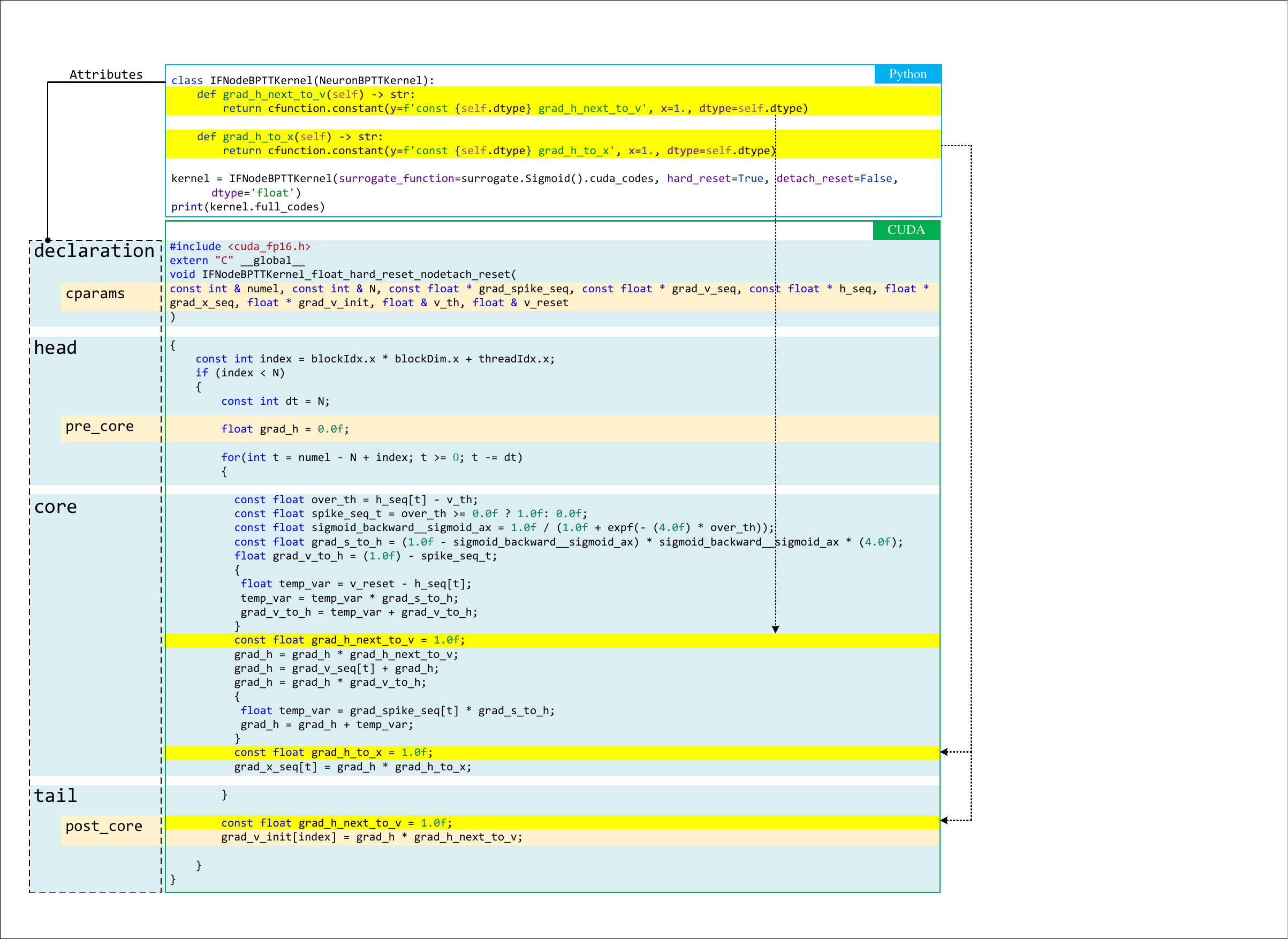}
	\caption{\textbf{Python codes of an IF neuron for the backward CUDA kernel class \textit{IFNodeBPTTKernel} and the generated CUDA codes.} These codes implement the BPTT algorithm of the IF neuron described in Alg.~\ref{alg: IF BPTT}.
		Based on \textit{NeuronBPTTKernel}, \textit{IFNodeBPTTKernel} only needs to complete $\frac{\mathrm{d} H[t+1]}{\mathrm{d} V[t]}$ and $\frac{\mathrm{d} H[t]}{\mathrm{d} X[t]}$ via the \textit{grad\_h\_next\_to\_v} and \textit{grad\_h\_to\_v} functions, respectively, which are highlighted in yellow in the Python codes. The corresponding CUDA codes governed by these Python codes are also highlighted in yellow.
	}
	\label{figure: auto cuda kernel}
\end{figure}

The semiautomatic CUDA code generation method is a practical technique for SpikingJelly. This technique is based on CuPy, which provides access to low-level CUDA features and is a NumPy/SciPy-compatible array library with GPU acceleration. CuPy is able to compile and run CUDA kernels with CUDA codes generated from Python strings. This feature is fully exploited by SpikingJelly, which generates different CUDA codes via if-else statements in Python, which modularizes the writing of CUDA codes and reduces development costs. The array library compatibility means that tensors in Numpy-format libraries including PyTorch are able to call CUDA kernels compiled with CuPy without copying or moving data between PyTorch, CuPy, and CUDA arrays. In SpikingJelly, tensors on GPUs are first set contiguous in memory, and then the addresses of the first element of these tensors are regarded as pointers and passed to the CUDA kernel. Finally, CUDA kernels can be directly executed by treating these tensors as arrays in CUDA. The workflow for the standard PyTorch CUDA extension requires preparing the CUDA development environment, writing CUDA code, writing C++ wrappers, setting up and loading C++ and CUDA files. It can be found that, due to the use of CuPy in SpikingJelly, only CUDA code need to be written and other procedures can be discarded. Moreover, the writing is accelerated by the semiautomatic CUDA code generation technique. As a result, the cost of developing and using CUDA extensions is largely reduced.

The base kernel class \textit{CKernel} in SpikingJelly has attributes such as its \textit{declaration}, \textit{head}, \textit{core}, and \textit{tail}, which are in string format and used to joint the complete CUDA codes in a split matter. \textit{declaration} is generated from the \textit{cparams} attribute, which is an empty Python dictionary in \textit{CKernel}. Fig.~\ref{figure: ckernel}\textbf{a} shows the CUDA codes generated by \textit{CKernel}. \textit{CKernel1D} is extended from \textit{CKernel} for processing 1-D tensors with elementwise operations. The default value in \textit{cparams} is \textit{numel}, which is the number of elements. Fig.~\ref{figure: ckernel}\textbf{b} shows an example of \textit{CKernel1D} and its generated CUDA codes. To process sequence data with a shape of $(T, N)$, \textit{CKernel2D} is designed with an extra for-loop, as shown in Fig.~\ref{figure: ckernel}\textbf{c}. The default values in \textit{cparams} are \textit{numel}, which is the number of all elements, and \textit{N}, which is the number of elements at one time step. For more delicate coding control, the extra attributes \textit{pre\_core} and \textit{post\_core} are introduced to add CUDA codes before and after \textit{core}, respectively. For example, the temporary variables used inside the for-loop can be initialized in \textit{pre\_core}.

Fig.~\ref{figure: ckernel1d example} shows an example of using \textit{CKernel1D}. To implement an elementwise CUDA kernel, the user only needs to add the function arguments, e.g., the input and output tensors, by calling the Python function \textit{add\_param} and the core CUDA codes by setting the attribute \textit{core}. In \textit{CKernel1D}, the CUDA thread index and the element index are identical, which is the \textit{index} in CUDA codes. Thus, the user can utilize \textit{[index]} to implement the elementwise operations. Fig.~\ref{figure: ckernel1d example}\textbf{a} shows the Python codes required to implement the Heaviside function and the generated CUDA codes. The Python codes for adding function arguments and setting CUDA core codes are colored and the corresponding CUDA codes with the same color are controlled. Fig.~\ref{figure: ckernel1d example}\textbf{b} shows how to execute this kernel. We find that the execution process is implemented in a pure Python environment, reducing the amount of work required to bind C++/CUDA to Python. Fig.~\ref{figure: ckernel1d example}\textbf{c} shows an example of \textit{cfunction.heaviside} with different data types and the generated CUDA codes. \textit{cfunction} is a subpackage of SpikingJelly that provides functions to generate CUDA codes that support both the \textit{float} and \textit{half2} data types. By switching the \textit{dtype}, the user can easily obtain CUDA codes for different data types. In this way, the user does not need to write kernels for both data types.

Compared with \textit{CKernel1D}, \textit{CKernel2D} adds a for-loop concerning the time steps, while the other components are similar. Based on \textit{CKernel2D}, the kernel for a spiking neuron's FPTT/BPTT can be easily implemented. Fig.~\ref{figure: auto cuda kernel} takes an IF neuron as the example and illustrates the details of how the BPTT algorithm described in Alg.~\ref{alg: IF BPTT} is implemented by the backward kernel \textit{IFNodeBPTTKernel}. The base class \textit{NeuronBPTTKernel} is inherited from \textit{CKernel2D} and has defined CUDA codes, including its \textit{declaration}, \textit{head}, \textit{pre\_core}, \textit{core}, \textit{post\_core} and \textit{tail}. As the yellow highlights for the CUDA codes in Fig.~\ref{figure: auto cuda kernel} show, three placeholders in \textit{core} and \textit{post\_core} are reserved for defining $\frac{\mathrm{d} H[t+1]}{\mathrm{d} V[t]}$ and $\frac{\mathrm{d} H[t]}{\mathrm{d} X[t]}$. These two equations are completed by the functions \textit{grad\_h\_next\_to\_v} and \textit{grad\_h\_to\_v} in the child class, as shown by the yellow highlights for the Python codes in Fig.~\ref{figure: auto cuda kernel}.

\subsection*{Exchange Modules}
Running SNNs on neuromorphic chips is a resource-constrained task; e.g., the typical precision levels are 16 bits in Lynxi KA200 and 8 bits in Loihi. In most cases, the neuromorphic chips have their own toolchains for deployment, such as the lyngor for the Lynxi chips and the Lava and the NxSDK for the Loihi. To be compatible with these toolchains, SpikingJelly defines the exchange modules, which are modules that support running in SpikingJelly and converting to the target toolchain. For each toolchain of a chip, an exchange package is defined, e.g., the \textit{lynxi\_exhange} package for the Lynxi KA200 and the \textit{lava\_exchange} package for the Loihi. The exchange module has the same behavior as the target module in the target toolchain. For example, the \textit{CubaLIFNode} in \textit{lava\_exchange} corresponds to the \textit{slayer.neuron.cuba.Neuron} in the Lava-DL framework. The identical behavior ensures that an SNN built by exchange modules in SpikingJelly can be converted to the target toolchain losslessly, and then be deployed to the target chip.

\begin{figure}
	\centering
	\includegraphics[width=1.\textwidth,trim=14 400 50 20,clip]{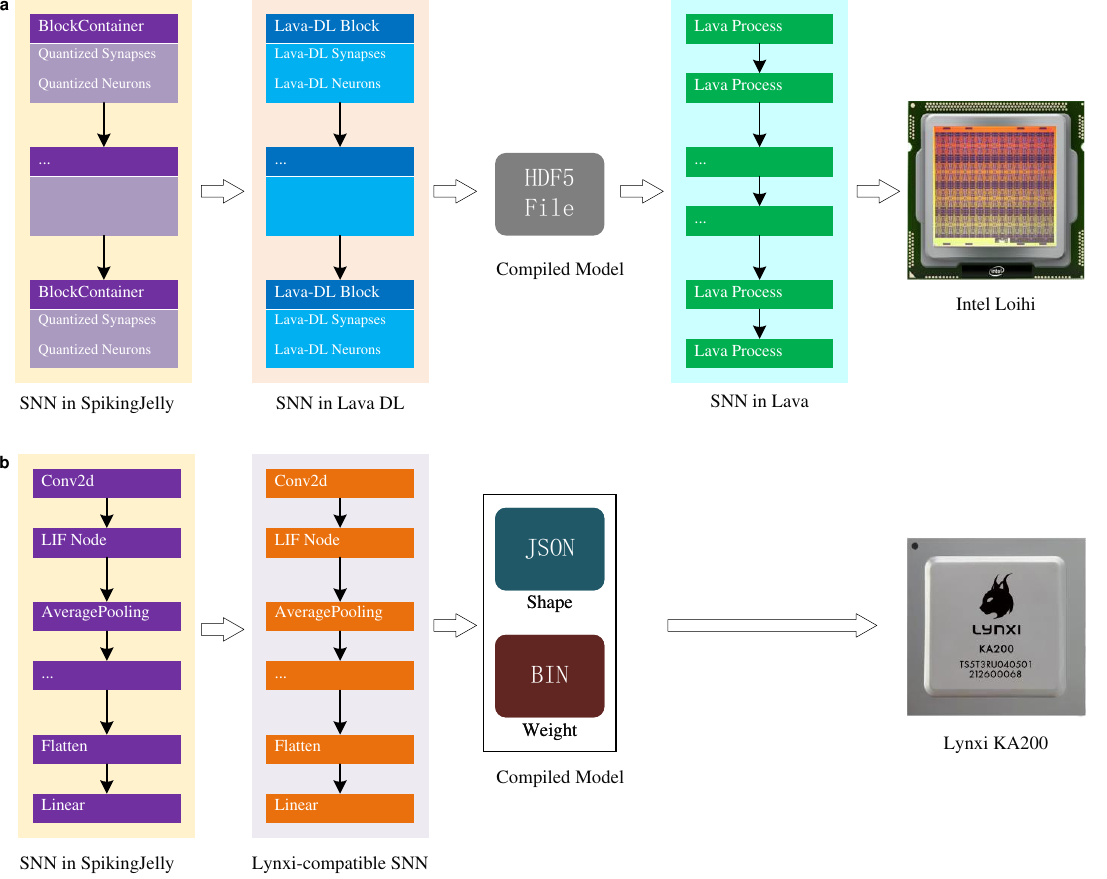}
	\caption{\textbf{Workflow for deploying SNNs to neuromorphic chips with SpikingJelly.} 
		\textbf{a}. Workflow for Intel Loihi. The SNN built by the \textit{BlockContainer} of SpikingJelly can be converted to an SNN in the format of the Lava-DL framework, which is the official framework for the Loihi chip. The SNN can then participate in the standard deployment process, which consists of exporting the SNN from Lava DL to an HDF5 file, reconstructing the SNN from HDF5 in Lava, and compiling the SNN into Loihi.
		\textbf{b}. Workflow for the Lynxi KA200. The SNN built by SpikingJelly can be converted to a Lynxi-compatible SNN by replacing the synapses and neurons with their reduced versions. The Lynxi-compatible SNN can be compiled into model files. Then, KA200 can reconstruct the SNN by loading its model files and run the inference process.
	}
	\label{figure: exchange module}
\end{figure}

Fig.~\ref{figure: exchange module}\textbf{a} is the workflow for Intel Loihi. The official software framework for Loihi is Lava, which compiles SNNs into chips. Lava-DL is a subpackage of Lava, which is based on PyTorch and supports building quantized SNNs with identical behaviors on both CPUs/GPUs and Loihi as long as the SNNs are composed of \textit{Block} modules. Thus, SpikingJelly provides a corresponding \textit{BlockContainer} module, which contains quantized synapses/neurons and has the same behavior as that of the \textit{Block} in Lava-DL. SNNs built by \textit{BlockContainer} can be converted to SNNs in Lava-DL with identical outputs when given the same inputs.

Fig.~\ref{figure: exchange module}\textbf{b} shows the workflow for Lynxi KA200. To run an SNN built by SpikingJelly on KA200, the synapses, and neurons are converted to their simplified versions, which are defined in the exchange module, and only their core functions are retained. For example, the CuPy functions are removed as they only support GPUs. The SNN built with the reduced synapses and neurons is then compatible with the KA200 and can be compiled into model files, which include a JSON file for defining the network shape and a BIN file for recovering the network weights. Finally, KA200 can reconstruct the SNN by loading the model files and run the inference process. 

\subsection*{Neuromorphic Datasets}
\begin{figure}
	\centering
	\includegraphics[width=1.\textwidth,trim=70 720 60 830,clip]{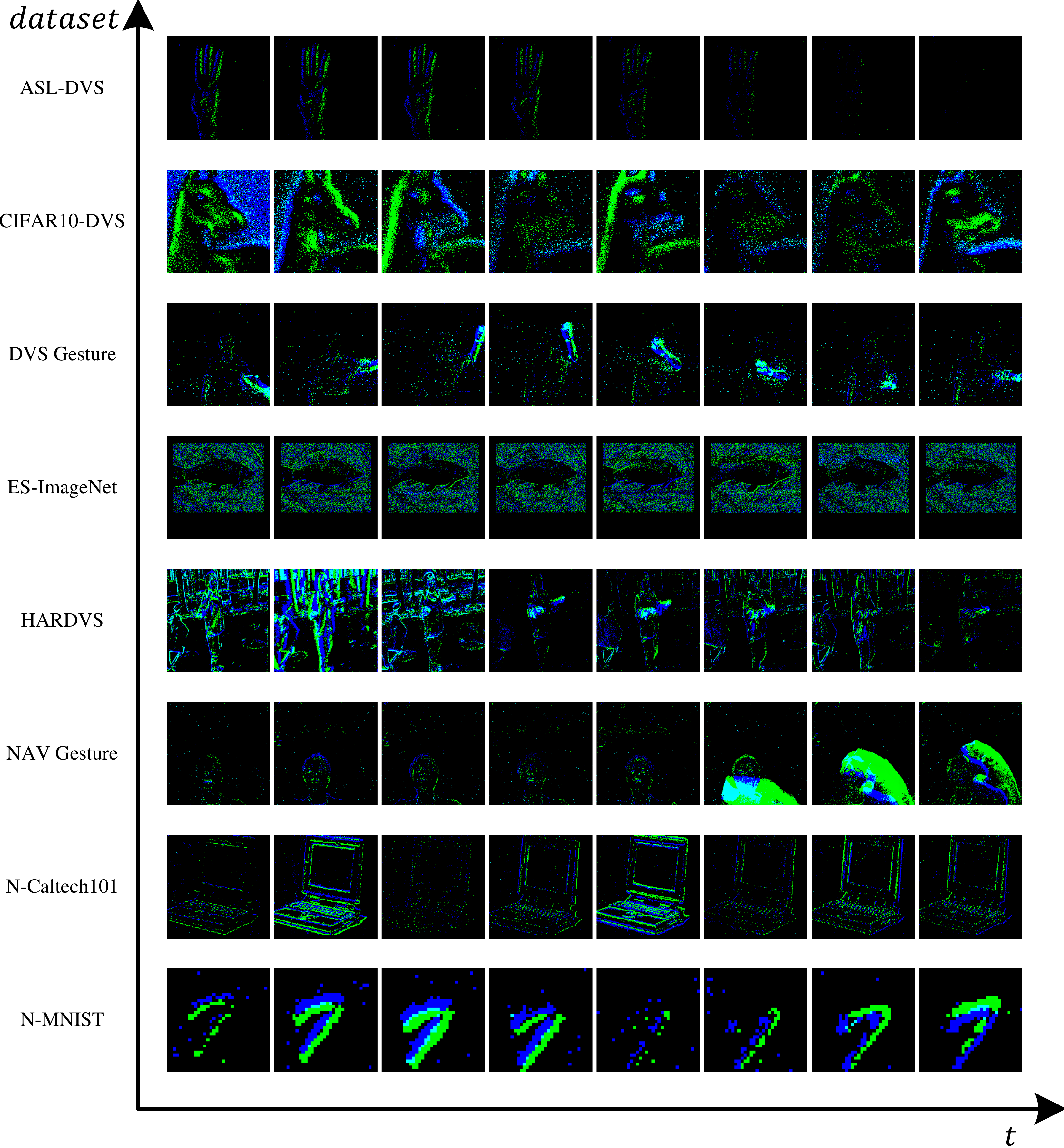}
	\caption{\textbf{Visualizations of all neuromorphic vision datasets integrated into SpikingJelly.} We choose one sample from each dataset and downsample the events to 8 frames.}
	\label{figure: dataset samples}
\end{figure}

Most commonly used neuromorphic datasets have been integrated into SpikingJelly, including ASL-DVS\cite{Bi_2019_ICCV}, CIFAR10-DVS\cite{10.3389/fnins.2017.00309}, DVS Gesture\cite{Amir_2017_CVPR}, ES-ImageNet\cite{10.3389/fnins.2021.726582}, HARDVS\cite{wang2022hardvs} N-Caltech101\cite{10.3389/fnins.2015.00437}, N-MNIST\cite{10.3389/fnins.2015.00437}, Nav Gesture\cite{10.3389/fnins.2020.00275}, and Spiking Heidelberg Digits \cite{shd}. Fig.~\ref{figure: dataset samples} shows visualizations of all of these neuromorphic vision datasets.

\subsection*{Details of Three Applications in Fig.~\ref{figure: cases}}
\textbf{Application a} In this application, we first use the \textit{DVS128Gesture} dataset class from SpikingJelly to process the original event data from the IBM DVS128 Gesture dataset. The original events are sliced and integrated to $T=64$ frames as Eq.~(\ref{eq: integrate event split by number}). For now, modules in Lava only support binary spikes as inputs. Thus, these frames are also binarized by setting all non-zero values as $1$. Meanwhile, we downsampled the size of frames from $128 \times 128$ to $64 \times 64$, which is observed to get higher accuracy in training. We set the batch size to 16. Finally, the inputs for the SNN are binary spikes with a shape of $(T, N, C, H, W) = (64, 16, 2, 64, 64)$.

The network structure is $\{C32k3s1-LIF-C32k2s2-LIF\} * 3 - FC11-LIF$. $C32k3s1$ is the convolution layer with 32 channels, kernel size 3, and stride 1. $C32k2s2$ is the convolution layer with 32 channels, kernel size 2, and stride 2. $LIF$ is the LIF neuron layer. $*3$ means 3 repeated structures. $FC11$ is the fully connected layer with 11 output features. Note that we use the quantization-aware training and all synapses and neuronal dynamics are quantized to 8 bits.

We use the Adam optimizer with a learning rate of 0.005 and the number of training epochs is 128. We use the cosine annealing learning rate schedule \cite{loshchilov2016sgdr} with $T_{max} = 128$. This SNN achieved 85.42\% accuracy on the test set.

All layers of this SNN are composed of modules from the \textit{lava\_exchange} subpackage from SpikingJelly. Thus, this SNN can be converted to the format of Lava. As the conversion workflow shown in Fig.\ref{figure: exchange module}, the SNN is firstly converted to the Lava-DL format, and then the compiled network in the Lava HDF5 format is exported. 
The HDF5 model files can be read by the Lava framework and run in Loihi/Loihi2 or with the Loihi/Loihi2 simulator on a CPU. We use the \textit{Loihi1SimCfg} provided in Lava to run this SNN in CPU with the Loihi simulator and get the same accuracy as running the original SNN in SpkingJelly. The users with access to Loihi cloud Systems provided through Intel vLab Virtual Machines can use the \textit{Loihi1HwCfg} or \textit{Loihi2HwCfg} to run this SNN in Loihi or Loihi2.

Note that this application is designed to demonstrate SpikingJelly's full-stack ability to process neuromorphic datasets, build, train, and deploy SNNs, rather than achieve state-of-the-art accuracy. We did not use large networks and complex training tricks, which resulted in the plain 85.42\% accuracy. It is worth noting that \cite{shymyrbay2023low} uses SpikingJelly and achieves 95.45\% accuracy on the DVS Gesture dataset with SNNs quantized to only 1 bit.

\textbf{Application b} In this application, we first obtain the observation from the state of the environment in the OpenAI Gym. There are notable differences in the state and action dimensions for different tasks, which affect the network structure. Taking Ant-v3 in the figure as an example, its state dimension $N_S$ is 111, and its action dimension $N_A$ is 8. Since we set the batch size as 100, the observation is floating-point values with a shape of $(N,N_S)=(100,111)$. 

Similar to the population encoder proposed by \cite{tang2021deep}, the spike encoder uses a population of neurons with different learnable Gaussian receptive fields to encode each state dimension into a spike train. Finally, the inputs for the SNN are binary spikes with a shape of $(T,N,N_S \cdot P_{in})=(5,100,1110)$, where $P_{in}$ represents the input population sizes per state dimension ($P_{in}=10$). The main difference is that our application uses the arctangent function as the surrogate function. The SNN connects the spike encoder and the spike decoder, whose input and output are spike trains. The network structure of the SNN is $FC256-LIF-FC256-LIF-FC80-LIF$. $LIF$ is the LIF neuron layer. $FC80$ is the fully connected layer with 80 output features, which equals to $N_A \cdot P_{out}$. We equally divide the spiking neurons of the last layer into $N_A$ output population with a size of $P_{out}$, which equals 10. Each output population has a corresponding spike decoder. Each spike decoder consists of a learnable layer ($FC1$) and an integrated neuron. The integrated neuron continuously accumulates the output of the learnable layer into its membrane voltage without firing. After every $T$ simulation time steps, the input of the spike decoder is decoded into the last membrane voltage of the integrated neuron, which represents the corresponding value of the action dimension.
Our spiking actor network (SAN) is functionally equivalent to a deep actor network (DAN), which can be trained in conjunction with a deep critic network using TD3 algorithms \cite{fujimoto2018addressing}. During training, SAN builds a mapping between states and actions to represent the agent's policy. The deep critic network estimates the corresponding Q-value, which can guide the SAN to learn a better policy. During evaluation, the trained SAN is able to predict the action with the largest Q-value produced by the trained critic network.
To ensure reproducibility, each of the models is trained for ten rounds, corresponding to ten random seeds. In each round, the task is trained for 1 million steps and evaluated every 10K steps, where each evaluation reports the average reward over 10 episodes using the deterministic policy. Each episode can last up to 1000 execution steps. For other hyper-parameters, we use the same configurations as the open-source code of PopSAN \cite{tang2021deep}. Each module in our SAN can be implemented by the layer and neuron subpackages.

We compare the performance of our SAN with PopSAN, taking the average performance ratio (APR) of different SANs to the corresponding DANs across all the tasks as the measurement standard. Our SAN achieved 88.45\% APR on the four most commonly used tasks from OpenAI gym, including Ant-v3, HalfCheetah-v3, Hopper-v3, and Walker2d-v3. Compared to our SAN implemented by SpikingJelly, PopSAN not only has a lower APR of 71.94\%, but also has a slower training speed. This application demonstrates the potential of SpikingJelly as an energy efficient alternative for real-time robotic control tasks.

\textbf{Application c} In this application, we first use the module \textit{activation\_based.model} from SpikingJelly to pretrain the SEW ResNet \cite{SEWResNet} on the ImageNet dataset. The network depths are 18, 34, 50, 101, and 152. For all networks, we use the same training settings and set the simulation time steps $T=4$. Each network is trained for 320 epochs with a mini-batch size of 32. The optimizer is SGD with momentum 0.9. The initial learning rate is 0.1 and decays by the cosine annealing learning rate schedule, where $T_{max}=320$ is the same as the number of epochs. All of these networks achieve an accuracy of 63\% or more on the ImageNet.

After pre-training these networks, we evaluate their brain-likeness by comparing their neural representations with those of biological visual systems. For biological visual systems, we take the neural response when the animal is presented with some static image as the neural representation. For the network, we feed the same static images used for the biological visual system, with the same simulation time steps as the training procedure. We then extract features from all layers of the network as neural representations. Finally, we apply three metrics \cite{huang2023deep} to compute the similarity between the neural representations of the networks and the biological visual system.

We compare the representational similarity of SEW ResNet with its counterparts of CNN (ResNet) and find out that the similarity of the former are higher than the latter with an average of 6.6\%. With this application, we demonstrate that SpikingJelly can be used as an effective tool for constructing more brain-like computational models.

\subsection*{Performance Comparison on Recurrent Structure}

\begin{figure}
	\centering
	\includegraphics[width=1.\textwidth,trim=75 66 160 60,clip]{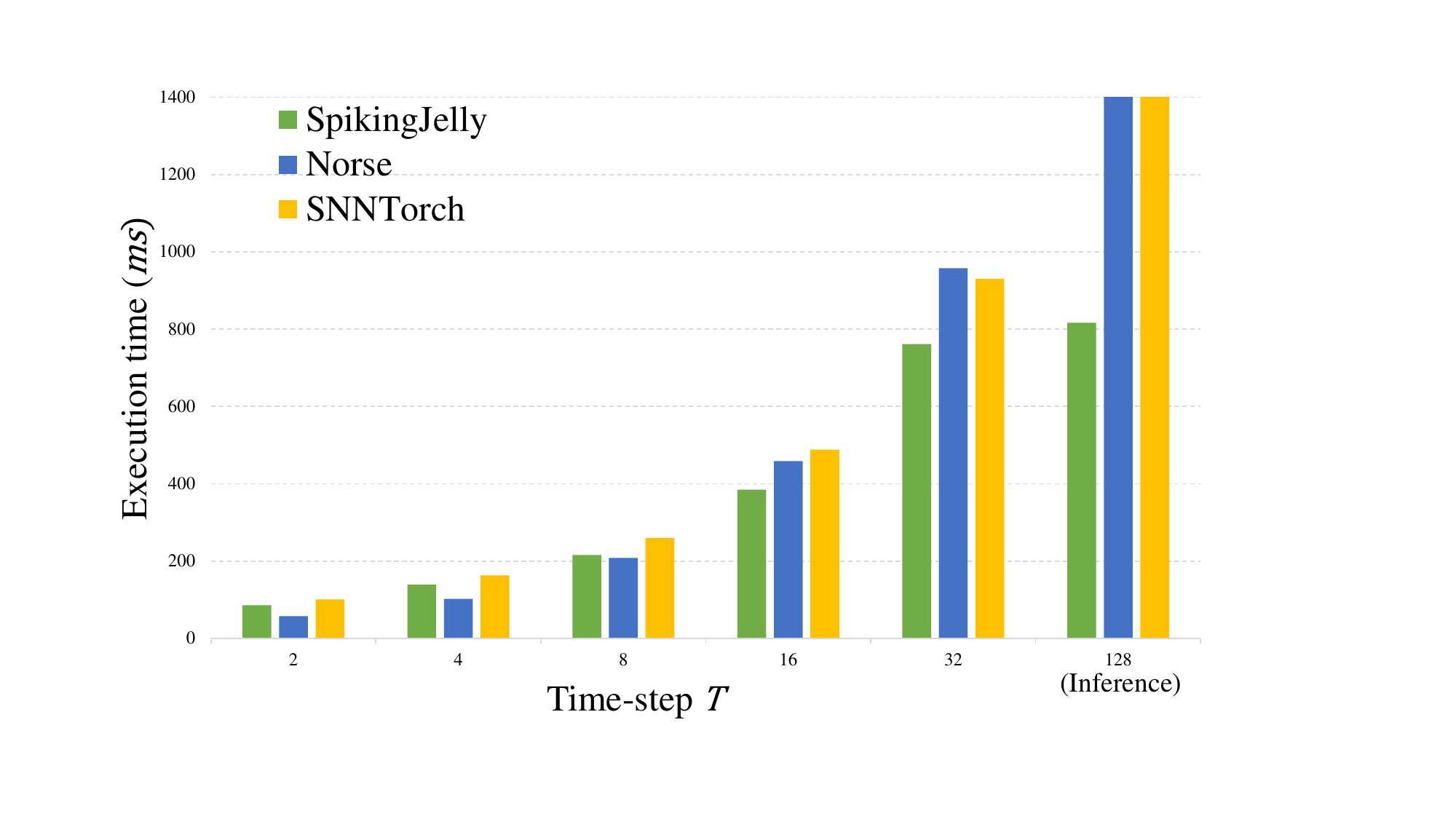}
	\caption{The execution times required for a single training iteration with $T=2,4,8,16,32$ and inference with $T=128$ using recurrent SEW ResNet-18 \cite{huang2023deep} built by SpikingJelly, Norse and SNNTorch.}
	\label{figure: recurrent snn speed}
\end{figure}

Fig.~\ref{figure: framework}\textbf{d} has shown the performance comparison on the feedforward SNN with SpikingJelly, Norse, and SNNTorch. Compared with the feedforward SNNs that are widely used for static and neuromorphic datasets classification, the recurrent SNNs have the potential of high ability in long-term learning and have been used in temporal tasks \cite{yin2021accurate, rao2022long, huang2023deep}. To compare the performance of different frameworks on simulating the recurrent SNN, we conduct the experiments of benchmarking the recurrent SEW ResNet-18 used in \cite{huang2023deep}. 
The experiment options are the same as those in Fig.~\ref{figure: framework}\textbf{d}, and the results are shown in Fig.~\ref{figure: recurrent snn speed}. Since the recurrent SEW ResNet uses recurrent connections in all four stages and only the single-step mode can be used, the acceleration methods based on the multistep mode in SpikingJelly cannot be applied. In this case, JIT is the only speedup method in SpikingJelly. For training, the experiment results show that when $T$ is small, e.g., $T=2, 4, 8$, the speed rank is Norse $ > $ SpikingJelly $>$ SNNTorch. 
When $T$ is large, e.g., $T=16, 32$, SpikingJelly is faster than Norse, and the SNNTorch is the slowest. The result of inference is identical to that in Fig.~\ref{figure: framework}\textbf{d}, and SpikingJelly only takes 0.57$\times$ execution times of those in Norse and SNNTorch, which take close execution times in inference. The source code and data of these experiments are also provided in the supplementary materials.

As a conclusion, we can find that SpikingJelly, Norse and SNNTorch achieve close performance in training fully recurrent structures. Norse is faster when $T$ is small, and SpikingJelly is faster when using a large $T$. For inference, SpikingJelly remains superior in terms of speed. Note that for the SNNs with partial recurrent connections, the hybrid propagation patterns can be used as we have shown in the section \textit{Using Examples}, and the user can still use the acceleration methods based on the multistep mode in the feedforward part of SNNs.

\subsection*{Limitations and Future Development Plans}
Although SpikingJelly is a mature framework after three years of development, it still has some limitations.  Based on the experience of the developers and discussions with users, we summarize the current limitations and potential future development plans below.

\textbf{Toolkits for timing-based SNNs} Most of the deep SNNs in recent years use the activation-based representation method, which uses binary tensors $S$ as spikes. The early research such as SpikeProp \cite{BOHTE200217}, Tempotron \cite{tempotron} uses the timing-based representation method, which uses the firing times $t_{f}$ as spikes. For example, suppose the time unit is 1, then $S=[0, 1, 0, 1]$ in the activation-based representation is identical to $t_{f} = [1, 3]$ in the timing-based representation. It can be found that timing-based representations are sparser, memory efficient, and have the potential to work in continuous time instead of discrete time steps.
However, due to the difficulty of implementing complex operations such as convolutions and matrix multiplication, the time-based SNNs advance much slower than the activation-based SNNs. Correspondingly, although there exists a \textit{timing\_based} subpackage in SpikingJelly, the development of this subpackage is primitive.
Considering the promising advantages of the timing-based representation, designing toolkits for timing-based SNNs and completing the \textit{timing\_based} subpackage is a worthwhile development plan.

\textbf{Sparse acceleration} Although spikes between layers are sparse in SNNs, these spikes are still stored in dense tensors, which are compatible with frequently-used operations in deep learning, but do not make full use of the sparse characteristic. When the firing rates, which are also the sparsities, of spikes are low enough, training SNNs with sparse computation libraries such as cuSPARSE for basic linear algebra and Minkowski Engine \cite{choy20194d} for sparse convolutions/poolings can reduce memory consumption and accelerate simulation speed. In the early 0.0.0.0.4 version of SpikingJelly, there exists a \textit{AutoSparseLinear} layer, which runs benchmark in the first running to get the critical sparsity. The critical sparsity is the sparsity at which sparse matrix multiplication and dense matrix multiplication have the same speed. Sparse matrix multiplication is used if the sparsity of the input spikes is higher than the critical sparsity, otherwise, dense matrix multiplication is used. Thus, the \textit{AutoSparseLinear} layer can automatically use the fastest matrix multiplication during training. However, exploration of sparse acceleration has been put on hold since version 0.0.0.0.4 due to the limited time and effort of the developers. In the future, we will also consider restarting the development on sparse acceleration.

\textbf{Automatic CUDA code generation for complex neurons} Implementing complex spiking neurons without the CuPy backend does not take much developing cost. However, the complex neuronal dynamics indicate tiny but numerous calling of CUDA kernels by PyTorch, which slows down the training of SNNs especially when $T$ is large. As we have shown in Fig.~\ref{figure: accelceration}\textbf{c}, even the IF neuron with the simplest neuronal dynamics still suffers from this issue.
The slow training speed may be the reason why complex spiking neurons are rarely used in deep SNNs. Correspondingly, there is only rare research that uses the spiking neuron with the same complexity at the level of the Izhikevich neuron \cite{jin2022sit} and high-order spiking neurons \cite{zhang2022multi} in deep SNNs.
The use of a large CUDA kernel to fuse all neuronal dynamics as a CuPy backend for complex neurons solves the problem of slow simulation speed. However, writing CUDA codes for both forward and backward propagations consumes a lot of developer effort. The semiautomatic code generation technique in SpikingJelly solves this problem partly but is not the most perfect solution. The ideal CUDA code generation method is fully automatic and generates CUDA codes from Python codes that define neuronal dynamics. In the 0.0.0.0.14 version of SpikingJelly, a primary generator subpackage \textit{spikingjelly.activation\_based.auto\_cuda.generator} is introduced as the prototype of the automatic CUDA code generation. Although this subpackage is not yet complete, the development roadmap has identified that the generator creates a computational graph based on the neuronal dynamics defined in the Python code, analyzes the variables in the computational graph and their mathematical relationships, and generates CUDA codes based on the variables.

\subsection*{Statistical Trends}
\begin{figure}
	\centering
	\includegraphics[width=1.\textwidth,trim=20 660 280 20,clip]{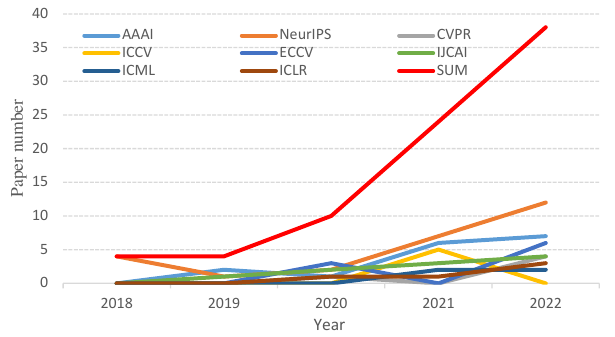}
	\caption{\textbf{The numbers of accepted papers related to spiking deep learning at the top AI conferences.} Note that "SUM" is the summation of the values for all conferences. It can be found that research interest began to increase in 2019.}
	\label{figure: ai cf paper number}
\end{figure}

We counted the number of accepted papers related to spiking deep learning at the top AI conferences, which are shown in Fig.~\ref{figure: ai cf paper number}. The chosen conferences included the AAAI Conference on Artificial Intelligence (AAAI), the Annual Conference on Neural Information Processing Systems (NeurIPS), the IEEE Conference on Computer Vision and Pattern Recognition (CVPR), the International Conference on Computer Vision (ICCV), the European Conference on Computer Vision (ECCV), the International Joint Conference on Artificial Intelligence (IJCAI), the International Conference on Machine Learning (ICML) and International Conference on Learning Representation (ICLR), which are well recognized as the primary conferences of the AI research community. Since 2019, there has been a dramatic increase in the number of accepted papers on spiking deep learning, indicating that research interest also started to increase at that time. Correspondingly, SpikingJelly was opened-sourced in December 2019. We also counted the number of publications using SpikingJelly during each year starting from 2020; these included 1 publication in 2020, 10 publications in 2021, 42 publications in 2022, and 17 publications in 2023 (until Mar.1, 2023). These statistical results demonstrate that SpikingJelly fulfills the need of researchers for a spiking deep learning toolkit and accelerates community boosting.

\end{document}